\documentclass[10pt,twocolumn,letterpaper]{article}

\usepackage{cvpr}              
\usepackage[accsupp]{axessibility}
\usepackage{graphicx}
\usepackage{amsmath}
\usepackage{amssymb}
\usepackage{booktabs}
\usepackage{multirow}
\usepackage{comment}
\usepackage{amsmath,amssymb} 
\usepackage{color}
\usepackage{graphicx}
\usepackage{booktabs}
\usepackage{bbm}
\usepackage{makecell}
\usepackage{bm}
\usepackage{multicol}
\usepackage{multirow}
\usepackage{caption}
\usepackage{xcolor}
\usepackage{bbding}
\usepackage{pifont}
\usepackage{wrapfig}

\def\red#1{\textcolor{red}{#1}}

\usepackage[pagebackref,breaklinks,colorlinks]{hyperref}

\usepackage[capitalize]{cleveref}
\crefname{section}{Sec.}{Secs.}
\Crefname{section}{Section}{Sections}
\Crefname{table}{Table}{Tables}
\crefname{table}{Tab.}{Tabs.}
\makeatletter
\newcommand{\printfnsymbol}[1]{%
  \textsuperscript{\@fnsymbol{#1}}%
}
\makeatother

\def\cvprPaperID{3685} 
\def\confName{CVPR}
\def\confYear{2023}

\begin{document}

\title{Backdoor Defense via Adaptively Splitting Poisoned Dataset}

\author{Kuofeng Gao\textsuperscript{\rm 1}\thanks{Equal contribution.}, \ Yang Bai \textsuperscript{\rm 2}\printfnsymbol{1}, \ Jindong Gu\textsuperscript{\rm 3}\thanks{Corresponding author.}, \  Yong Yang\textsuperscript{\rm 4},  \ Shu-Tao Xia\textsuperscript{\rm 1\rm 5}\printfnsymbol{2}\\
\textsuperscript{\rm 1} Tsinghua University \quad
\textsuperscript{\rm 2} Tencent Security Zhuque Lab \quad
\textsuperscript{\rm 3} University of Oxford \\
\textsuperscript{\rm 4} Tencent Security Platform Department \quad
\textsuperscript{\rm 5} Peng Cheng Laboratory
\\
\tt\small gkf21@mails.tsinghua.edu.cn, \{mavisbai,coolcyang\}@tencent.com \\
\tt\small jindong.gu@eng.ox.ac.uk, xiast@sz.tsinghua.edu.cn
}


\maketitle

\begin{abstract}
Backdoor defenses have been studied to alleviate the threat of deep neural networks (DNNs) being backdoor attacked and thus maliciously altered. Since DNNs usually adopt some external training data from an untrusted third party, a robust backdoor defense strategy during the training stage is of importance. We argue that the core of training-time defense is to select poisoned samples and to handle them properly. 
In this work, we summarize the training-time defenses from a unified framework as splitting the poisoned dataset into two data pools. Under our framework, we propose an \underline{a}daptively \underline{s}plitting dataset-based \underline{d}efense (ASD). Concretely, we apply loss-guided split and meta-learning-inspired split to dynamically update two data pools. With the split clean data pool and polluted data pool, ASD successfully defends against backdoor attacks during training. Extensive experiments on multiple benchmark datasets and DNN models against six state-of-the-art backdoor attacks demonstrate the superiority of our ASD. Our code is available at \url{https://github.com/KuofengGao/ASD}.

\end{abstract}

\section{Introduction}
\label{sec:intro}
Backdoor attacks can induce malicious model behaviors by injecting a small portion of poisoned samples into the training dataset with specific trigger patterns. The attacks have posed a significant threat to deep neural networks (DNNs) \cite{ma2022visual,li2014common,li2015learning,liu2006spatio,li2016mutual,tang2004video}, especially when DNNs are deployed in safety-critical scenarios, such as autonomous driving \cite{ding2019trojan}. To alleviate the threats, backdoor defenses have been intensively explored in the community, which can be roughly grouped into post-processing defenses and training-time ones. Since the training data collection is usually time-consuming and expensive, it is common to use external data for training without security guarantees \cite{gu2017badnets,shafahi2018poison,qi2023revisiting,li2023backdoorbox,li2022untargeted}. The common practice makes backdoor attacks feasible in real-world applications, which highlights the importance of training-time defenses. We argue that such defenses are to solve two core problems, \textit{i.e.}, to select poisoned samples and to handle them properly.

\begin{table}[t]
\footnotesize
\caption{Summary of the representative training-time backdoor defenses under our framework.}
\label{summary abl, dbd and asd}
\vspace{-2em}
\begin{center}
\setlength{\tabcolsep}{0.87mm}
{
\begin{tabular}{ccccc}
\toprule[0.68pt]
\multirow{2}{*}{Methods}  & {\# Pool} & {\# Pool} & {\# Pool} & {\# Clean Hard} \\
 & Initialization & Maintenance & Operation & Sample Selection \\
\midrule
ABL & Fast & Static & Unlearn & No\\
DBD & Slow & Adaptive & Purify & No\\
ASD (Ours) & Fast & Adaptive & Purify & Yes\\
\bottomrule[0.68pt]
\end{tabular}}
\end{center}
\vspace{-1.5em}
\end{table}

In this work, we formulate the training-time defenses into a unified \textit{framework} as splitting the poisoned dataset into two data pools. Concretely, a clean data pool contains selected clean samples with trustworthy labels and a polluted data pool is composed of poisoned samples and remaining clean samples. Under this framework, the mechanisms of these defenses can be summarized into three parts, \textit{i.e.}, \textit{pool initialization}, \textit{pool maintenance}, and \textit{pool operation}. To be more specific, they need to first initialize two data pools, deploy some data pool maintenance strategies, and take different training strategies on those split (clean and polluted) data pools. We illustrate our framework with two representative training-time defenses, \textit{i.e.}, anti-backdoor learning (ABL) \cite{li2021anti} as well as decoupled-based defense (DBD) \cite{huang2022backdoor}. ABL statically initializes a polluted pool by the loss-guided division. The polluted pool is fixed and unlearned during training. Similarly, DBD initializes two data pools after computationally expensive training. Then the model is fine-tuned by semi-supervised learning with two dynamically updated data pools. (More details of these two methods are introduced in Sec. \ref{sec:relatedwork_defense}.)

Despite their impressive results, there is still room for improvement. ABL initializes two pools with static data selection, which raises the concern of mixing poisoned data into clean ones. Once they are mixed, it is hard to alleviate it in the followed training process. Besides, unlearning poisoned data directly can lose some useful semantic features and degrade the model's clean accuracy. As for DBD, its pool initialization is computationally expensive and is hard to be implemented end-to-end. Moreover, DBD adopts supervised learning for the linear layer in the whole poisoned dataset during the second stage, which can potentially implant the backdoor in models. 

\begin{figure}[t]
	\centering
    \begin{subfigure}[]{0.32\linewidth}
        \includegraphics[width=\linewidth]{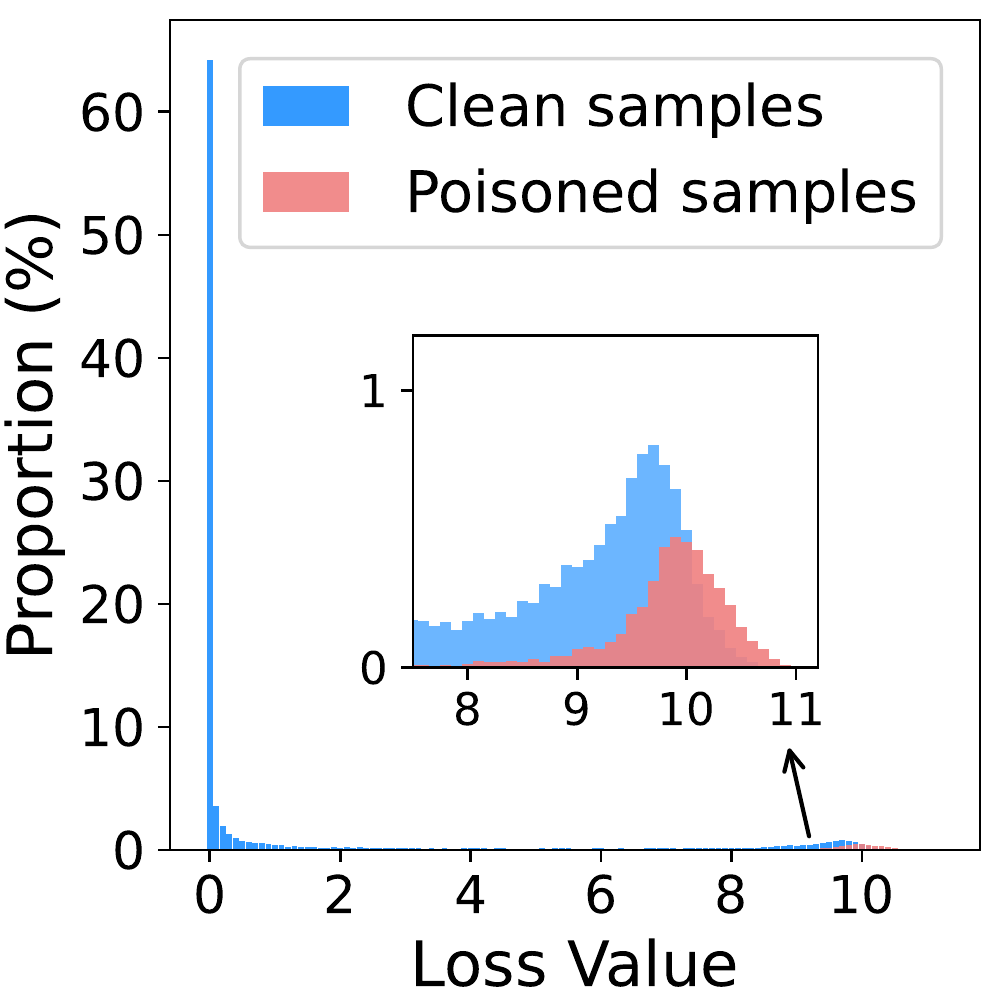}
        \caption{ABL}
        \label{intro_ABL}
    \end{subfigure}
    \begin{subfigure}[]{0.32\linewidth}
        \includegraphics[width=\linewidth]{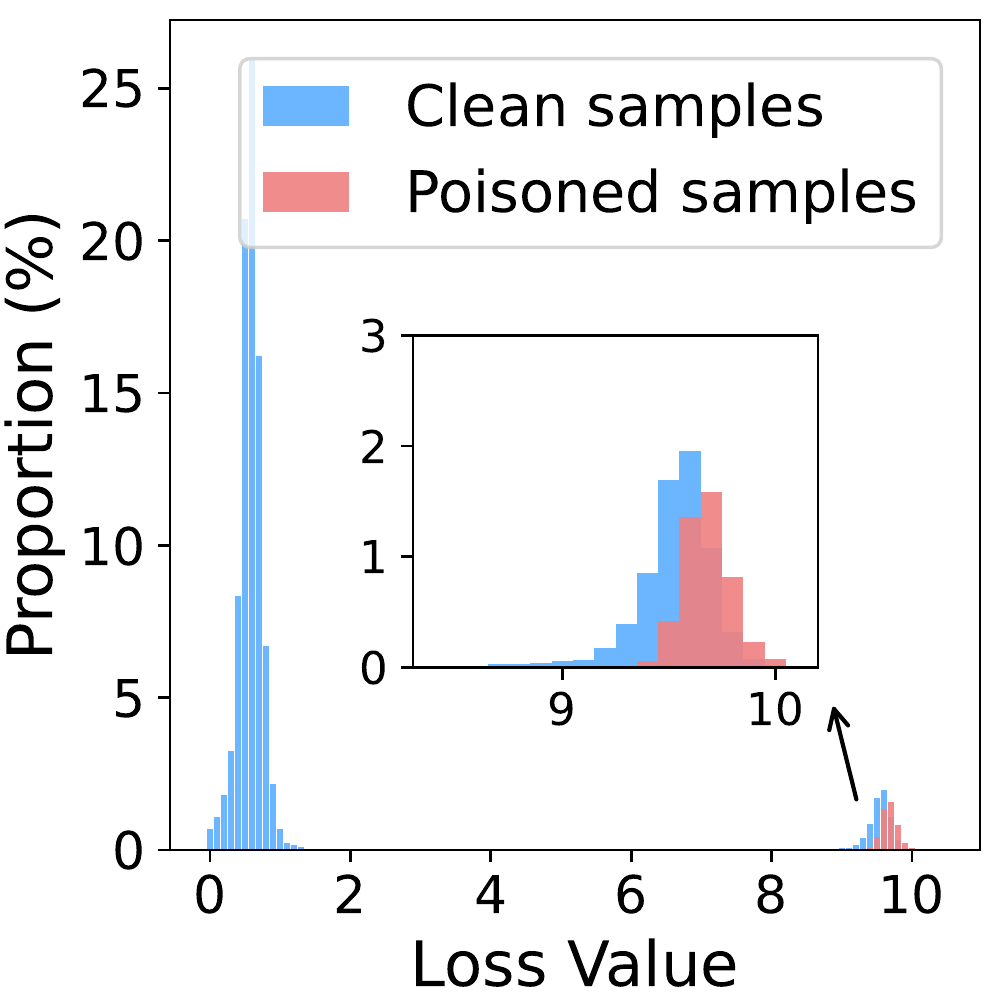}
        \caption{DBD}
        \label{intro_DBD}
    \end{subfigure}
    \begin{subfigure}[]{0.32\linewidth}
        \includegraphics[width=\linewidth]{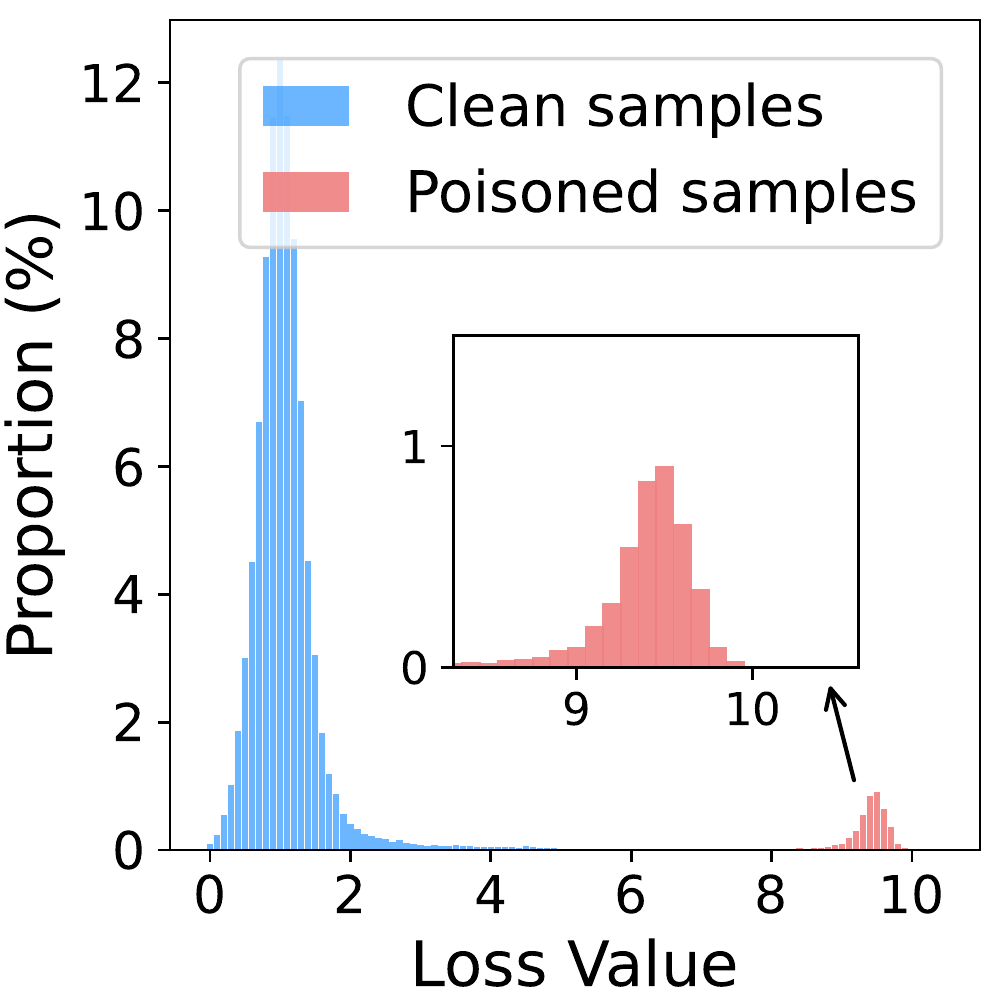}
        \caption{ASD (Ours)}
        \label{intro_wanet_loss_epoch120}
    \end{subfigure}
	\centering
 \vspace{-0.5em}
	\caption{Loss distribution of samples on the model trained by ABL, DBD and our ASD against WaNet. Compared with ABL and DBD, our proposed ASD can clearly separate clean samples and poisoned ones better by a novel meta-split.}
 \label{fig:intro}
\end{figure}

Under our framework, we introduce an \underline{a}daptively \underline{s}plitting dataset-based \underline{d}efense (ASD). With two initialized data pools, we first adopt the loss-guided \cite{wang2019symmetric} split to update two data pools. However, some (model-dependent) clean hard samples can not be distinguished from poisoned ones directly by their loss magnitudes. As shown in Fig. \ref{fig:intro}, ABL and DBD adopting loss-guided split have failed to completely separate clean samples from poisoned samples. Instead, we propose a novel meta-learning-inspired split (meta-split), which can make a successful separation.
Then we treat the clean data pool as a labeled data container and the polluted one as unlabeled, where we adopt semi-supervised learning on two data pools.
As such, we can utilize the semantic information of poisoned data without labels to keep the clean accuracy meanwhile to avoid backdoor injection, which can be regarded as a fashion of purifying poisoned samples. Note that, our ASD introduces clean seed samples (\textit{i.e.}, only 10 images per class) in pool initialization, which could be further extended to a transfer-based version, by collecting clean seed samples from another classical dataset. Given previous methods \cite{liu2018fine,li2021neural,wu2021adversarial,wang2022trap} usually assume they can obtain much more clean samples than ours, our requirements are easier to meet.
The properties of ABL, DBD and our ASD are briefly listed in Table \ref{summary abl, dbd and asd}.

In summary, our main contributions are three-fold: 
\begin{itemize}
\vspace{-0.5em}
\item We provide a framework to revisit existing training-time backdoor defenses from a unified perspective, namely, splitting the poisoned dataset into a clean pool and a polluted pool. Under our framework, we propose an end-to-end backdoor defense, ASD, via splitting poisoned dataset adaptively. 
\vspace{-0.5em}
\item We propose a fast pool initialization method and adaptively update two data pools in two splitting manners, \textit{i.e.}, loss-guided split and meta-split. Especially, the proposed meta-split focuses on how to mine clean hard samples and clearly improves model performance.
\vspace{-0.5em}
\item With two split data pools, we propose to train a model on the clean data pool with labels and the polluted data pool without using labels. Extensive experiment results demonstrate the superiority of our ASD to previous state-of-the-art backdoor defenses.
\end{itemize}

\section{Related Work}
\label{sec:related work}

\subsection{Backdoor Attack}
Backdoor attacks are often implemented by injecting a few poisoned samples to construct a poisoned dataset. Once a model is trained on the constructed poisoned dataset, the model will perform the hidden backdoor behavior, \textit{e.g.}, classifying samples equipped with trigger to the target label. Apart from the malicious behavior, the backdoored model behaves normally when the trigger is absent. Existing backdoor attacks can be divided into two categories: (1) \textbf{poison-label backdoor attacks} \cite{gu2017badnets,wenger2021backdoor,zhao2022defeat,li2021backdoor,bai2022hardly} connect the trigger with the target class by relabelling poisoned samples as target labels.
Trigger patterns have been designed to enhance the attack strength \cite{gu2017badnets,nguyen2020input} or stealthiness \cite{chen2017targeted,nguyen2021wanet,liu2020reflection}. 
(2) \textbf{clean-label backdoor attacks} \cite{turner2018clean,gan2021triggerless} only poison the samples from the target class with labels unchanged. Although they are stealthier than poison-label attacks, clean-label attacks may fail to implant the model sometimes \cite{li2022backdoor,zhao2020clean}. 

\begin{figure*}[t]
\centering
    \includegraphics[width=\linewidth]{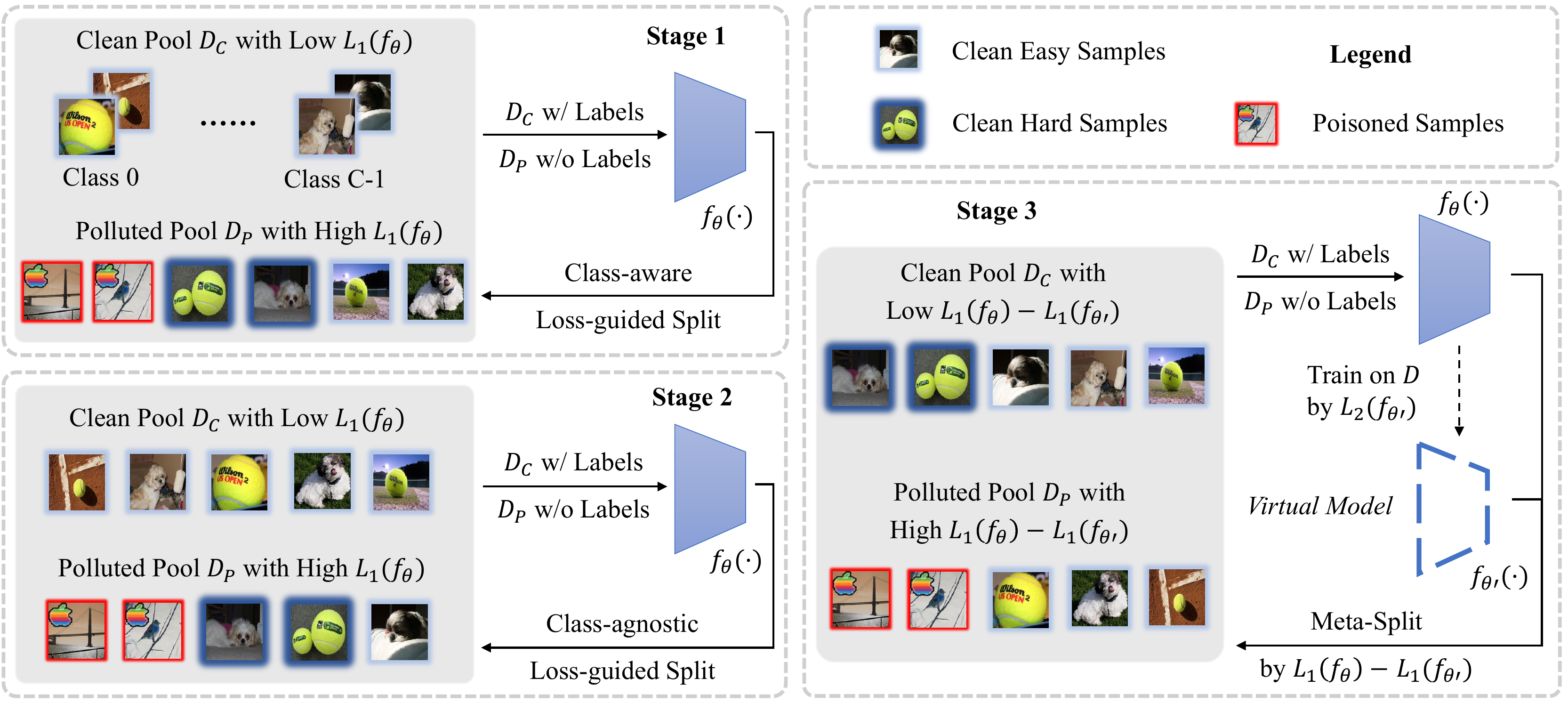}
\caption{Pipeline of our ASD. It mainly contains three stages. (1) Clean data pool $\mathcal{D}_{C}$ is initialized with a few clean seed samples. Polluted data pool $\mathcal{D}_{P}$ is initialized with the poisoned training dataset $\mathcal{D}$. Then $\mathcal{D}_{C}$ is supplemented by samples in $\mathcal{D}$ with the lowest $\mathcal{L}_1(f_{\bm{\theta}})$ losses, which are equally selected from each class, dubbed \textit{class-aware loss-guided split}. (2) $\mathcal{D}_{C}$ is substantially supplemented by samples with the lowest $\mathcal{L}_1(f_{\bm{\theta}})$ losses in the entire dataset $\mathcal{D}$, dubbed \textit{class-agnostic loss-guided split}. (3) A novel \textit{meta-split} is proposed to add clean hard samples into $\mathcal{D}_{C}$ and further improve the performance. Note that two data pools are adaptively updated at every epoch, where $\mathcal{D}_{C}$ is directly supplemented and the remaining samples in $\mathcal{D}$ are divided as $\mathcal{D}_{P}$. During the whole process, labels in $\mathcal{D}_{C}$ are preserved while labels in $\mathcal{D}_{P}$ are removed. The model $f_{\bm{\theta}}$ is trained on $\mathcal{D}_{C}$ and $\mathcal{D}_{P}$ by semi-supervised learning following Eq. \ref{eq:formulation semi-supervised 2 loss}.}
\label{Pipeline}
\end{figure*}

\subsection{Backdoor Defense}
\label{sec:relatedwork_defense}
Recent work has explored various methods to mitigate the backdoor threat. Existing backdoor defenses can be grouped into two categories: (1) \textbf{post-processing backdoor defenses} \cite{gao2019strip,kolouri2020universal,zeng2021adversarial} aim to repair a local backdoored model with a set of local prepared data. A straightforward way is to reconstruct the trigger pattern \cite{wang2019neural,dong2021black,chen2022quarantine,tao2022better,guan2022few} and then to unlearn the trigger pattern for repairing the model. Apart from the trigger-synthesis defenses, other methods are also widely used in erasing the backdoor, \textit{e.g.}, pruning \cite{liu2018fine,wu2021adversarial,zheng2022data}, model distillation \cite{li2021neural} and mode connectivity \cite{zhao2020bridging}. (2) \textbf{training-time backdoor defenses} \cite{tran2018spectral,weber2020rab,wu2022backdoordefense} intend to train a clean model directly on the poisoned dataset. Anti-backdoor learning (ABL) \cite{li2021anti} observes that the training losses of poisoned samples drop abruptly in the early training stage. Thus ABL proposes to first isolate a few samples with the lowest losses in early epochs, then train a model without isolated samples and finally unlearn isolated samples during the last few training epochs. Decouple-based backdoor defense (DBD) \cite{huang2022backdoor}, first adopts self-supervised learning to obtain the feature extractor. Then it uses supervised learning to update the linear layer. After identifying clean samples by the symmetric cross-entropy loss, DBD conducts semi-supervised learning on the labeled clean data and unlabeled remaining data. Besides, adopting differential-privacy SGD \cite{du2019robust} and strong data augmentation \cite{borgnia2021strong} can also defend against backdoor attacks to some degree. Our proposed ASD belongs to the training-time backdoor defense.


\section{Preliminaries}
\label{sec:preliminaries}
\noindent \textbf{Threat model}. We adopt the poisoning-based threat model used in previous works \cite{gu2017badnets,turner2018clean,chen2017targeted}, where attackers provide a poisoned training dataset containing a set of pre-created poisoned samples. Following previous training-time defenses \cite{huang2022backdoor,li2021anti,borgnia2021strong,du2019robust}, we assume that defenders can control the training process. Besides, a few clean samples of each class are available as seed samples. The goal of defenders is to obtain a well-performed model without suffering backdoor attacks. 
\par

\noindent \textbf{Problem formulation}. Given a classification model $f_{\bm{\theta}}$ with randomly initialized parameters $\bm{\theta}$ and a training dataset $\mathcal{D}=\{(\bm{x}_i,y_i)\}^N_{i=1}$, the training dataset $\mathcal{D}$ contains $N$ samples $\bm{x}_i \in \mathbb{R}^{d}, i=1,..., N$, and their ground-truth labels $y_i \in \{0,1,...,C-1\}$ where $C$ is the number of classes. Poisoned samples might be included in $\mathcal{D}$. Under our unified framework, we propose to divide the dataset $\mathcal{D}$ into two disjoint data pools adaptively, \textit{i.e.}, a clean data pool $\mathcal{D}_{C}$ with labels and a polluted data pool $\mathcal{D}_{P}$, whose labels will not be used. Moreover, we train the model on the clean data pool and polluted data pool in a semi-supervised learning-based manner by treating the polluted pool as unlabeled data, denoted as:
\begin{equation}
\begin{aligned}
\min_{\bm{\theta}} \mathcal{L}\left(\mathcal{D}_{C}, \mathcal{D}_{P}; \bm{\theta}\right),
\end{aligned}
\label{eq:formulation semi-supervised}
\end{equation}
where $\mathcal{D}_{C} \subset \mathcal{D}$ and $\mathcal{D}_{P}=\{\bm{x} | (\bm{x}, y) \in \mathcal{D} {\backslash} \mathcal{D}_{C}\}$. $\mathcal{L}(\cdot)$ denotes the semi-supervised loss function. In view of previous semi-supervised learning methods \cite{berthelot2019mixmatch,berthelot2019remixmatch,zheng2022simmatch,oh2021distribution}, $\mathcal{L}(\cdot)$ usually contains two losses:
\begin{equation}
\begin{aligned}
\mathcal{L}=\sum_{(\bm{x},y) \in \mathcal{D}_{C}} \mathcal{L}_s(\bm{x}, y; \bm{\theta})+\lambda \sum_{\bm{x} \in \mathcal{D}_{P}} \mathcal{L}_u(\bm{x}; \bm{\theta}),
\end{aligned}
\label{eq:formulation semi-supervised 2 loss}
\end{equation}
where the supervised loss $\mathcal{L}_s$ is adopted on the clean data pool $\mathcal{D}_{C}$ with labels, the unsupervised loss $\mathcal{L}_u$ is used on the polluted data pool $\mathcal{D}_{P}$ without using labels, $\lambda$ denotes the trade-off between $\mathcal{L}_s$ and $\mathcal{L}_u$.
Since $\mathcal{L}_s$ can obtain the precise relationship between images and labels. Hence, it is critical to ensure as many clean samples and few poisoned samples as possible in the clean data pool $\mathcal{D}_{C}$.

\begin{figure*}[t]
	\centering
    \begin{subfigure}[]{0.24\linewidth}
        \includegraphics[width=\linewidth]{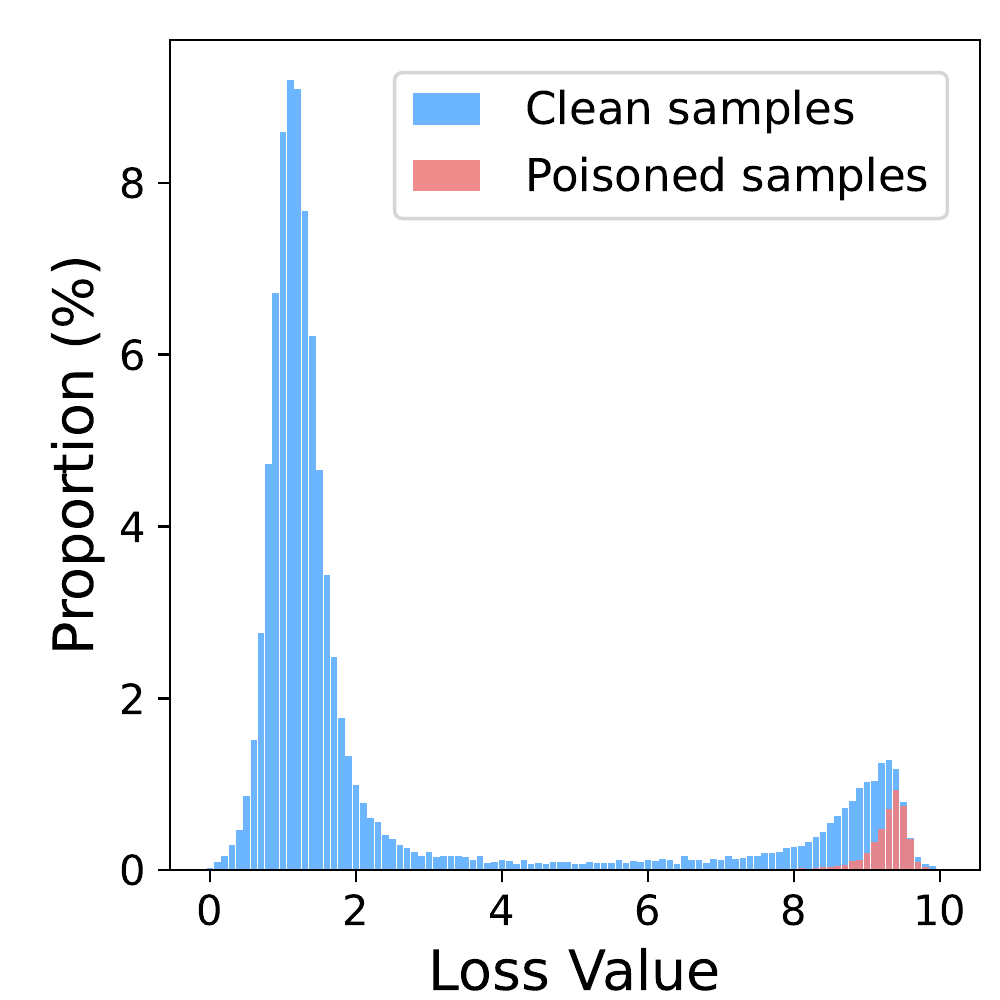}
        \caption{}
        \label{badnets_loss_epoch91}
    \end{subfigure}
    \begin{subfigure}[]{0.24\linewidth}
        \includegraphics[width=\linewidth]{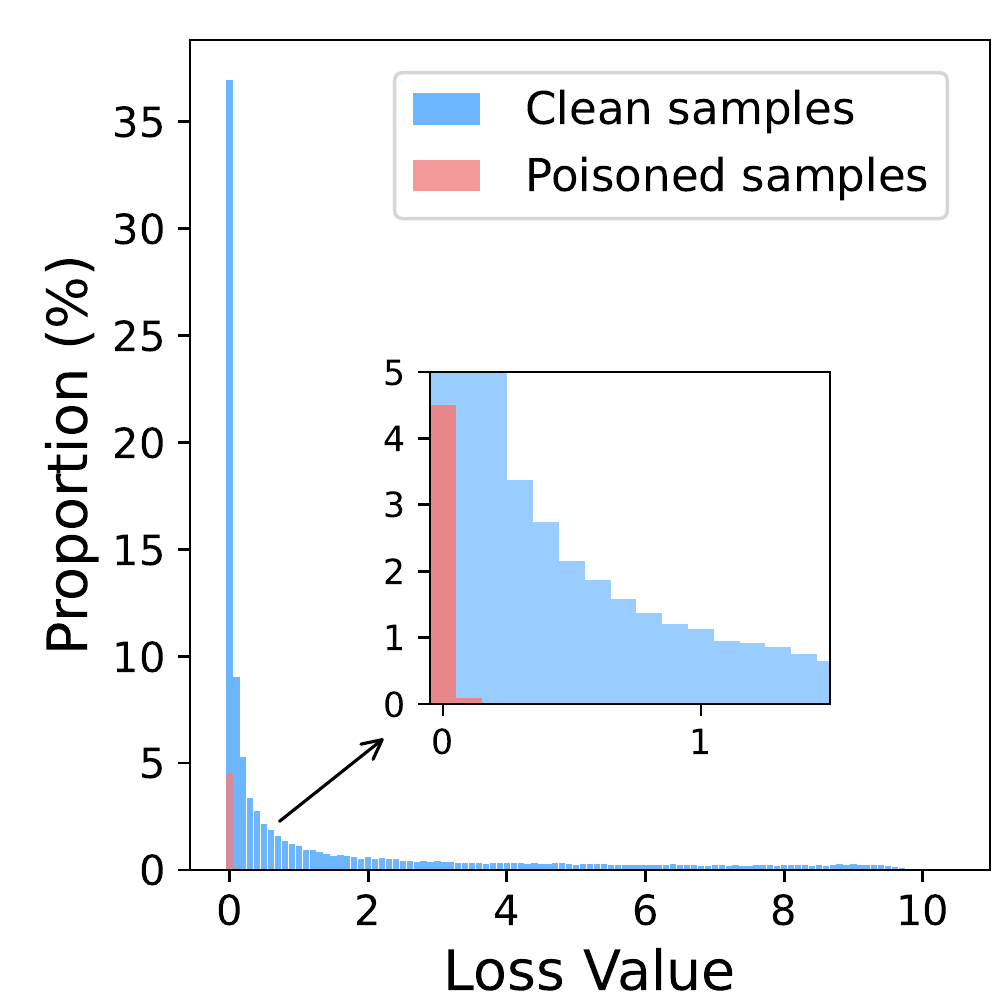}
        \caption{}
        \label{badnets_loss_epoch91_after_one_forward}
    \end{subfigure}
    \begin{subfigure}[]{0.24\linewidth}
        \includegraphics[width=\linewidth]{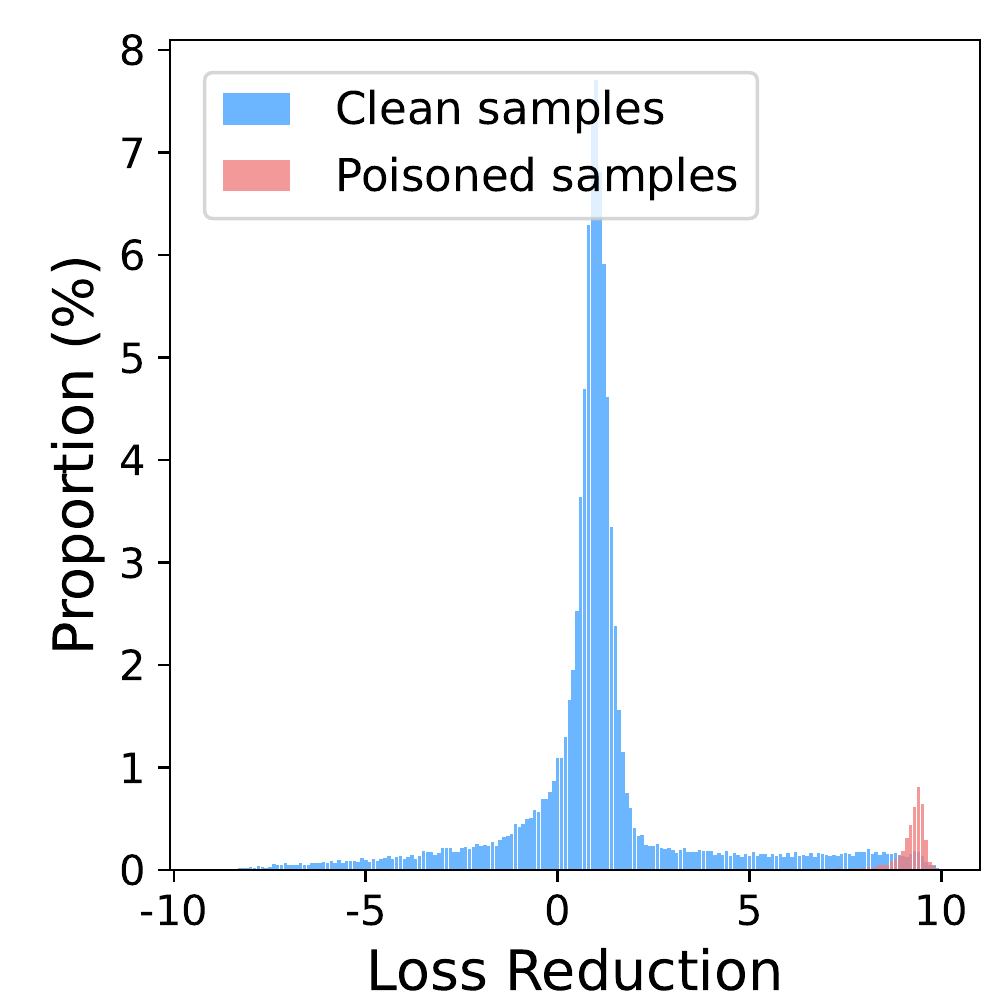}
        \caption{}
        \label{badnets_loss_reduction_epoch91}
    \end{subfigure}
    \begin{subfigure}[]{0.24\linewidth}
        \includegraphics[width=\linewidth]{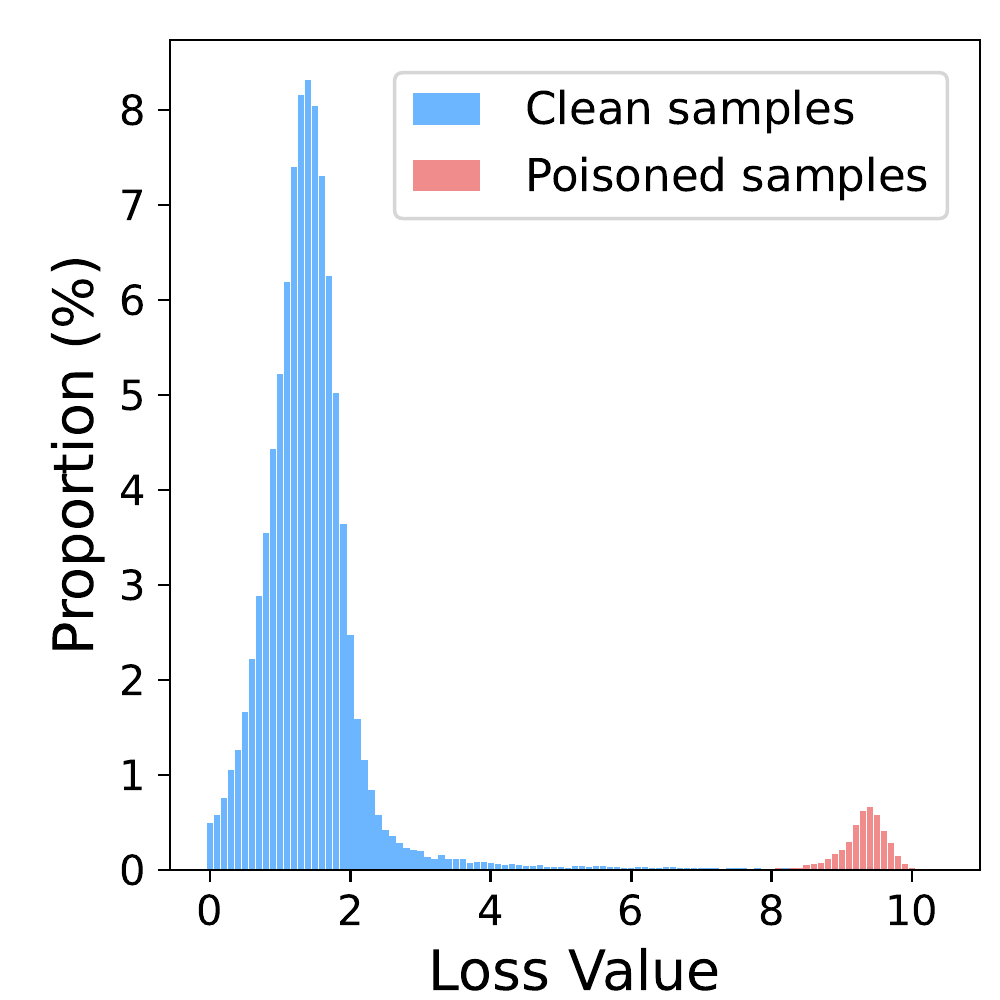}
        \caption{}
        \label{badnets_loss_epoch120}
    \end{subfigure}
	\centering
 \vspace{-0.5em}
	\caption{Loss distribution of samples on the model after different operations, which motivates us to propose the meta-split method in stage 3. (a) The model $f_{\bm{\theta}}$ after previous two stages. The clean hard samples have similar high losses to the poisoned ones, which can not be separated  by the loss-guided split. (b) The `virtual model' $f_{\bm{\theta'}}$ in Fig. \ref{badnets_loss_epoch91} after one-epoch supervised learning. The losses of poisoned samples reduce to zero, which shows the poisoned samples can be fast learned. (c) The loss reduction between $f_{\bm{\theta}}$ in Fig. \ref{badnets_loss_epoch91} and $f_{\bm{\theta'}}$ in Fig. \ref{badnets_loss_epoch91_after_one_forward}. (d) The model $f_{\bm{\theta}}$ after stage 3. Compared with Fig. \ref{badnets_loss_epoch91}, poisoned samples and clean samples are better separated, especially for clean hard samples, which benefits from our proposed meta-split.}
\end{figure*}

\section{Proposed Backdoor Defense: ASD}
In this section, we present the pipeline of our ASD. 
The key challenge is to adaptively update and maintain two pools, \textit{i.e.}, a clean data pool $\mathcal{D}_{C}$ and a polluted data pool $\mathcal{D}_{P}$. 
As shown in Fig. \ref{Pipeline}, our ASD is summarized in three stages: (1) We first initialize $\mathcal{D}_{C}$ with several fixed clean seed samples and $\mathcal{D}_{P}$ with the whole poisoned dataset $\mathcal{D}$. We perform the warming-up training and update $\mathcal{D}_{C}$ by class-aware loss-guided split. (2) Then, we adopt class-agnostic loss-guided split on the entire dataset $\mathcal{D}$ and supplement $\mathcal{D}_{C}$ to accelerate the defense process. (3) To add more clean hard samples into $\mathcal{D}_{C}$, we propose a meta-learning-inspired (meta-split) method. During these three stages, $\mathcal{D}_{P}$ is composed of all remaining samples in $\mathcal{D}$ except for those selected in $\mathcal{D}_{C}$. The model $f_{\bm{\theta}}$ is trained by Eq. \ref{eq:formulation semi-supervised 2 loss}. Algorithm of our ASD is shown in Appendix \red{A}.\par

\subsection{Warming-up and training with loss-guided split}
We introduce the first two stages of ASD. Both two stages utilize the loss-guided data split to update the clean data pool $\mathcal{D}_{C}$ and the polluted data pool $\mathcal{D}_{P}$ adaptively. To be more specific, given a model $f_{\bm{\theta}}$ at any epoch during stage 1 and stage 2, $\mathcal{D}_{C}$ is supplemented by samples with the lowest $\mathcal{L}_1(f_{\bm{\theta}})$ losses, while $\mathcal{D}_{P}$ is composed of remaining samples. As suggested in \cite{huang2022backdoor}, symmetric cross-entropy (SCE) \cite{wang2019symmetric} loss is adopted as $\mathcal{L}_1(\cdot)$ because it can amplify the difference between clean samples and poisoned samples when compared with cross-entropy (CE) loss. 
\par

\noindent \textbf{Warming up with class-aware loss-guided split.} In the warming-up stage, we first initial $\mathcal{D}_{C}$ with the clean seed samples and $\mathcal{D}_{P}$ with all the poisoned training data. Since only a few clean seed samples are available in $\mathcal{D}_{C}$, we progressively increase the number of samples in $\mathcal{D}_{C}$, namely we add $n$ every $t$ epochs in each class. Next, we add samples with the lowest $\mathcal{L}_1(\cdot)$ losses \textit{in each class} to $\mathcal{D}_{C}$ dynamically, and remaining samples are used as $\mathcal{D}_{P}$. The reason for the class-aware way is to prevent the performance collapse caused by the class imbalance in the small clean data pool. Based on the tiny but progressively growing $\mathcal{D}_{C}$ and its complement $\mathcal{D}_{P}$, the model will be warmed up according to Eq. \ref{eq:formulation semi-supervised 2 loss} during the first $T_1$ epochs. \par

\noindent \textbf{Training with class-agnostic loss-guided split.}  
After the first stage, the model with certain accuracy can be used to better distinguish clean samples and poisoned samples. In the second stage, we further enlarge $|\mathcal{D}_{C}|$ to accelerate the defense process. We directly add $\alpha\%$ samples with the lowest $\mathcal{L}_1(\cdot)$ losses in the entire dataset into $\mathcal{D}_{C}$, and remaining samples are used as $\mathcal{D}_{P}$. Such a class-agnostic loss-guided data split method can avoid selecting poisoned samples into $\mathcal{D}_{C}$ from the target class and further suppress the attack success rate. With two split data pools, we adopt Eq. \ref{eq:formulation semi-supervised 2 loss} to train the model from epoch $T_1$ to epoch $T_2$. 

\setlength{\columnsep}{9pt}
\begin{wrapfigure}{r}{0.4\linewidth}
  \vspace{-2em}
  \begin{center}
    \includegraphics[width=\linewidth]{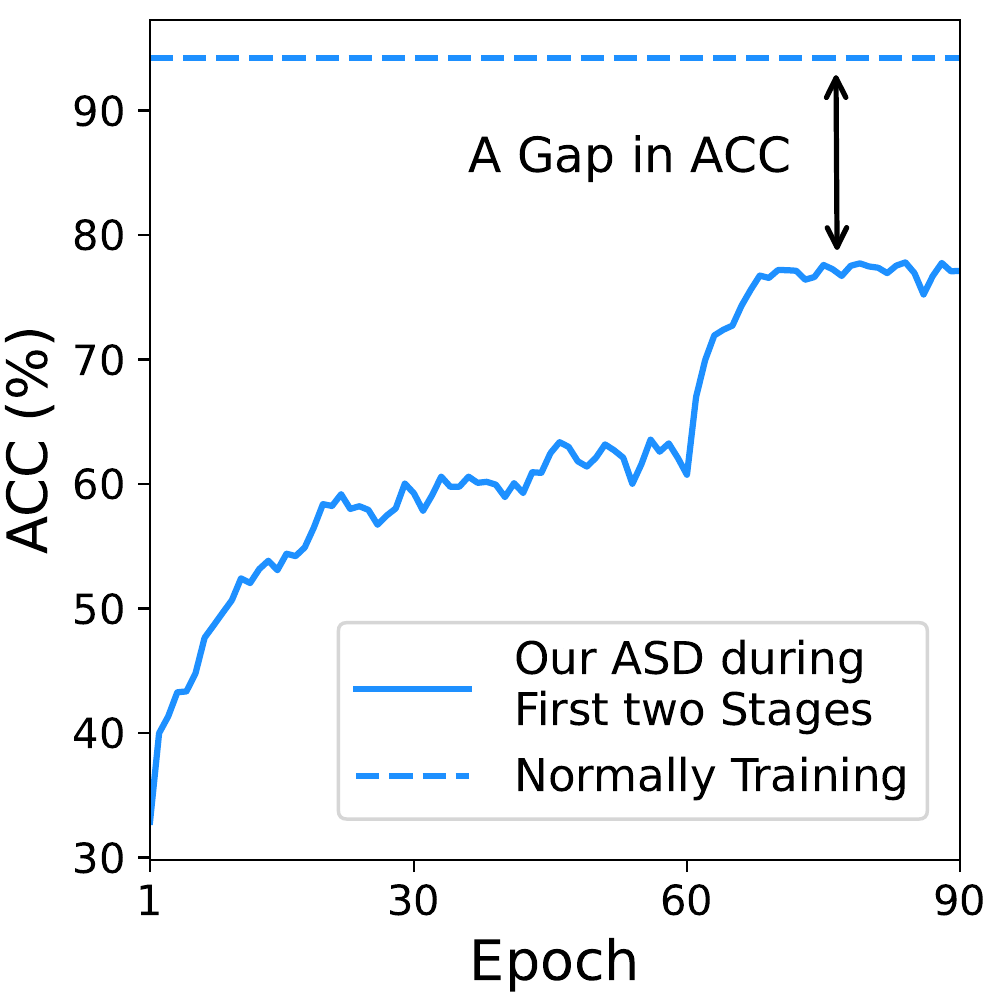}
  \end{center}
  \setlength{\abovecaptionskip}{-10pt}
  \caption{The clean accuracy of our ASD defended model during previous two stages and normally trained model.}
  \vspace{-0.8em}
  \label{fig:a gap in acc}
\end{wrapfigure}

\subsection{Hard sample training with meta-split}
After previous two stages, there is still a small gap in clean accuracy between our ASD defended model and the normally trained model (in Fig. \ref{fig:a gap in acc}). We assume that the gap is caused by that some clean samples can not be accessed by the loss-guided split. To verify our assumption, we calculate the losses of all training data for BadNets attack \cite{gu2017badnets}. In Fig. \ref{badnets_loss_epoch91}, though the losses of most clean samples are nearly zeros, there are still some clean samples with high losses, which are similar to poisoned ones. Such phenomena indicate that most clean data have been well learned while poisoned data are not by our model after previous two stages. However, there are still some \textit{(model-dependent) clean hard samples} \footnote{Here `model-dependent' refers to model checkpoint-dependent , since clean hard samples are samples with the largest losses on the current model at each epoch.}, which are difficult to be separated from poisoned ones just by loss magnitudes. We argue such clean hard samples are quite difficult to learn by a model yet poisoned samples can be easy to learn during supervised learning. Fig. \ref{badnets_loss_epoch91_after_one_forward} and Fig. \ref{badnets_loss_reduction_epoch91} can verify our claim and show that losses of poisoned samples drop much faster than those of clean hard samples after one-epoch supervised learning, which inspires us to provide a new solution to successfully separate clean hard samples and poisoned ones. 

To distinguish clean hard samples from poisoned ones and inspired by meta-learning \cite{hospedales2021meta,finn2017model}, we propose a novel manner to supplement $\mathcal{D}_{C}$, called meta-split. Given a model $f_{\bm{\theta}}$ at any epoch in the third stage, we first create a new `\textit{virtual model}' $f_{\bm{\theta}'}$ with the same parameters and architecture as $f_{\bm{\theta}}$. The virtual model $f_{\bm{\theta}'}$ is updated on the entire poisoned dataset $\mathcal{D}$ by the loss $\mathcal{L}_2(\cdot)$ with learning rate $\beta$ which can be denoted as:
\begin{equation}
\begin{aligned}
\bm{\theta}' & \gets \bm{\theta}, \\
\bm{\theta}' \gets \bm{\theta}' - \beta & \nabla_{\bm{\theta}'} \mathcal{L}_2(f_{\bm{\theta}'}(\bm{x}),y),
\end{aligned}
\label{eq:meta-split}
\end{equation}
where cross-entropy (CE) loss is adopted as $\mathcal{L}_2(\cdot)$. Finally, $\gamma\%$ samples with the least loss reduction $\mathcal{L}_1(f_{\bm{\theta}})-\mathcal{L}_1(f_{\bm{\theta'}})$ are chosen to supplement $\mathcal{D}_{C}$.

Note that the virtual model $f_{\bm{\theta}'}$ is only used for sample separation and is not involved in the followed training process. After supplementing $\mathcal{D}_{C}$ by meta-split and consistently updating the model $f_{\bm{\theta}}$ using Eq. \ref{eq:formulation semi-supervised 2 loss} until training end at epoch $T_3$, $f_{\bm{\theta}}$ can successfully split more clean hard samples in $\mathcal{D}_{C}$ and thus obtain higher accuracy. The loss distribution of samples on $f_{\bm{\theta}}$ at $T_3$ in Fig. \ref{badnets_loss_epoch120} shows clean samples and poisoned samples have been successfully separated, which further demonstrates the effectiveness of meta-split during the third stage of our ASD.\par

\begin{table*}[t]
\caption{The clean accuracy (ACC \%) and the attack success rate (ASR \%) of five backdoor defenses against six backdoor attacks across three datasets, including CIFAR-10, GTSRB and ImageNet. Note that the results of FP, NAD and ABL are reported by grid-searching the best ones in different hyper-parameters. In contrast, the results of DBD and our ASD are provided in the same hyper-parameters. Best results among five defenses are highlighted in \textbf{bold}. } 
\label{main results}
\footnotesize
\setlength{\tabcolsep}{3mm}
{
\begin{tabular}{l|l|ll|ll|ll|ll|ll|ll}

\toprule[0.68pt]
\addlinespace[0pt]
\multicolumn{1}{c|}{\multirow{2}{*}{Dataset}} & \multicolumn{1}{c|}{\multirow{2}{*}{Attack}} & \multicolumn{2}{c|}{No Defense}                    & \multicolumn{2}{c|}{FP }                            & \multicolumn{2}{c|}{NAD }                                                    & \multicolumn{2}{c|}{ABL }                           & \multicolumn{2}{c|}{DBD }                           & \multicolumn{2}{c}{ASD (Ours)}      \\ \cline{3-14} 
\multicolumn{1}{c|}{}       & \multicolumn{1}{c|}{}                   & \multicolumn{1}{c}{ACC} & \multicolumn{1}{c|}{ASR} & \multicolumn{1}{c}{ACC} & \multicolumn{1}{c|}{ASR} & \multicolumn{1}{c}{ACC} & \multicolumn{1}{c|}{ASR} & \multicolumn{1}{c}{ACC} & \multicolumn{1}{c|}{ASR} & \multicolumn{1}{c}{ACC} & \multicolumn{1}{c|}{ASR} & \multicolumn{1}{c}{ACC} & \multicolumn{1}{c}{ASR}  \\ \hline
\multicolumn{1}{c|}{\multirow{7}{*}{CIFAR-10}}  & \multicolumn{1}{l|}{BadNets }                 & 94.9 & 100                              & \textbf{93.9} & 1.8                              & 88.2 & 4.6                                                           &  93.8 & 1.1                              & 92.3 & \textbf{0.8}                              & 93.4 & 1.2         \\
\multicolumn{1}{c|}{}       & Blend                                       & 94.1 & 98.3                             & 92.9 & 77.1                              & 85.8 & 3.4                                                            & 91.9 & 1.6                               & 91.7 & \textbf{0.7}                              & \textbf{93.7} & 1.6         \\
\multicolumn{1}{c|}{}       & WaNet                                        & 93.6 & 99.9                               &  90.4 & 98.6                               & 71.3 & 6.7                                                            &  84.1 & 2.2                               & 91.4 & \textbf{0}                              & \textbf{93.1} & 1.7          \\
\multicolumn{1}{c|}{}       & IAB                                         & 94.2 & 100                              &  89.3 & 98.1                             & 82.8 & 4.2                                                           &  \textbf{93.4} & 5.1                               & 91.6 & 100                              & 93.2 & \textbf{1.3}      \\
\multicolumn{1}{c|}{}       & Refool                                          & 93.8 & 98.2                              & 92.1 & 86.1                              & 86.2 & 3.6                                                            &  82.7 & 1.3                              & 91.5 & 0.5                              & \textbf{93.5} & \textbf{0}         \\
\multicolumn{1}{c|}{}       & CLB                                          & 94.4 & 99.9                             & 90.2 & 92.8                             & 86.4 & 9.5                                                           &  86.6 & 1.3                              & 90.6 & \textbf{0.1}                              & \textbf{93.1} & 0.9         \\
\cline{2-14} 
\multicolumn{1}{c|}{}       & Average                                      &    94.2                     &       99.4                   &              91.5           &     75.8                     &       83.5                  &            5.3              &            88.7             &         2.1                                                                    &                91.5         &   17.0                       &      \textbf{93.3}                  &   \textbf{1.1}  \\ \hline

\multicolumn{1}{c|}{\multirow{7}{*}{GTSRB}}  & \multicolumn{1}{c|}{BadNets}                & 97.6 & 100                           & 84.2 & \textbf{0}                              & \textbf{97.1} & 0.2                                                           & \textbf{97.1} & \textbf{0}                               & 91.4 & \textbf{0}                              & 96.7 & \textbf{0}          \\
\multicolumn{1}{c|}{}       & Blend                                        & 97.2 & 99.4                             & 91.4 & 68.1                              & 93.3 & 62.4                                                            &  \textbf{97.1} & 0.5                               & 91.5 & 99.9                             & \textbf{97.1} & \textbf{0.3}         \\
\multicolumn{1}{c|}{}       & WaNet                                        & 97.2 & 100                              & 92.5 & 21.4                             & 96.5 & 47.1                                                           &  97.0 & 0.4                               & 89.6 & \textbf{0}                             & \textbf{97.2} & 0.3         \\
\multicolumn{1}{c|}{}       & IAB                                          & 97.3 & 100                              & 86.9 & \textbf{0}                              & 97.1 & 0.1                                                            &  \textbf{97.4} & 0.6                              & 90.9 & 100                              & 96.9 & \textbf{0}          \\
\multicolumn{1}{c|}{}       & Refool                                          & 97.5 & 99.8                              & 91.5 & 0.2                              & 95.5 & 1.4                                                          &  96.2 & \textbf{0}                              & 91.4 & 0.4                             & \textbf{96.8} & \textbf{0}          \\ 
\multicolumn{1}{c|}{}       & CLB                                          & 97.3 & 100                              & 93.6 & 99.3                              & 3.3 & 21.1                                                          &  90.4 & 2.3                               & 89.7 & 0.3                              & \textbf{97.3} & \textbf{0}          \\
\cline{2-14} 
\multicolumn{1}{c|}{}       & Average                                      &    97.4                     &         99.9                 &   90.0                      &          31.5                &              80.5           &      22.1                    &           95.9              &      0.6                                                                      &    90.8                     &  33.4                        &            \textbf{97.0}             &  \textbf{0.1}   \\ \hline

\multicolumn{1}{c|}{\multirow{7}{*}{ImageNet}}  & \multicolumn{1}{c|}{BadNets}                 & 79.5 & 99.8                              & 70.3 & 1.6                              & 65.1 & 5.1                                                            &  83.1 & \textbf{0}                              & 81.9 & 0.3                             & \textbf{83.3} & 0.1          \\
\multicolumn{1}{c|}{}       & Blend                                        & 82.5 & 99.5                              & 63.4 & 9.5                             & 64.8 & 0.3                                                           &  \textbf{82.6} & 0.7                              & 82.3 & 100                              & 82.5 & \textbf{0.2}          \\
\multicolumn{1}{c|}{}       & WaNet                                        & 79.1 & 98.9                              & 58.2 & 84.4                              & 63.8 & 1.3                                                         &  74.9 & 1.1                               & 80.6 & 9.8                             & \textbf{84.1} & \textbf{0.8}          \\
\multicolumn{1}{c|}{}       & IAB                                          & 78.2 & 99.6                              & 58.7 & 84.2                              & 63.8 & 0.6                                                           &  81.7 & \textbf{0}                               & \textbf{83.1} & \textbf{0}                             & 81.6 & 0.5          \\
\multicolumn{1}{c|}{}       & Refool                                          & 80.6 & 99.9                              & 61.4 & 10.3                             & 63.7 & 0.3                                                           &  76.2 & 0.2                               & 82.5 & 0.1                             & \textbf{82.6} & \textbf{0}          \\ 
\multicolumn{1}{c|}{}       & CLB                                          & 80.1 & 42.8                              & 73.2 & 38.3                              & 62.7 & 1.7                                                           &  \textbf{82.8} & 0.8                               & 81.8 & \textbf{0}                              & 82.2 & \textbf{0}          \\ 
\cline{2-14} 
\multicolumn{1}{c|}{}       & Average                                      &       80.0                  &             90.1             &      64.2                      &              38.1         &           64.0              &      1.5                    &            80.2            &       0.5                                                                      &   82.0                      &                18.4          &              \textbf{82.7}           &   \textbf{0.3}  \\ 
\addlinespace[-0.22em]
\bottomrule[0.68pt]

\end{tabular}}
\end{table*}

\begin{table}[]
\vspace{-1em}
\caption{The average training time (s) of ABL, DBD and our ASD.
}
\label{time cost}
\footnotesize
\setlength{\tabcolsep}{1.9mm}{
\begin{tabular}{clllll}
\toprule[0.68pt]
\addlinespace[0pt]
Dataset & Method & Stage 1 & Stage 2 & Stage 3 & Total  \\ \hline
\multirow{3}{*}{CIFAR-10}   & ABL  &  740   &    2,450 &   10  &  3,200  \\ \cline{2-6} 
& DBD  &   30,000  &  80   &   15,770  &  45,850 \\ \cline{2-6} 
& ASD (Ours)  &  4,980   &  2,490   &  2,520  &  9,990  \\ \hline
\multirow{3}{*}{GTSRB}     & ABL  &  520 &  1,750   &   5   &  2,275     \\ \cline{2-6} 
& DBD  &  23,000   &  57   &   15,580  &  38,637 \\ \cline{2-6} 
& ASD (Ours)  &  4,920  &  2,460   &   830   &   8,210    \\ \hline
\multirow{3}{*}{ImageNet}     & ABL  &  900  &   2,940 &  15    &   3,855     \\ \cline{2-6} 
& DBD  &  180,000   &  350   &   42,750  &  223,100 \\ \cline{2-6} 
& ASD (Ours)  &  13,500  &  6,750   &   6,780   &  27,030     \\ 
\addlinespace[-0.22em]
\bottomrule[0.68pt]
\end{tabular}}
\end{table}

\section{Experiments}
\label{sec:experiments}

\subsection{Experimental Setups}
\noindent \textbf{Datasets and DNN models.} We adopt three benchmark datasets to evaluate all the backdoor defenses, \textit{i.e.}, CIFAR-10 \cite{deng2009imagenet}, GTSRB \cite{stallkamp2011german} and an ImageNet \cite{deng2009imagenet} subset. ResNet-18 \cite{he2016deep} is set as the default model in our experiments. More details are listed in Appendix \red{B.1}. Moreover, we also provide the results on VGGFace2 dataset \cite{cao2018vggface2} and DenseNet-121 \cite{huang2017densely} in Appendix \red{C}. \par

\noindent \textbf{Attack baselines and setups.} We conduct six state-of-the-art backdoor attacks, including BadNets \cite{gu2017badnets}, Blend backdoor attack (Blend) \cite{chen2017targeted}, Warping-based backdoor attack (WaNet) \cite{nguyen2021wanet}, Input-aware backdoor attack (IAB) \cite{nguyen2020input}, Reflection-based attack (Refool) \cite{liu2020reflection} and clean-label attack with adversarial perturbations (dubbed ‘CLB’) \cite{turner2018clean}. All these backdoor attacks are implemented as suggested in \cite{huang2022backdoor} and their original papers.  
Following \cite{huang2022backdoor}, we choose 3 ($y_t=3$) as the target label and set the poisoned rate as 25\% poisoned samples in the target class for the clean-label attack and 5\% for other five backdoor attacks. More details about the attack setups are summarized in Appendix \red{B.2}.
Besides, we also provide the results against the sample-specific attack \cite{li2021invisible} and \textit{all2all} attack in Appendix \red{D}.
\par

\noindent \textbf{Defense baselines and setups.} We compare our proposed method with four existing backdoor defenses, including Fine-pruning (FP) \cite{liu2018fine}, Neural Attention Distillation (NAD) \cite{li2021neural}, Anti-Backdoor Learning (ABL) \cite{li2021anti} and decoupling-based backdoor defense (DBD) \cite{huang2022backdoor}. 
Since FP, NAD and ABL are sensitive to their hyper-parameters, we report their best results optimized by grid-search (See Appendix \red{N}). DBD is implemented based on the original paper \cite{huang2022backdoor}. Besides, FP and NAD are assumed to have access to 5\% of the clean training data.
Furthermore, we also provide the results of cutmix-based defense \cite{borgnia2021strong} and differential privacy SGD-based defense \cite{du2019robust} in Appendix \red{E}.
\par
For our ASD, we follow DBD \cite{huang2022backdoor} to adopt MixMatch \cite{berthelot2019mixmatch} as our default semi-supervised method. The initial number of clean seed samples $w$ is $10$ in each class and it will increase $n=10$ at every $t=5$ epochs. After stage 1, the filtering rate $\alpha\%$ is set to $50\%$. For meta-split, we choose Adam \cite{kingma2014adam} optimizer
with the learning rate $\beta=0.015$ 
to perform one-epoch supervised learning on a virtual model. In particular, we only update the parameters of the last three layers. The discussions about the hyper-parameters for meta-split are demonstrated in Appendix \red{I}. On CIFAR-10 and ImageNet, $T_1$, $T_2$ and $T_3$ are chosen as 60, 90 and 120, and $T_3$ is chosen as 100 on GTSRB. More details about the defense setups are in Appendix \red{B.3}.\par

\noindent \textbf{Evaluation metrics.} We evaluate 
backdoor defenses by two widely used metrics, \textit{i.e.}, the accuracy on clean dataset (ACC) and the attack success rate (ASR). Specifically, ASR is the fraction of the samples from the non-target class with the trigger classified to the target label by the backdoored model. For a backdoor defense, the higher ACC and the lower ASR correspond to better performance.\par

\subsection{Main Results}
To verify the superiority of our backdoor defense, we summarize the ACC and ASR of five backdoor defenses against six backdoor attacks on three datasets in Table \ref{main results}. We also report the time cost of three training-time defenses in Table \ref{time cost}. Table \ref{main results} shows that our ASD can achieve low ASRs meanwhile maintain high ACCs on three datasets. In particular, ASD can purify and better utilize poisoned samples, even outperforming `No Defense' on ImageNet on average.
In comparison, although FP and NAD require a larger amount of local clean samples (2,500) than ours (100), FP provides a limited reduction on ASR and NAD damages the ACC of the models by a large margin, which constrains their deployment in repairing the model. 
Table \ref{time cost} shows that ABL requires the least training time among three training-time defenses but needs grid-search for hyper-parameters to defend against different attacks. Besides, when ABL encounters the WaNet or Refool, its static backdoor isolation samples can be mixed with some clean samples. Once ABL adopts these backdoor isolation samples to unlearn for the backdoored model, the ACC will drop. In contrast, our adaptive split can update two data pools dynamically, which stabilizes the defense process of our ASD.

\begin{figure}[t]
    \vspace{-1em}
    \centering
    \begin{subfigure}[]{0.48\columnwidth}
        \includegraphics[width=\columnwidth]{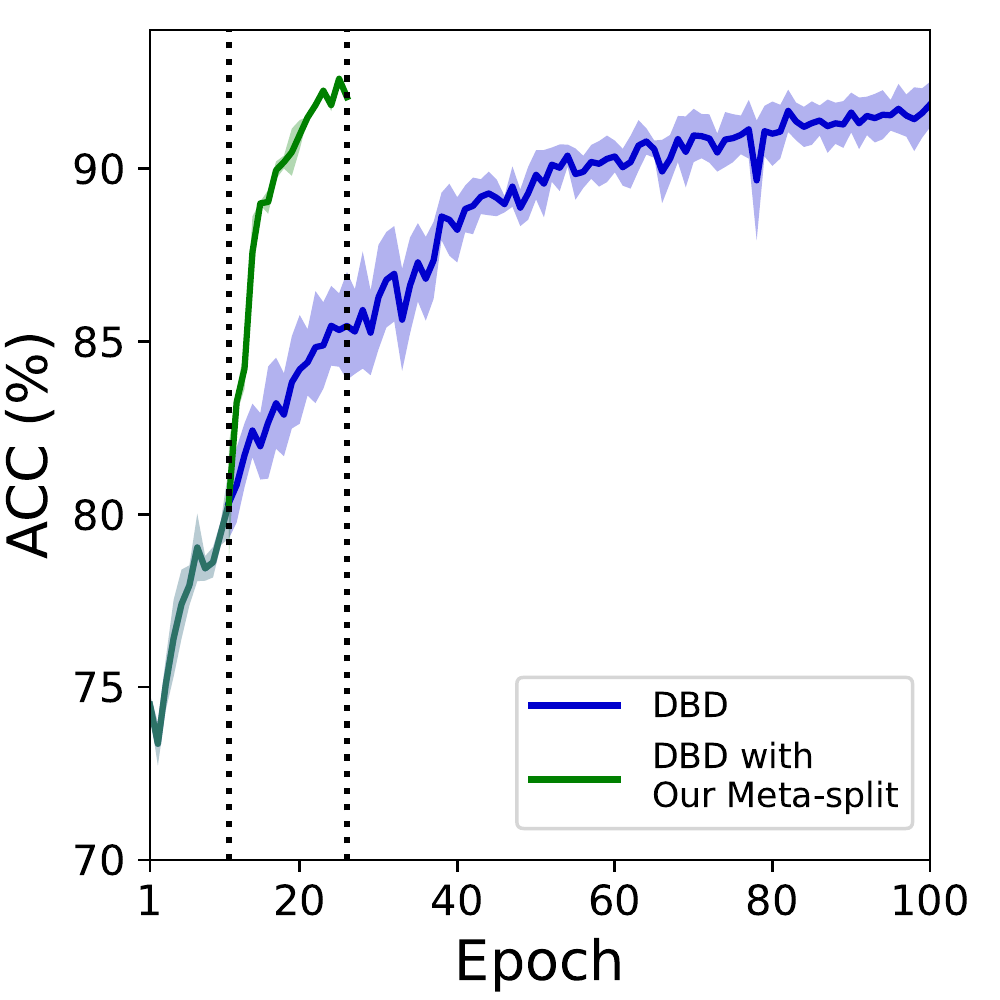}
        \caption{}
        \label{BadNets_with_meta_split}
    \end{subfigure}
    \begin{subfigure}[]{0.48\columnwidth}
        \includegraphics[width=\columnwidth]{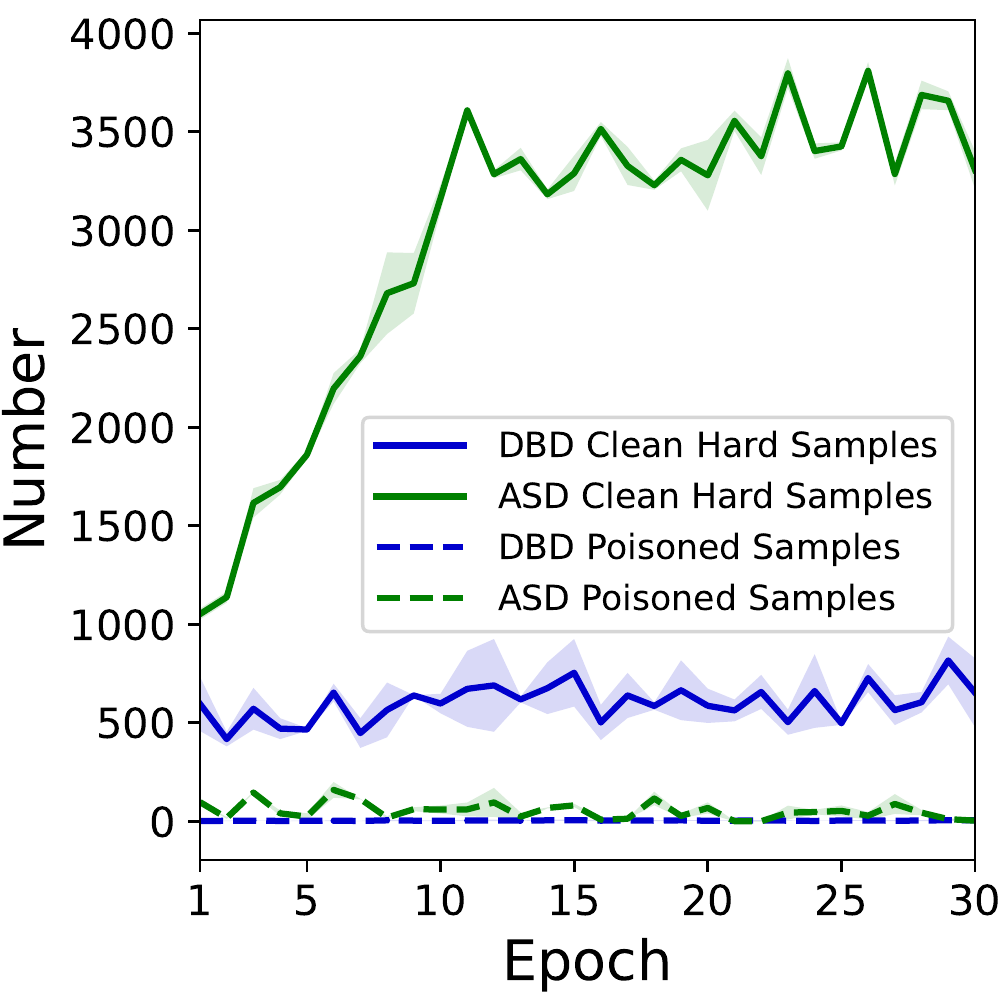}
        \caption{}
        \label{BadNets_DBD_ASD}
    \end{subfigure}
	\centering
	\caption{Combination and comparison between DBD and our ASD. The experiments ($\pm$std over 5 random runs) are conducted on CIFAR-10 for BadNets. (a) Combine DBD and our meta-split to converge faster. (b) The number of clean hard samples and poisoned samples in the clean data pool $\mathcal{D}_{C}$ during the final 30 epochs. Our ASD can successfully select clean hard examples. }
    \label{DBD with our meta-split}
\end{figure}

\begin{figure*}[t]
	\centering
\begin{minipage}[]{0.75\textwidth}
\centering
\includegraphics[width=\linewidth]{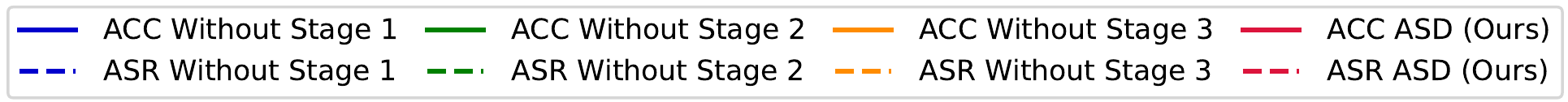}
\end{minipage}
\begin{minipage}[]{\textwidth}
    \begin{subfigure}[]{0.24\linewidth}
        \includegraphics[width=\linewidth]{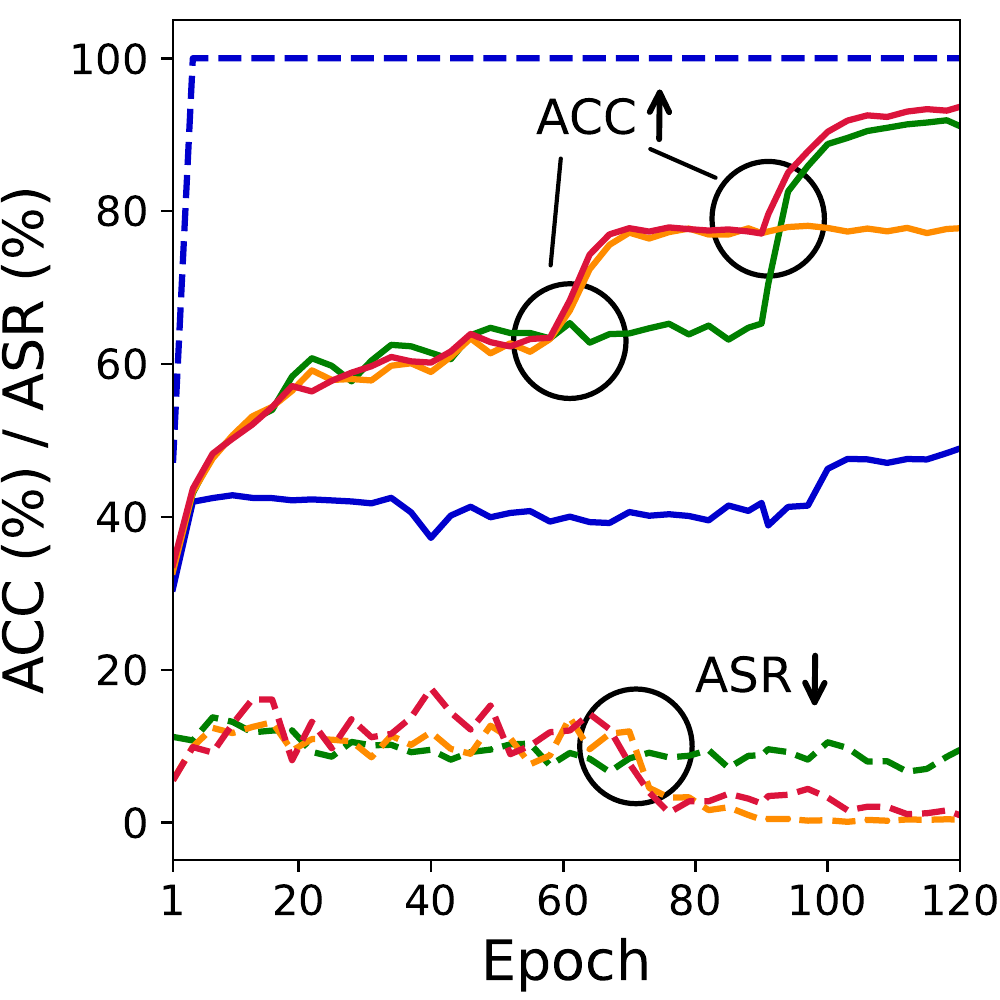}
        \caption{BadNets}
        \label{without-BadNets}
    \end{subfigure}
    \begin{subfigure}[]{0.24\linewidth}
        \includegraphics[width=\linewidth]{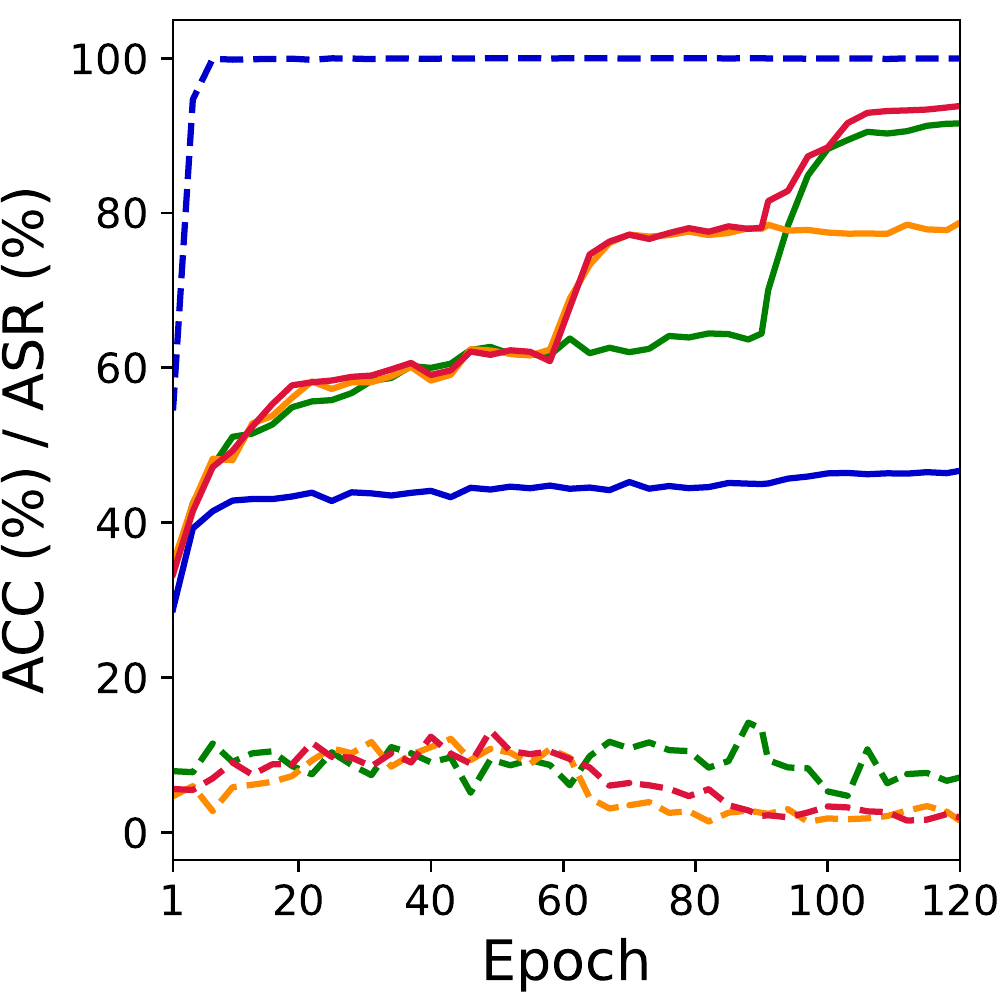}
        \caption{Blend}
        \label{without-Blend}
    \end{subfigure}
    \begin{subfigure}[]{0.24\linewidth}
        \includegraphics[width=\linewidth]{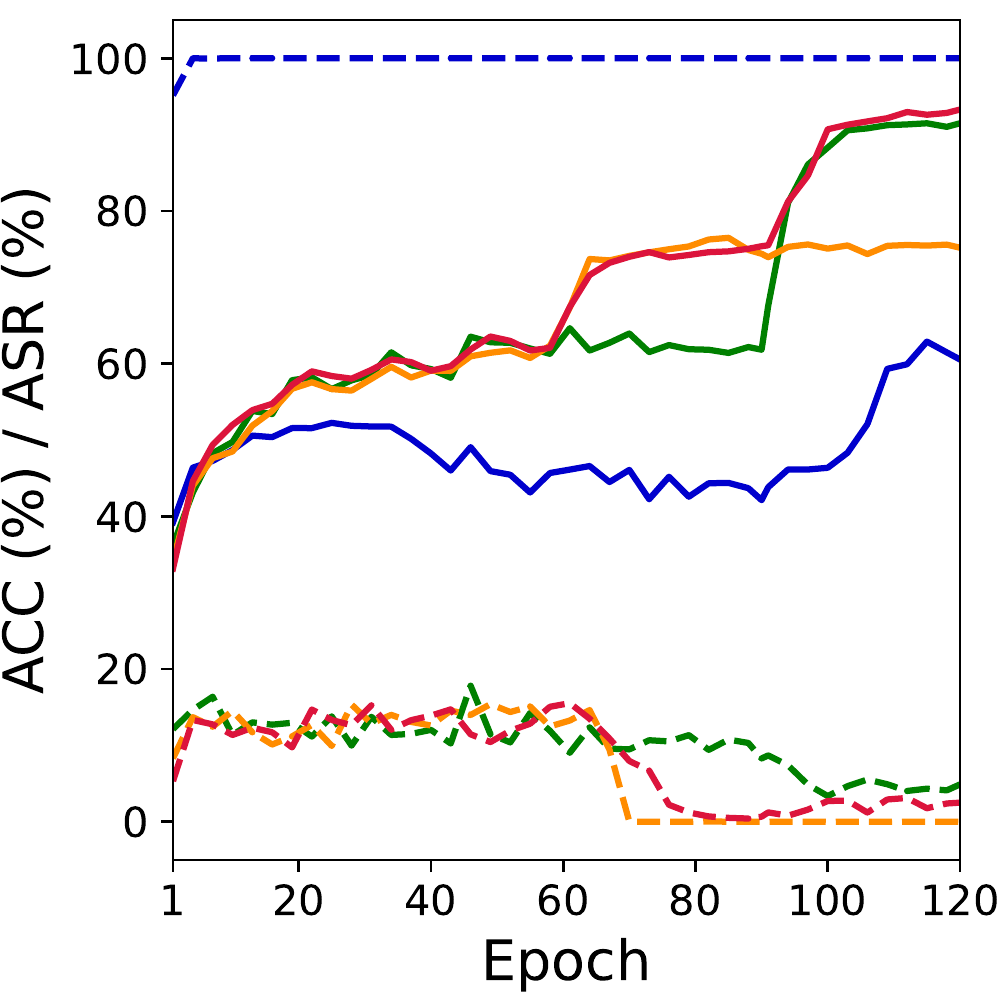}
        \caption{WaNet}
        \label{without-wanet}
    \end{subfigure}
    \begin{subfigure}[]{0.24\linewidth}
        \includegraphics[width=\linewidth]{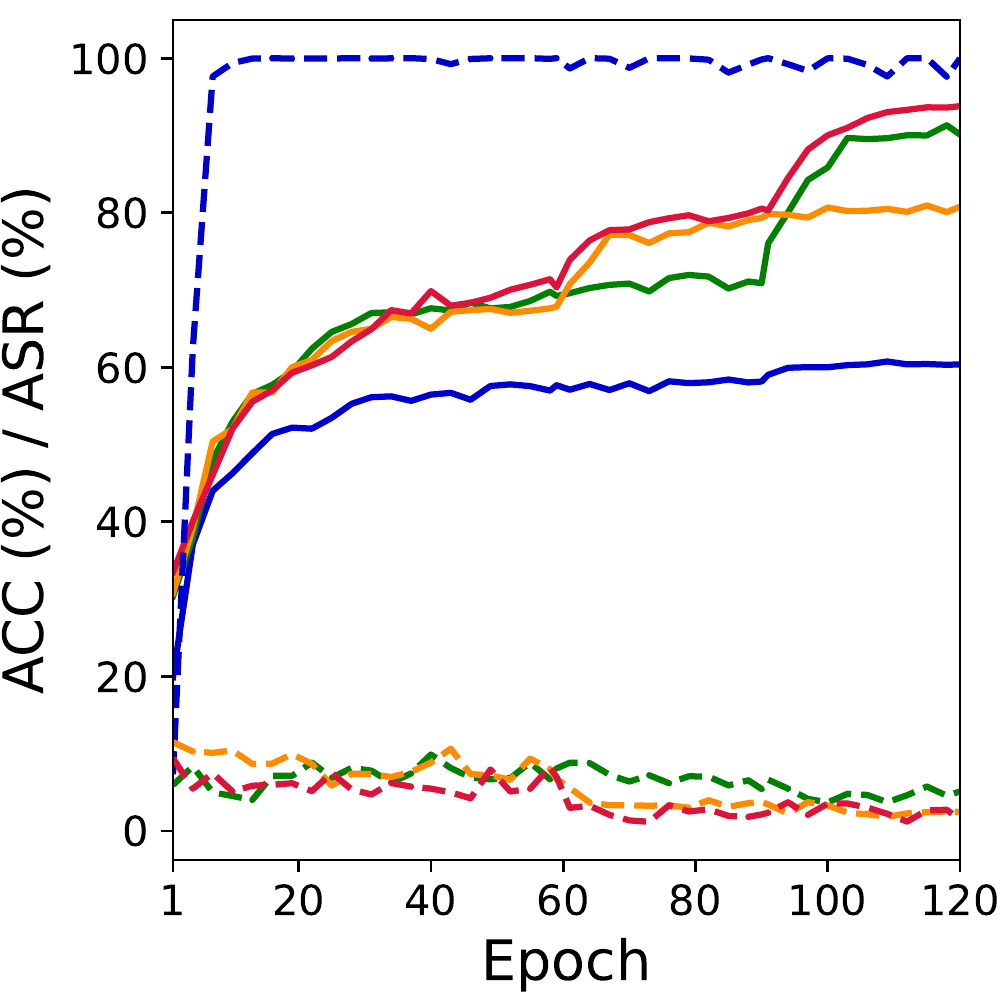}
        \caption{CLB}
        \label{without-clb}
    \end{subfigure}
	\centering
  \end{minipage}
	\caption{The clean accuracy (ACC \%) and the attack success rate (ASR \%) of different ablation studies for three stages in ASD on CIFAR-10 for four backdoor attacks, which shows the necessity of each stage in our ASD. 
    }
    \label{different without}
    \vspace{-0.8em}
\end{figure*}

\begin{figure*}[t]
	\centering
\begin{minipage}[]{0.75\textwidth}
\centering
\includegraphics[width=\linewidth]{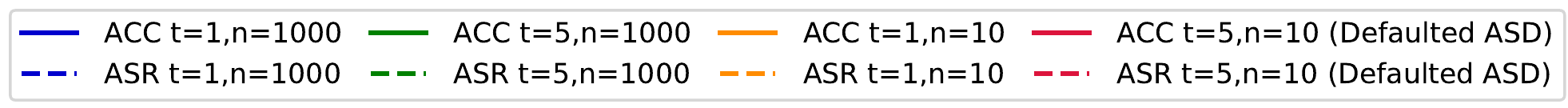}
\end{minipage}
\begin{minipage}[]{\textwidth}
    \begin{subfigure}[]{0.24\linewidth}
        \includegraphics[width=\linewidth]{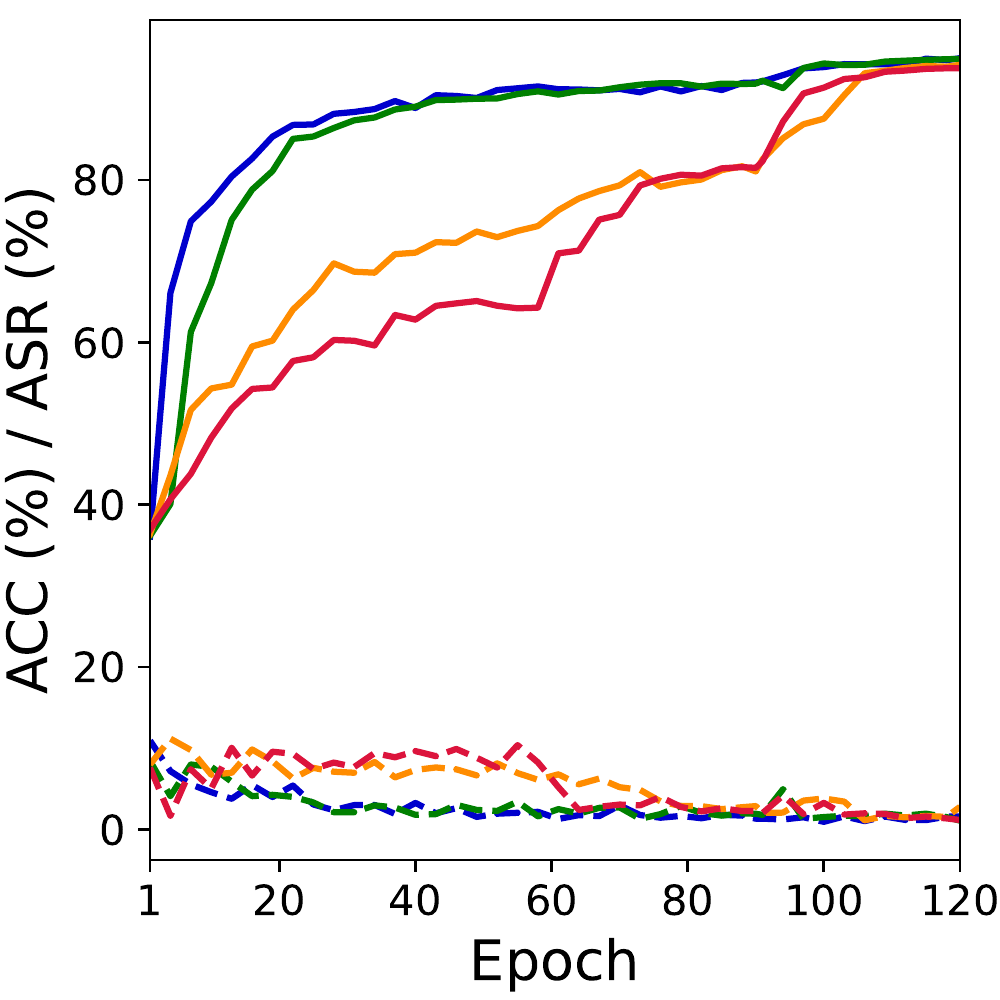}
        \caption{BadNets}
        \label{warm-BadNets}
    \end{subfigure}
    \begin{subfigure}[]{0.24\linewidth}
        \includegraphics[width=\linewidth]{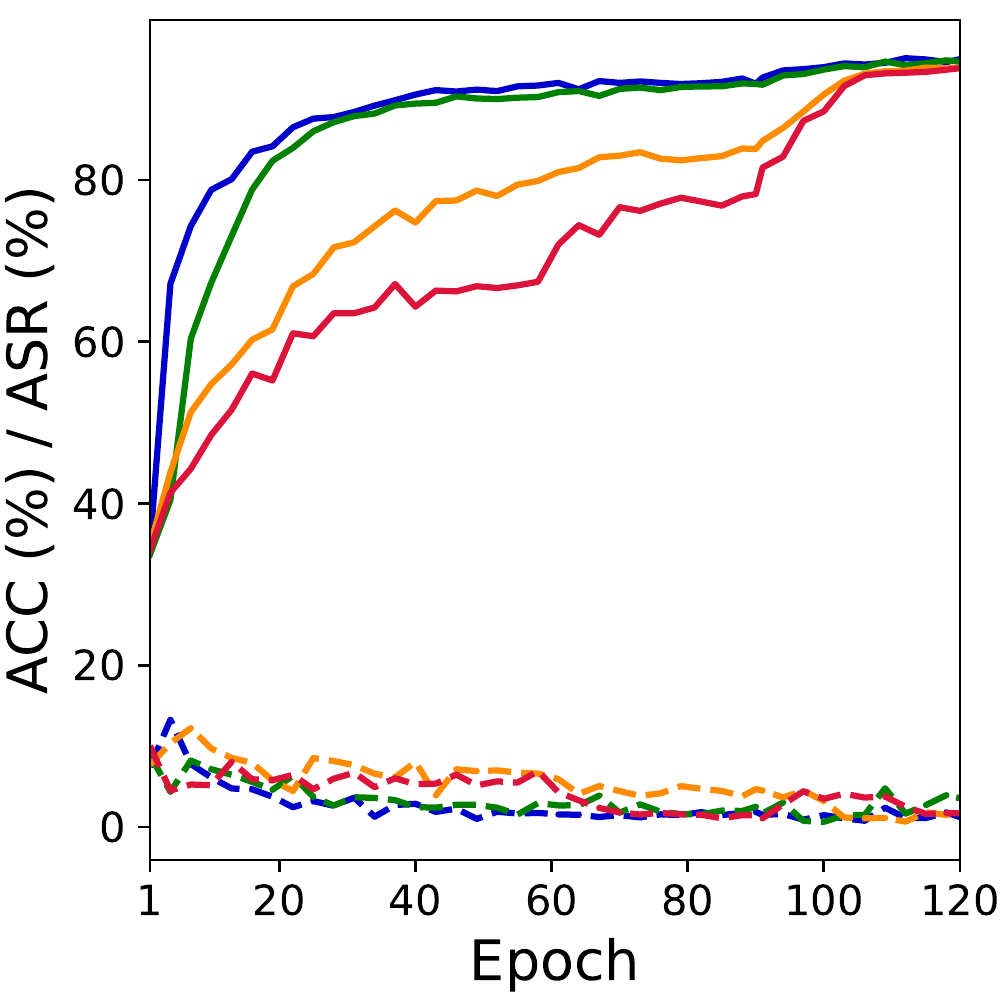}
        \caption{Blend}
        \label{warm-Blend}
    \end{subfigure}
    \begin{subfigure}[]{0.24\linewidth}
        \includegraphics[width=\linewidth]{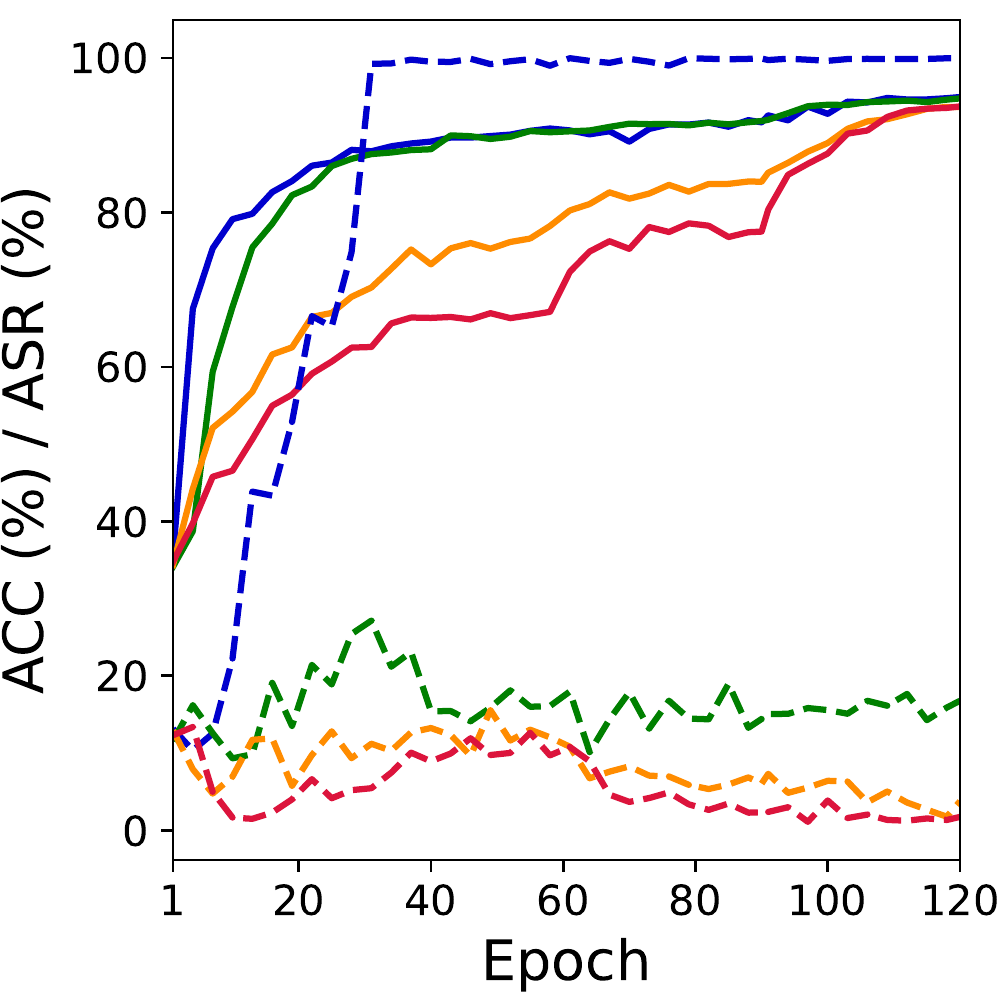}
        \caption{WaNet}
        \label{warm-wanet}
    \end{subfigure}
    \begin{subfigure}[]{0.24\linewidth}
        \includegraphics[width=\linewidth]{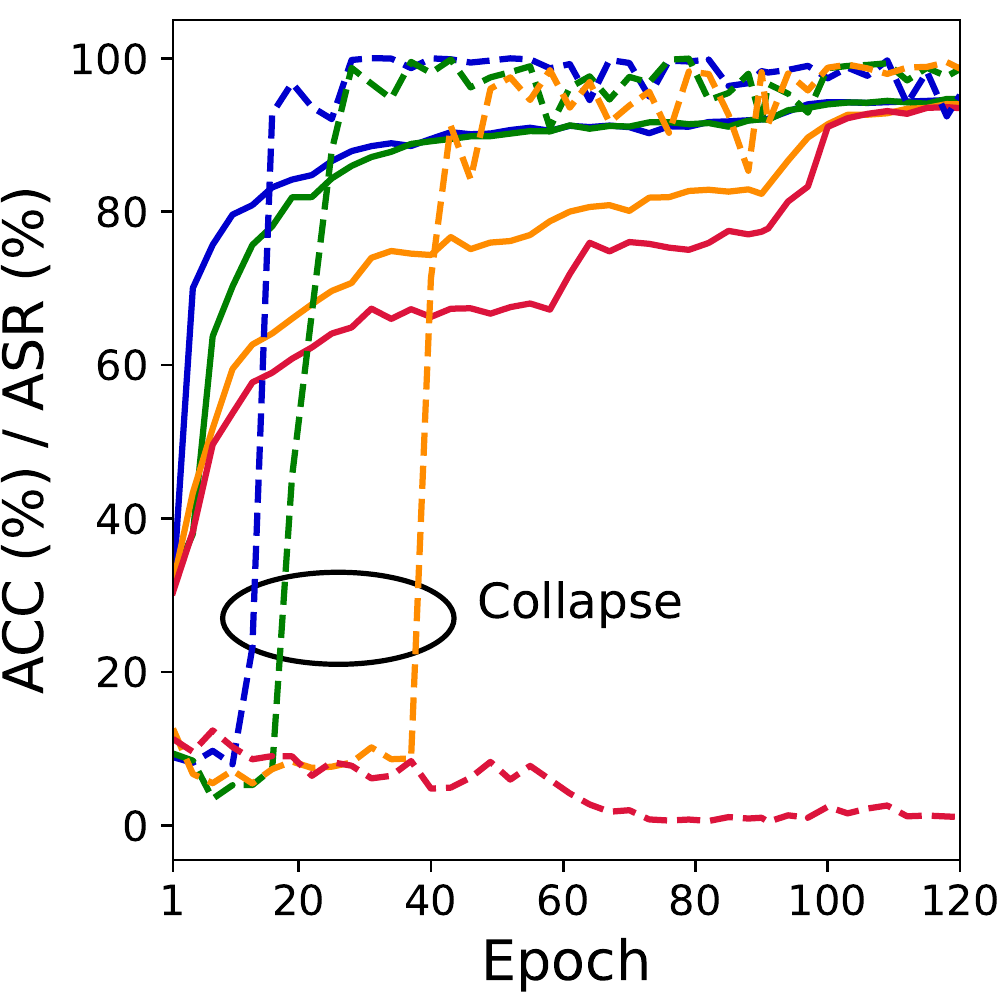}
        \caption{CLB}
        \label{warm-clb}
    \end{subfigure}
	\centering
 \end{minipage}
	\caption{The clean accuracy (ACC \%) and the attack success rate (ASR \%) of different warming-up strategies in stage 1 on CIFAR-10 for four backdoor attacks. A smaller $t$ and a larger $n$ correspond to a faster warming-up. The $t$ and $n$ should be set carefully to progressively increase the number of samples in $\mathcal{D}_C$ during stage 1 instead of building it in short time, which can prevent the collapse of ASD. 
    }
    \label{different warm-up strategies}
\end{figure*}

DBD can achieve a low ASR in most cases but fails sometimes, \textit{e.g.}, when it defends the IAB attack on CIFAR-10. The reason behind the observation is that DBD trains the linear layer on the whole poisoned dataset during stage 2 for 10 epochs, which introduces the risk to implant the backdoor. 
Besides, DBD needs a larger amount of time than our ASD, \textit{e.g.}, 8 times our ASD on ImageNet. 
Stage 3 of DBD under our framework also splits the poisoned dataset into two data pools adaptively. 
Hence, after stage 3 of DBD for 10 epochs, we replace the original data split method in DBD with our proposed meta-split. As shown in Fig. \ref{BadNets_with_meta_split}, stage 3 of DBD can be compressed from the original 190 epochs to 25 epochs and with 91$+\%$ ACC and 4$-\%$ ASR, which shows that our meta-split can help accelerate DBD. Moreover, we compare the number of clean hard samples and poisoned samples in $\mathcal{D}_{C}$ during the final 30 epochs of DBD and our ASD. Specifically, we choose 5,000 samples with the largest $\mathcal{L}_1(\cdot)$ losses chosen by the model as the clean hard samples. Fig. \ref{BadNets_DBD_ASD} demonstrates that our ASD can access much more clean hard samples than DBD and poisoned samples with a similar low scale. 
More results about the combination and comparison between DBD and our ASD are in Appendix \red{F} and \red{G}.

\subsection{Ablation Study on Defense Settings}
In summary, ASD is composed of three stages. Here we study the necessity of each stage by conducting the following experiments. Results are shown in Fig. \ref{different without}. We set the poisoned rate of BadNets, Blend and WaNet as 20\% and remain other settings unchanged. (1) \textbf{Without Stage 1}. The two data pools will approximate random initialization if we directly start our defense without stage 1. Fig. \ref{different without} indicates that the defense process will be completely disrupted with two randomly initialized data pools. (2) \textbf{Without Stage 2}. 
In stage 2, since less poisoned samples from the target class will be introduced to a larger $\mathcal{D}_{C}$ by class-agnostic loss-guided split, it can rapidly promote the ACC and suppress the ASR by a large margin. The defaulted ASD can achieve a lower ASR and a higher ACC than that without stage 2. 
(3) \textbf{Without stage 3}. As shown in Fig. \ref{different without}, ACC will achieve only about 80\% without stage 3 owing to the lack of the model-dependent clean hard samples in $\mathcal{D}_{C}$. More results about the ablation study on attack settings and defense settings are shown in Appendix \red{H} and \red{I}. \par

\noindent \textbf{Different warming-up strategies.} Compared with our default setting ($t=5$ and $n=10$), ACC can increase faster when building $\mathcal{D}_{C}$ in shorter time, \textit{i.e.}, $t$ is smaller and $n$ is larger, as shown in Fig. \ref{different warm-up strategies}. However, it can wrongly introduce a large number of poisoned samples into $\mathcal{D}_{C}$ and result in the failure of our ASD, especially under WaNet and CLB. Hence, it is necessary to control the speed to build $\mathcal{D}_{C}$ and constrain the number of samples in $\mathcal{D}_{C}$ during stage 1.
\par

\noindent \textbf{Different semi-supervised learning methods.} We treat the samples in $\mathcal{D}_P$ as unlabeled and apply semi-supervised learning to learn from both data pools. In this experiment, we show our ASD can work well with various semi-supervised learning, \textit{e.g.}, UDA \cite{xie2020unsupervised} and ReMixMatch \cite{berthelot2019remixmatch}. We keep all settings unchanged. As shown in Table \ref{different semi-supervised}, UDA and ReMixMatch can still have similar robustness against backdoor attacks compared with MixMatch \cite{berthelot2019mixmatch} under our proposed ASD. 
More details about these three  semi-supervised learning methods are in Appendix \red{J}. 
\par

\begin{table}[]
\caption{The clean accuracy (ACC \%) and the attack success rate (ASR \%) on CIFAR-10 of our ASD implemented by different semi-supervised methods. Consistent satisfactory results show the stability of ASD.}
\label{different semi-supervised}
\footnotesize
\setlength{\tabcolsep}{1.3mm}{
\begin{tabular}{l|llllllll}

\toprule[0.68pt]
\addlinespace[0pt]
\multicolumn{1}{c|}{\multirow{2}{*}{Method}} & \multicolumn{2}{c}{BadNets} & \multicolumn{2}{c}{Blend} & \multicolumn{2}{c}{WaNet} & \multicolumn{2}{c}{CLB} \\ \cline{2-9} 
\multicolumn{1}{c|}{}                  & ACC          & ASR          & ACC         & ASR         & ACC         & ASR         & ACC        & ASR        \\ \hline
MixMatch                               & 93.4 & 1.2	& 93.7 & 1.6	& 93.1 & 1.7 & 	93.1 & 0.9            \\
UDA                                   & 92.6 & 2.1	& 91.9 & 2.3	& 92.5 & 1.6	& 92.1 & 2.8           \\
ReMixMatch                             & 91.6 & 1.4	& 91.5 & 1.0	& 91.9 & 0	& 91.1 & 0.1           \\ 
\addlinespace[-0.22em]
\bottomrule[0.68pt]
\end{tabular}}
\end{table}

\begin{table}[]
\centering
\caption{The clean accuracy (ACC \%) and attack success rate (ASR \%) on CIFAR-10 for different numbers of poisoned samples in the seed sample.}
\label{seed sample with poisoned sample}
\footnotesize
\setlength{\tabcolsep}{3.1mm}{
\begin{tabular}{cllllll}
\toprule[0.68pt]
\addlinespace[0pt]
\multicolumn{2}{c}{Poisoned Number} & 0 & 1 & 2 & 3 & 4 \\ \hline
\multirow{2}{*}{BadNets}   & ACC  & 93.4 &  94.1 & 93.6 & 93.6 & 93.5  \\ \cline{2-7} 
                           & ASR  &  1.2 & 1.5 & 2.5 & 1.4 & 1.3  \\ \hline
\multirow{2}{*}{Blend}     & ACC  &  93.7 & 93.6 & 93.5 & 93.5 & \textit{93.1}   \\ \cline{2-7} 
                           & ASR  &  1.6 & 2.5 & 2.7 & 0.8 & \textit{99.9}     \\ \hline
\multirow{2}{*}{WaNet}     & ACC  &  93.1 & 93.6 & 93.7 & 93.4 & 93.5   \\ \cline{2-7} 
                           & ASR  &  1.7 & 1.4 & 2.2 & 4.1 & 5.4    \\ \hline
\multirow{2}{*}{CLB}     & ACC  &   93.1 & 93.5 & 91.3 & 93.7 & \textit{93.2}     \\ \cline{2-7} 
                           & ASR  &  0.9 & 1.3 & 2.5 & 1.5 & \textit{98.9}    \\ 
\addlinespace[-0.22em]
\bottomrule[0.68pt]
\end{tabular}}
\end{table}

\subsection{Ablation Study on Seed Sample Selection}
\label{sec:transfer-based}
In our method, we utilize the seed samples with 10 clean samples per class to warm up the model. Here, we discuss the flexibility of the seed sample selection. (1) Seed samples contain a few poisoned samples. (2) Seed samples are from another classical dataset, \textit{e.g.}, ImageNet. \par

First, we discuss the case that some poisoned samples are introduced in the seed samples and the results are shown in Table \ref{seed sample with poisoned sample}. It can be seen that our ASD can still exceed 91\% ACC and suppress the creation of backdoor even though the seed samples contain 1 $\sim$ 3 poisoned samples. Meanwhile, as the poisoned number increases to 4, our ASD can also defend against BadNets and WaNet successfully. This illustrates that our method has certain resistance to the seed samples mixed with a few poisoned samples. Then, we introduce the transfer learning-based ASD when adopting the seed samples from another classical dataset as follows.
\par

\noindent \textbf{Threat model}. Considering a more realistic scenario, we cannot obtain any clean sample from the source dataset. Here, we specify the source training data as CIFAR-10. However, only 100 clean samples from the classical ImageNet dataset are available.
\par

\noindent \textbf{Methods}. We first assign the 100 ImageNet clean samples as $\mathcal{D}_{C}$ and remove the labels of the entire poisoned CIFAR-10 as $\mathcal{D}_{P}$ and perform the semi-supervised learning for 10 epochs. Then we freeze the pre-trained backbone and fine-tune the linear layer on the entire poisoned dataset via supervised learning for 1 epoch. Finally, this model will be regarded as the initialized model of our ASD and other settings of our ASD remain unchanged.\par

\noindent \textbf{Results}. As shown in Table \ref{transfer-based experiment}, after the above transfer-based pre-training, the model will achieve about 52\% ACC and 9\% ASR. By adopting this transfer-based initialized model, our ASD achieves 92$+\%$ ACC and 4$-\%$ ASR, which shows our ASD can obtain robustness against backdoor attacks without clean seed samples from the poisoned training dataset. More results are in Appendix \red{I}.

\begin{table}[]
\caption{The clean accuracy (ACC \%) and the attack success rate (ASR \%) on CIFAR-10 under the proposed transfer-based setting. Transfer learning-based ASD works well. }
\label{transfer-based experiment}
\footnotesize
\setlength{\tabcolsep}{1.2mm}{
\begin{tabular}{l|llllllll}
\toprule[0.68pt]
\addlinespace[0pt]
\multicolumn{1}{c|}{\multirow{2}{*}{Method}}                                             & \multicolumn{2}{c}{BadNets}           & \multicolumn{2}{c}{Blend}             & \multicolumn{2}{c}{WaNet}             & \multicolumn{2}{c}{CLB}               \\ \cline{2-9} 
\multicolumn{1}{c|}{}                                                              & ACC               & ASR               & ACC               & ASR               & ACC               & ASR               & ACC               & ASR               \\ \hline
\multirow{2}{*}{\begin{tabular}[c]{@{}l@{}}Transfer-based \\ pre-training\end{tabular}} &  \multirow{2}{*}{52.6} & \multirow{2}{*}{8.9} & \multirow{2}{*}{51.5} & \multirow{2}{*}{10.2} & \multirow{2}{*}{53.4} & \multirow{2}{*}{9.3} & \multirow{2}{*}{52.6} & \multirow{2}{*}{9.5} \\
    &                   &                   &                   &                   &                   &                   &                   &                   \\ \hline
\multirow{2}{*}{\begin{tabular}[c]{@{}l@{}}Transfer-based \\ ASD\end{tabular}} & \multirow{2}{*}{92.9} & \multirow{2}{*}{2.4} & \multirow{2}{*}{92.5} & \multirow{2}{*}{2.6} & \multirow{2}{*}{92.5} & \multirow{2}{*}{3.5} & \multirow{2}{*}{92.1} & \multirow{2}{*}{2.8} \\
    &                   &                   &                   &                   &                   &                   &                   &                   \\ 
\addlinespace[-0.22em]
\bottomrule[0.68pt]
\end{tabular}
}
\end{table}

\subsection{Resistance to Potential Adaptive Attacks}
In the above experiments, we assume that attackers have no information about our backdoor defense. In this section, we consider a more challenging setting, where the attackers know the existence of our defense and can construct the poisoned dataset with an adaptive attack.  \par

\noindent \textbf{Threat model for the attackers.} Following existing work \cite{gu2017badnets,chen2017targeted,turner2018clean}, we assume that the attackers can access the entire dataset and know the architecture of the victim model. However, the attackers can not control the training process after poisoned samples are injected into the training dataset.

\noindent \textbf{Methods.} Our defense separates samples by the magnitude of the loss reduction in the final stage, so adaptive attacks should aim to minimize the difference in the loss reduction between clean samples and poisoned samples. First, the attackers train a clean model in advance. Then, since the gradient determines the loss reduction of the model \cite{zhang2004solving,kingma2014adam}, the trigger pattern can be optimized by minimizing the gradient for poisoned samples \textit{w.r.t} the trained model and maximizing that for clean samples.

\noindent \textbf{Settings.} We conduct experiments on CIFAR-10. Based on the clean model, we adopt projected gradient descent (PGD) \cite{madry2017towards} to optimize the trigger pattern for 200 iterations with a step size 0.001. Besides, we set the perturbation magnitude as $32/255$ and the trigger size as 32$\times$32. 
\par

\noindent \textbf{Results.} The adaptive attack can achieve 94.8\% ACC and 99.8\% ASR without any defense. However, this attack can obtain 93.6\% ACC and only 1.4\% ASR under our ASD, which illustrates our defense can resist the adaptive attack. The probable reason is that the trigger pattern is optimized on the surrogate clean model and has low transferability. The details of this adaptive attack and another designed adaptive attack are stated in Appendix \red{K} and Appendix \red{L}.

\section{Conclusion}
\vspace{-0.2em}
In this paper, we revisit training-time backdoor defenses in a unified framework from the perspective of splitting the poisoned dataset into two data pools. Under our framework, we propose a backdoor defense via adaptively splitting the poisoned dataset. Extensive experiments show that our ASD can behave effectively and efficiently against six state-of-the-art backdoor attacks. Furthermore, we explore a transfer-based ASD to show the flexibility of seed sample selection in our method. In summary, we believe that our ASD can serve as an effective tool in the community to improve the robustness of DNNs against backdoor attacks.

\section*{Acknowledgement}
This work is supported in part by the National Natural Science Foundation of China under Grant 62771248, Shenzhen Science and Technology Program (JCYJ20220818101012025), and the PCNL KEY project (PCL2021A07).

{\small
\bibliographystyle{ieee_fullname}
\bibliography{egbib}
}

\clearpage

\renewcommand\thesection{\Alph{section}}
\setcounter{section}{0}

\section{Algorithm outline}
The algorithm outline of ASD is listed as Algorithm \red{1}.

\begin{figure}[ht]
    \includegraphics[width=\linewidth]{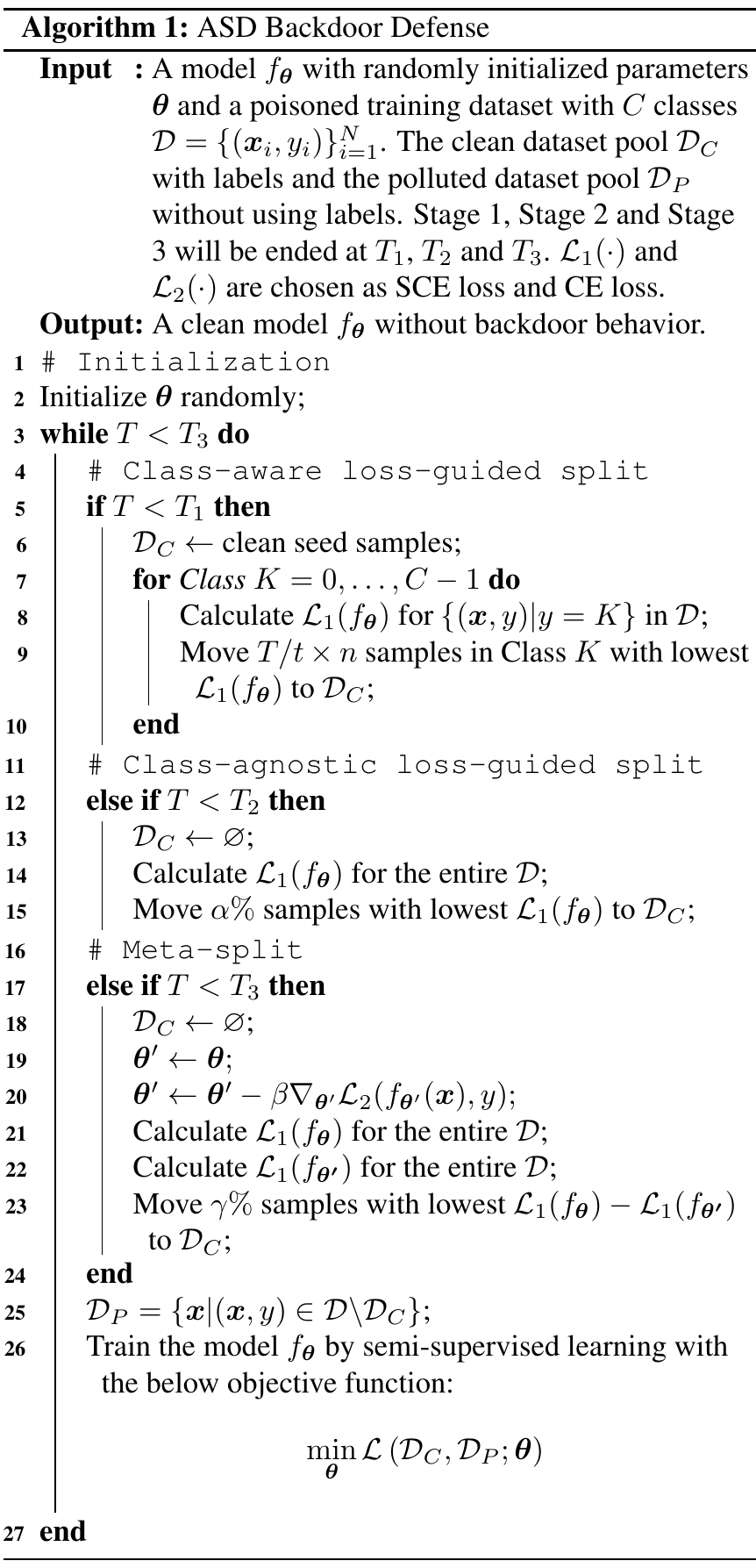}
\end{figure}

\section{Implementation details}
\label{sec:Implementation details}
In summary, we use the framework PyTorch \cite{paszke2019pytorch} to implement all the experiments. Note that the experiments on CIFAR-10 and GTSRB dataset are run on a NVIDIA GeForce RTX 2080 Ti GPU with 11GB memory and the experiments on ImageNet and VGGFace2 dataset are on a NVIDIA Tesla V100 GPU with 32GB memory.

\subsection{Datasets and DNN models}
The details of datasets and DNN models in our experiments are summarized in Table \ref{datasets and models}. Specially, we randomly choose 30 classes from ImageNet and VGGFace2 dataset to construct a subset due to the limitation of the computational time and costs.



\subsection{Attack setups}
\noindent \textbf{Training setups.} On the CIFAR-10 \cite{krizhevsky2009learning} and GTSRB \cite{stallkamp2011german} dataset, we perform backdoor attacks on ResNet-18 \cite{he2016deep} for 200 epochs with batch size 128. We adopt the stochastic gradient descent (SGD) \cite{zhang2004solving} optimizer with a learning rate 0.1, momentum 0.9, weight decay $5\times10^{-4}$. The learning rate is divided by 10 at epoch 100 and 150. On the ImageNet \cite{deng2009imagenet} dataset, we train ResNet-18 for 90 epochs with batch size 256. We utilize the SGD optimizer with a learning rate 0.1, momentum 0.9, weight decay $10^{-4}$. The learning rate is decreased by a factor of 10 at epoch 30 and 60. The image resolution will be resized to $224\times224\times3$ before attaching the trigger pattern. On VGGFace2 \cite{cao2018vggface2} dataset, the batch size is set to 32 and the targeted model is DenseNet-121 \cite{huang2017densely}. Other settings are the same as those used in training the models on ImageNet dataset.\par

\begin{table}[t]
\begin{center}
\caption{Summary of datasets and DNN models in our experiments.}
\label{datasets and models}
\footnotesize
\setlength{\tabcolsep}{0.3mm}{
\begin{tabular}{cccccc}
\toprule[0.68pt]
\multicolumn{1}{c}{\multirow{2}{*}{Dataset}} & \multicolumn{1}{c}{\multirow{2}{*}{\# Input size}}  & \multicolumn{1}{c}{\multirow{2}{*}{\# Classes}} & \multicolumn{1}{c}{\# Training}  & \multicolumn{1}{c}{\# Testing} & \multicolumn{1}{c}{\multirow{2}{*}{Models}}\\
 &  &  &  images &  images & \\
\midrule
CIFAR-10 & 3 $\times$ 32 $\times$ 32 & 10 & 50000 & 10000 & ResNet-18  \\
\midrule
GTSRB & 3 $\times$ 32 $\times$ 32 & 43 & 39209 & 12630 & ResNet-18  \\
\midrule
ImageNet & 3 $\times$ 224 $\times$ 224 & 30 & 38859 & 1500 & ResNet-18 \\
\midrule
VGGFace2 & 3 $\times$ 224 $\times$ 224 & 30 & 9000 & 2100 & DenseNet-121 \\
\bottomrule[0.68pt]
\end{tabular}}
\end{center}

\end{table}

\noindent \textbf{Settings for BadNets.} As suggested in \cite{huang2022backdoor,gu2017badnets}, we set a $2 \times 2$ square on the upper left corner as the trigger pattern on CIFAR-10 and GTSRB. For ImageNet and VGGFace2, we use a $32\times 32$ apple logo on the upper left corner. The ablation study for different trigger sizes and trigger locations has been shown in Appendix \ref{sec:ablation study on attack settings}.\par

\noindent \textbf{Settings for Blend.} Following \cite{huang2022backdoor,chen2017targeted}, we choose ``Hello Kitty" pattern on CIFAR-10 and GTSRB and the random noise pattern on ImageNet and VGGFace2. The blend ratio is set to 0.1. \par

\noindent \textbf{Settings for WaNet.} The original implementation of WaNet \cite{nguyen2021wanet} assumes that the attacker can control the training process. To apply WaNet in our poisoning-based attack threat model, we follow \cite{huang2022backdoor} to directly use the default warping-based operation to generate the trigger pattern. For CIFAR-10 and GTSRB, we set the noise rate $\rho_n$ = 0.2, control grid size $k$ = 4, and warping strength $s$ = 0.5. For ImageNet and VGGFace2, we choose the noise rate $\rho_n$ = 0.2, control grid size $k$ = 224, and warping strength $s$ = 1.\par

\noindent \textbf{Settings for IAB.} IAB \cite{nguyen2020input} also belongs to the control-training backdoor attacks as WaNet \cite{nguyen2021wanet}. As suggested in \cite{li2021anti}, we first reimplement the IAB attack as the original paper \cite{nguyen2020input} to obtain the trigger generator. Then we use the trigger generator to generate the poisoned samples in advance and conduct a poisoning-based backdoor attack.\par

\noindent \textbf{Settings for Refool.} Following \cite{liu2020reflection,li2021neural}, we randomly choose 5,000 images from PascalVOC \cite{everingham2012pascal} as the candidate reflection set $\mathcal{R}_{cand}$ and randomly choose one of the three reflection methods to generate the trigger pattern during the backdoor attack.

\noindent \textbf{Settings for CLB.} As the suggestions in \cite{huang2022backdoor,turner2018clean}, we adopt projected gradient descent (PGD) \cite{madry2017towards} to generate the adversarial perturbations \cite{bai2022improving,bai2020targeted,gu2020improving,wu2022towards,gu2022vision,liu2022watermark,bai2019hilbert,bai2021improving,bai2020improving} within $l_{\infty}$ ball and set its maximum magnitude $\epsilon=16$, step size 1.5, and 30 steps. The trigger pattern is the same as that in BadNets. More experiments of different settings for CLB are listed in Appendix \ref{sec:ablation study on attack settings}.\par

\subsection{Defense setups}
\noindent \textbf{Settings for FP.} FP \cite{liu2018fine} consists of two steps: pruning and fine-tuning. (1) We randomly select 5\% clean training samples as the local clean samples and forward them to obtain the activation values of neurons in the last convolutional layer. The dormant neurons on clean samples with the lowest $\alpha\%$ activation values will be pruned. (2) The pruned model will be fine-tuned on the local clean samples for 10 epochs. Specially, the learning rate is set as 0.01, 0.01, 0.1, 0.1 on CIFAR-10, GTSRB, ImageNet and VGGFace2. Unless otherwise specified, other settings are the same as those used in \cite{liu2018fine}.\par
Note that FP is sensitive to its hyper-parameters and we search for the best results by adjusting the pruned ratio $\alpha\% \in \{20\%, 30\%, 40\%, 50\%, 60\%, 70\%, 80\%, 90\%\}$ for six backdoor attacks on four datasets. \par

\begin{table*}[t]
\caption{The clean accuracy (ACC \%) and the attack success rate (ASR \%) of five backdoor defenses against six backdoor attacks on VGGFace2. Best results among five backdoor defenses are highlighted in \textbf{bold}.}
\label{results of vggface2}
\footnotesize
\setlength{\tabcolsep}{3.7mm}{
\begin{tabular}{l|ll|ll|ll|ll|ll|ll}
\toprule[0.68pt]
\addlinespace[0pt]
\multicolumn{1}{c|}{\multirow{2}{*}{Attack}} & \multicolumn{2}{c|}{No Defense}                    & \multicolumn{2}{c|}{FP }                            & \multicolumn{2}{c|}{NAD }                                                    & \multicolumn{2}{c|}{ABL }                           & \multicolumn{2}{c|}{DBD }                           & \multicolumn{2}{c}{ASD (Ours)}      \\ \cline{2-13} 
\multicolumn{1}{c|}{}                   & \multicolumn{1}{c}{ACC} & \multicolumn{1}{c|}{ASR} & \multicolumn{1}{c}{ACC} & \multicolumn{1}{c|}{ASR} & \multicolumn{1}{c}{ACC} & \multicolumn{1}{c|}{ASR} & \multicolumn{1}{c}{ACC} & \multicolumn{1}{c|}{ASR} & \multicolumn{1}{c}{ACC} & \multicolumn{1}{c|}{ASR} & \multicolumn{1}{c}{ACC} & \multicolumn{1}{c}{ASR}  \\ \hline
\multicolumn{1}{l|}{BadNets }                 &  91.7 & 99.9	& 91.5 & 100 & 56.1 & 6.5 & 91.2 & 19.6 & \textbf{91.6} & \textbf{0.4}	& 90.9 & 0.5	   \\
Blend                                       &   90.9 & 99.9	& 87.1 & 96.0 & 50.8 & 7.3  & 90.1 & 96.7  & 91.5 & 0.7	& \textbf{91.8} & \textbf{0.6}	 \\
WaNet                                        & 91.8 & 99.2	& 89.2 & 33.4 & 50.4 & 4.2 & \textbf{92.6} & 74.6 & 89.1 & 0.8	& 91.2 & \textbf{0.8}	
    \\
IAB                                         &   91.2 & 99.6	 & 90.6 & 97.2  & 43.1 & 5.5  & 91.3 & 59.7 & 90.1 & 2.5	& \textbf{92.1} & \textbf{0.4}	\\
Refool                                      &   90.7 & 98.3	 & 90.4 & 98.4   & 53.0 & 3.1  & 91.1 & 51.1 & \textbf{91.2} & \textbf{0.3} & 90.4 & 0.5	\\
CLB                                         &  91.8 & 98.9     & 90.9 & 99.9   & 40.0 & 3.3 & 91.3 & \textbf{0} & 90.4  & 0.3 & \textbf{91.8} & 0.2\\
\cline{1-13} 
Average                                     & 91.4 & 99.3 & 89.9 & 87.5 & 48.9 & 5.0 & 91.3 & 50.3 & 90.6 & 0.8 & \textbf{91.4} & \textbf{0.5}   
    \\ 
\addlinespace[-0.22em]
\bottomrule[0.68pt]
\end{tabular}}
\vspace{0.5em}
\end{table*}

\begin{table*}[t]
\caption{The clean accuracy (ACC \%) and the attack success rate (ASR \%) of five backdoor defenses against SSBA backdoor attack and \textit{all2all} attack on CIFAR-10. Best results among five backdoor defenses are highlighted in \textbf{bold}.}
\label{results of more attacks}
\footnotesize
\setlength{\tabcolsep}{3.78mm}{
\begin{tabular}{l|ll|ll|ll|ll|ll|ll}
\toprule[0.68pt]
\addlinespace[0pt]
\multicolumn{1}{c|}{\multirow{2}{*}{Attack}} & \multicolumn{2}{c|}{No Defense}                    & \multicolumn{2}{c|}{FP }                            & \multicolumn{2}{c|}{NAD }                                                    & \multicolumn{2}{c|}{ABL }                           & \multicolumn{2}{c|}{DBD }                           & \multicolumn{2}{c}{ASD (Ours)}      \\ \cline{2-13} 
\multicolumn{1}{c|}{}                   & \multicolumn{1}{c}{ACC} & \multicolumn{1}{c|}{ASR} & \multicolumn{1}{c}{ACC} & \multicolumn{1}{c|}{ASR} & \multicolumn{1}{c}{ACC} & \multicolumn{1}{c|}{ASR} & \multicolumn{1}{c}{ACC} & \multicolumn{1}{c|}{ASR} & \multicolumn{1}{c}{ACC} & \multicolumn{1}{c|}{ASR} & \multicolumn{1}{c}{ACC} & \multicolumn{1}{c}{ASR}  \\ \hline
\multicolumn{1}{l|}{SSBA}                 &  94.3 & 100 & \textbf{94.3} & 100 & 89.6 & 2.7 & 89.2 & 1.2 & 83.2 & \textbf{0.5} & 92.4 & 2.1  \\
\textit{all2all}                                   &  94.2 & 89.6 & 90.9 & 54.3 & 85.1 & 2.0 & 86.3 & 2.7 & 91.6 & \textbf{0.2} & \textbf{91.7} & 3.6	 \\
\addlinespace[-0.22em]
\bottomrule[0.68pt]
\end{tabular}}
\end{table*}

\begin{table}[t]
\caption{The clean accuracy (ACC \%) and the attack success rate (ASR \%) of three backdoor defenses against six backdoor attacks on CIFAR-10. Best results are highlighted in \textbf{bold}.}
\label{results of more defenses}
\footnotesize
\setlength{\tabcolsep}{2.8mm}{
\begin{tabular}{l|ll|ll|ll}
\toprule[0.68pt]
\addlinespace[0pt]
\multicolumn{1}{c|}{\multirow{2}{*}{Attack}}                & \multicolumn{2}{c|}{CutMix }                            & \multicolumn{2}{c|}{DPSGD }                         & \multicolumn{2}{c}{ASD (Ours)}      \\ \cline{2-7} 
\multicolumn{1}{c|}{}                   & \multicolumn{1}{c}{ACC} & \multicolumn{1}{c|}{ASR} & \multicolumn{1}{c}{ACC} & \multicolumn{1}{c|}{ASR} &  \multicolumn{1}{c}{ACC} & \multicolumn{1}{c}{ASR}   \\ \hline
\multicolumn{1}{l|}{BadNets}                 &  \textbf{95.8} & 99.9 & 55.9 & 10.9 & 93.4 & \textbf{1.2}   \\
Blend                                    &  \textbf{94.9} & 99.3 & 56.7 & 37.0 & 93.7 & \textbf{1.6} \\
WaNet                                    & \textbf{95.1} & 99.9 & 55.1 & 15.8 & 93.1 & \textbf{1.7} \\
IAB                                    & \textbf{94.9} & 100 & 85.9 & 99.7 & 93.2 & \textbf{1.3}	 \\
Refool                                    &  \textbf{95.4} & 99.9 & 55.4 & 59.2 & 93.5 & \textbf{0}  \\
CLB                                    &  \textbf{96.1} & 1.1 & 55.7 & 7.6 & 93.1 & \textbf{0.9} \\
\cline{1-7} 
Average                                     &   \textbf{95.4} & 83.4 & 60.8 & 38.4 & 93.3 & \textbf{1.1}
    \\ 
\addlinespace[-0.22em]
\bottomrule[0.68pt]
\end{tabular}}
\end{table}

\noindent \textbf{Settings for NAD.} NAD \cite{li2021neural} also aims to repair the backdoored model and need 5\% local clean training samples. NAD contains two steps: (1) We first use the local clean samples to fine-tune the backdoored model for 10 epochs. Specially, the learning rate is set as 0.01, 0.01, 0.1, 0.1 on CIFAR-10, GTSRB, ImageNet and VGGFace2. (2) The fine-tuned model and the backdoored model will be regarded as the teacher model and student model to perform the distillation process. . Unless otherwise specified, other settings are the same as those used in \cite{li2021neural}. \par
We find that NAD is sensitive to the hyper-parameter $\beta$ in the distillation loss. Therefore, we search for the best results by adjusting the hyper-parameter $\beta$ from \{500, 1000, 1500, 2000, 2500, 5000, 7500, 10000\} for six backdoor attacks on four datasets. \par

\noindent \textbf{Settings for ABL.} ABL \cite{li2021anti} contains three stages: (1) To obtain the poisoned samples, ABL first train the model on the poisoned dataset for 20 epochs by LGA loss \cite{li2021anti} and isolate 1\% training samples with the lowest loss. (2) Continue to train the model with the poisoned dataset after the backdoor isolation for 70 epochs. (3) Finally, the model will be unlearned by the isolation samples for 5 epochs. The learning rate is $5\times 10^{-4}$ at the unlearning stage. Unless otherwise specified, other settings are the same as those used in \cite{li2021anti}. \par
We find that ABL is sensitive to the hyper-parameter $\gamma$ in LGA loss. We search for the best results by adjusting the hyper-parameter $\gamma$ from \{0, 0.1, 0.2, 0.3, 0.4, 0.5\} for six backdoor attacks on four datasets.\par

\noindent \textbf{Settings for DBD.} DBD \cite{huang2022backdoor} contains three independent stages: (1) DBD uses SimCLR \cite{chen2020simple} to perform the self-supervised learning for 1,000 epochs. (2) Freeze the backbone and fine-tune the linear layer by supervised learning for 10 epochs. (3) Adopt the MixMatch \cite{berthelot2019mixmatch} to conduct the semi-supervised learning for 200 epochs on CIFAR-10 and GTSRB for 90 epochs on ImageNet and VGGFace2. Unless otherwise specified, other settings are the same as those used in \cite{huang2022backdoor}. Since DBD is a stable backdoor defense and not sensitive to its hyper-parameter, we directly use the default hyper-parameters and report the results. \par

\noindent \textbf{Settings for our ASD.} We adopt MixMatch \cite{berthelot2019mixmatch} as our semi-supervised learning framework and utilize Adam optimizer with a learning rate 0.002 and batch size 64 to conduct the semi-supervised training. The temperature $T$ is set as 0.5 and the weight of unsupervised loss $\lambda_u$ is set as 15. We treat the clean data pool $\mathcal{D}_C$ as a labeled container and the polluted data pool $\mathcal{D}_P$ as unlabeled. Our three stages are performed ended at $T_1=60$, $T_2=90$ and $T_3=120$ on CIFAR-10 and ImageNet and $T_3=100$ on GTSRB.\par 
In the first stage, we fixed the clean seed samples in $\mathcal{D}_{C}$ and these clean seed samples will not be removed in the first stage. The clean seed samples consist of 10 samples per class. Besides, we adopt class-aware loss-guided data split with $\mathcal{L}_1(\cdot)$ to progressively increase the number of the seed samples. The number of each class will add $n=10$ at every $t=5$ epochs. Specially, $\mathcal{L}_1(\cdot)$ is chosen as symmetric cross-entropy (SCE) \cite{wang2019symmetric}, as suggested in \cite{huang2022backdoor}. 
In the second stage, we use class-agnostic loss-guided data split to choose $\alpha\%=50\%$ samples with the lowest $\mathcal{L}_1(\cdot)$ losses into $\mathcal{D}_{C}$. 
In the third stage, we adopt the meta-split to split $\gamma\%=50\%$ samples with the lowest $\mathcal{L}_1(f_{\bm{\theta}})-\mathcal{L}_1(f_{\bm{\theta'}})$ losses into $\mathcal{D}_{C}$. In meta-split, we adopt stochastic
gradient descent (SGD) optimizer with the learning rate $\beta=0.015$ and batch size 128 to perform one supervised learning for $f_{\bm{\theta}}$ to obtain a virtual model $f_{\bm{\theta'}}$. For the virtual model, we update half of the layers of its feature extractor and its linear layer. Note that $f_{\bm{\theta'}}$ is only used for data splits and will not be involved in the followed training process. Besides, we adaptively split the poisoned training dataset every epoch and the polluted data pool $\mathcal{D}_P$ is formed with the remaining samples in the poisoned training dataset except the samples in the clean data pool $\mathcal{D}_C$.

\section{Results on VGGFace2 dataset}
We conduct the experiments on VGGFace2 \cite{cao2018vggface2} dataset and set the model architecture as DenseNet-121 \cite{huang2017densely}. The results against six backdoor attacks are shown in Table \ref{results of vggface2}. We also search for the best results for FP, NAD and ABL in different parameters. Besides, we set the learning rate of supervised training in the meta-split of ASD as 0.02. Unless otherwise specified, other settings remain unchanged.  Compared with the previous four backdoor defenses, our ASD can still achieve higher ACC and lower ASR on average, which verifies the superiority of our proposed ASD.

\begin{figure*}[t]
	\centering
    \begin{subfigure}[]{0.48\linewidth}
        \includegraphics[width=\linewidth]{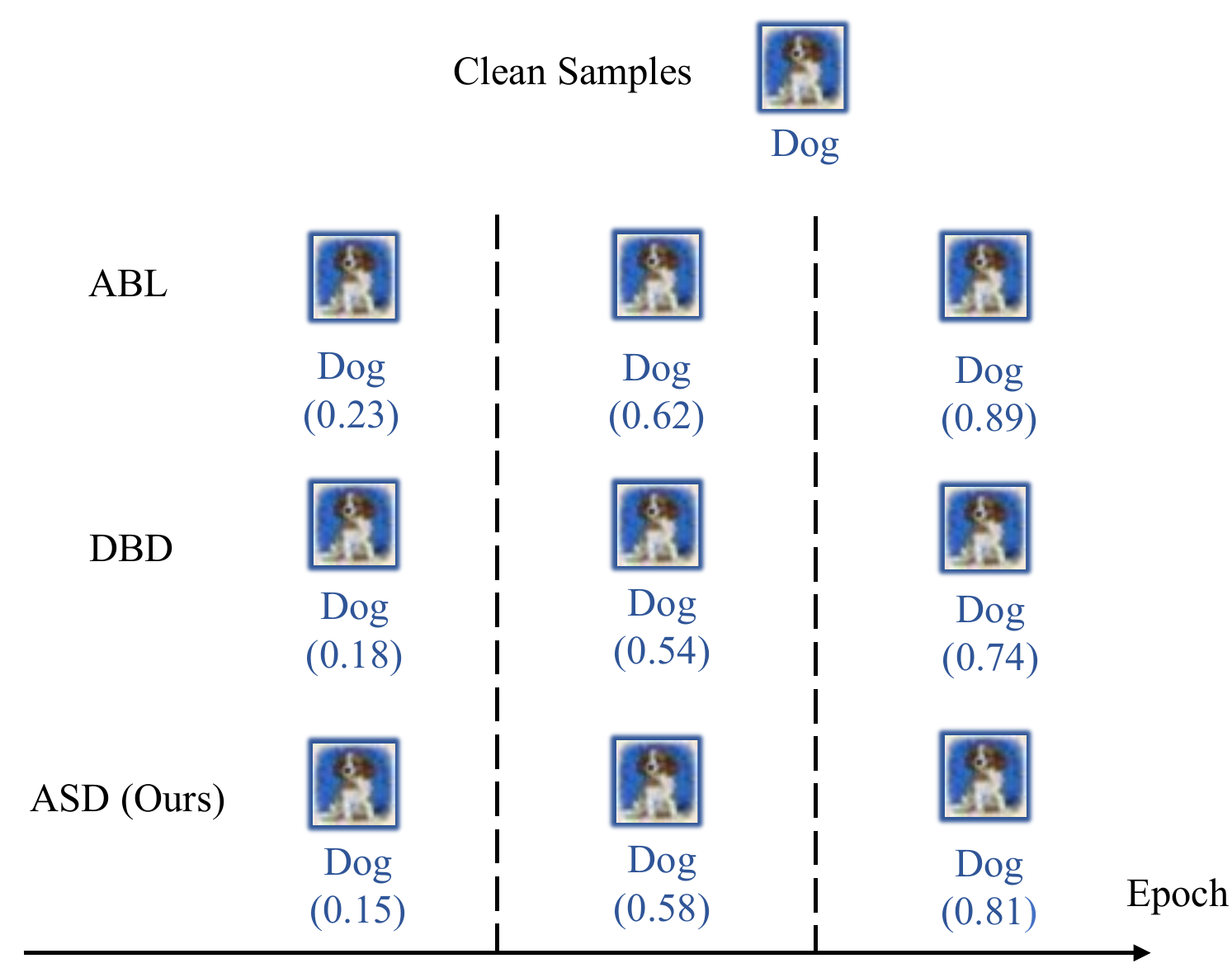}
        \caption{Clean Samples}
        \label{purify-clean}
    \end{subfigure}
    \begin{subfigure}[]{0.48\linewidth}
        \includegraphics[width=\linewidth]{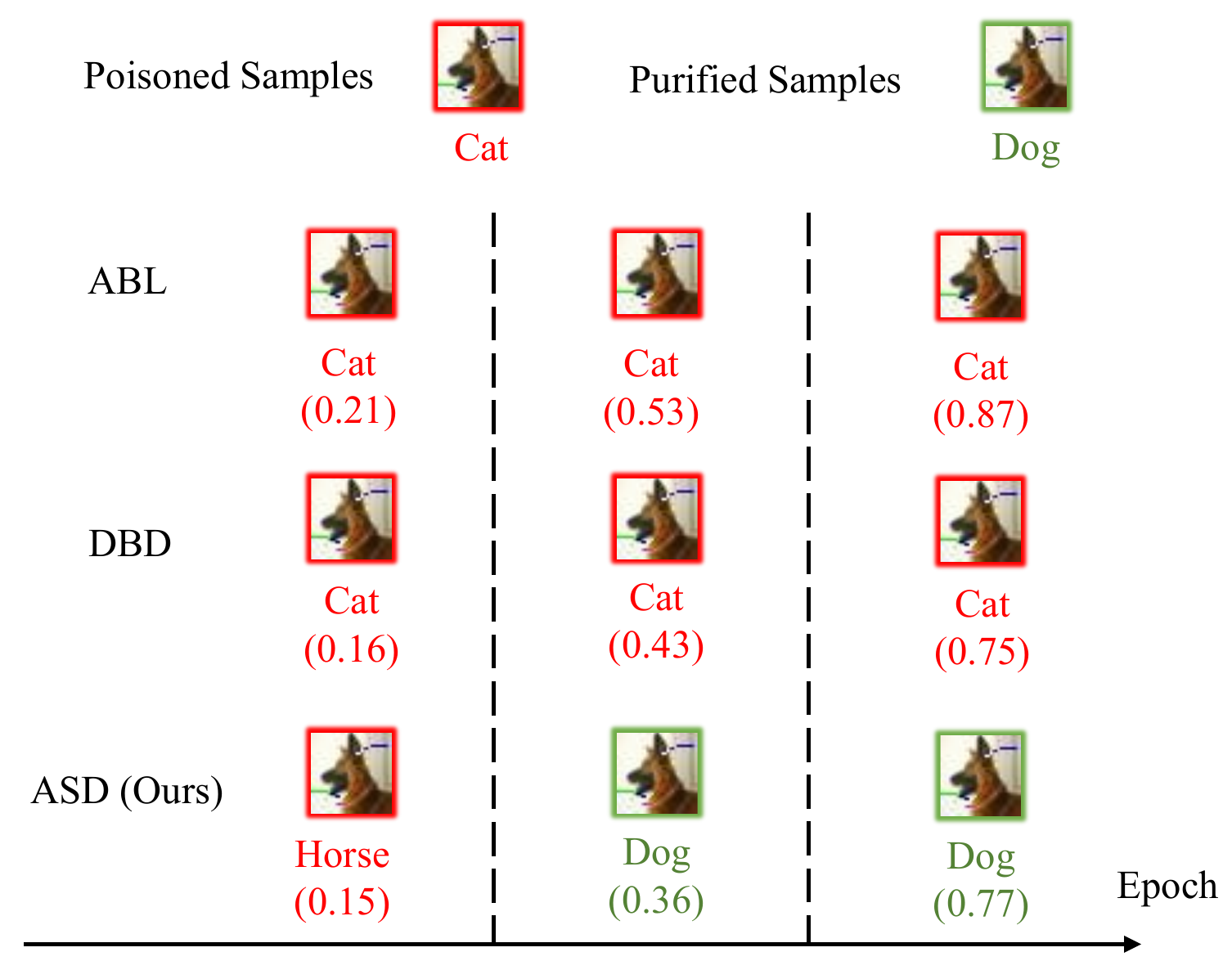}
        \caption{Poisoned Samples}
        \label{purify-backdoor}
    \end{subfigure}
	\centering
	\caption{The label and the logit score for the clean samples and poisoned samples on the model trained by ABL, DBD and our ASD against IAB. Our ASD can purify the poisoned samples successfully during the training process.}
    \label{purify all}
\end{figure*}

\section{Results about more backdoor attacks}
In addition to six backdoor attacks in our main experiment, we also test our ASD on another two backdoor attack paradigms, \textit{i.e.}, Sample-specific backdoor attack (SSBA) \cite{li2021invisible}, and \textit{all2all} backdoor attack (\textit{all2all}). For SSBA, we follow \cite{li2021invisible} to use the same encoder-decoder network to generate the poisoned samples. Note that all the backdoor attacks in the above experiments belong to the \textit{all2one} attack and they relabel the poisoned samples to a target label. For the \textit{all2all} attack, we relabel samples from class $i$ as class ($i+1$) and we adopt IAB \cite{nguyen2020input} as the trigger pattern, as suggested in \cite{wu2021adversarial}. 
The results are shown in Table \ref{results of more attacks}, which verifies that ASD can defend against these two backdoor attacks successfully.

\section{Results about more backdoor defenses}
We evaluate another two training-time backdoor defenses, \textit{i.e.}, CutMix-based backdoor defense (CutMix) \cite{borgnia2021strong} and differential privacy SGD-based backdoor defense (DPSGD) \cite{du2019robust}. For CutMix, we implement it as the defaulted setting in the original paper \cite{borgnia2021strong}. For DPSGD, we set the clipping bound $C=1$ and select the best noise scale $\sigma$ by the grid-search. We demonstrate the results in Table \ref{results of more defenses} and our ASD can still behave better than those two defenses on average. Besides, we also show the purification process of our ASD in Fig. \ref{purify all}.

\section{Combination between DBD and our meta-split}
\begin{figure*}[t]
	\centering
    \begin{subfigure}[]{0.24\linewidth}
        \includegraphics[width=\linewidth]{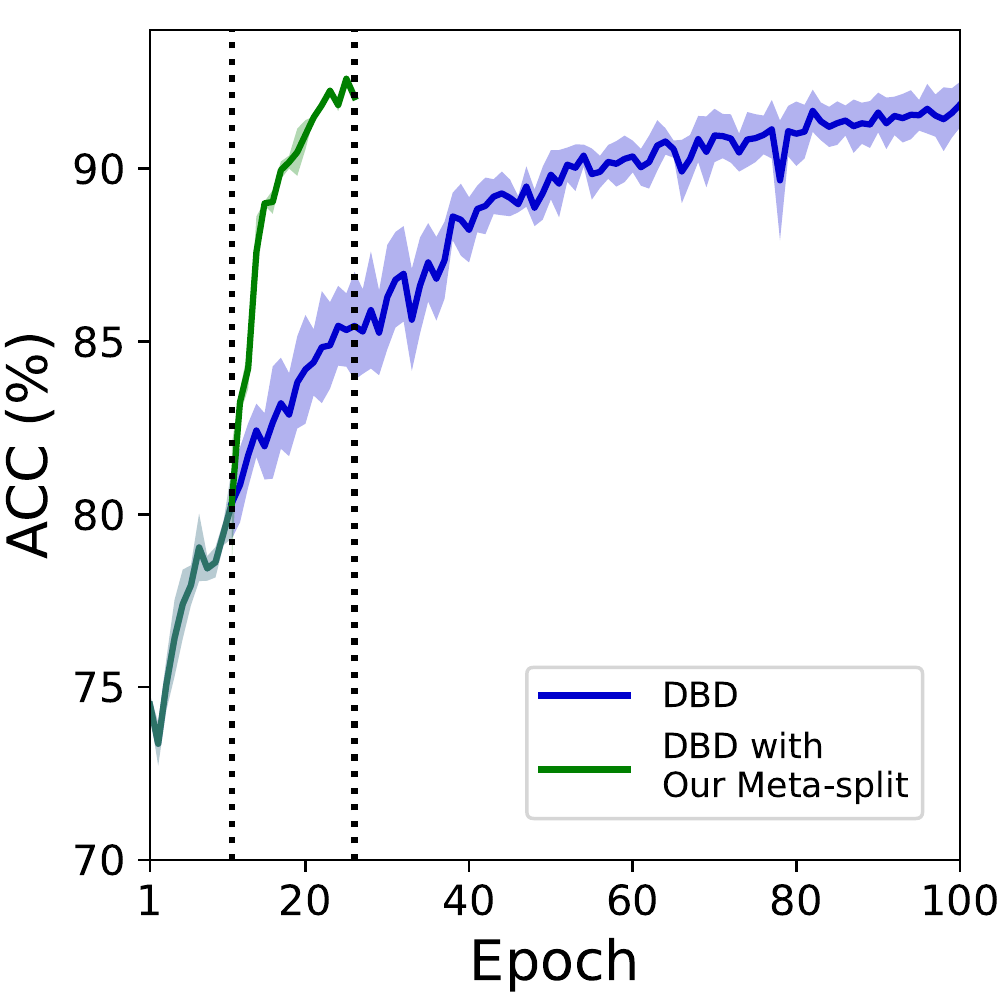}
        \caption{BadNets}
        \label{dbdtune-BadNets}
    \end{subfigure}
    \begin{subfigure}[]{0.24\linewidth}
        \includegraphics[width=\linewidth]{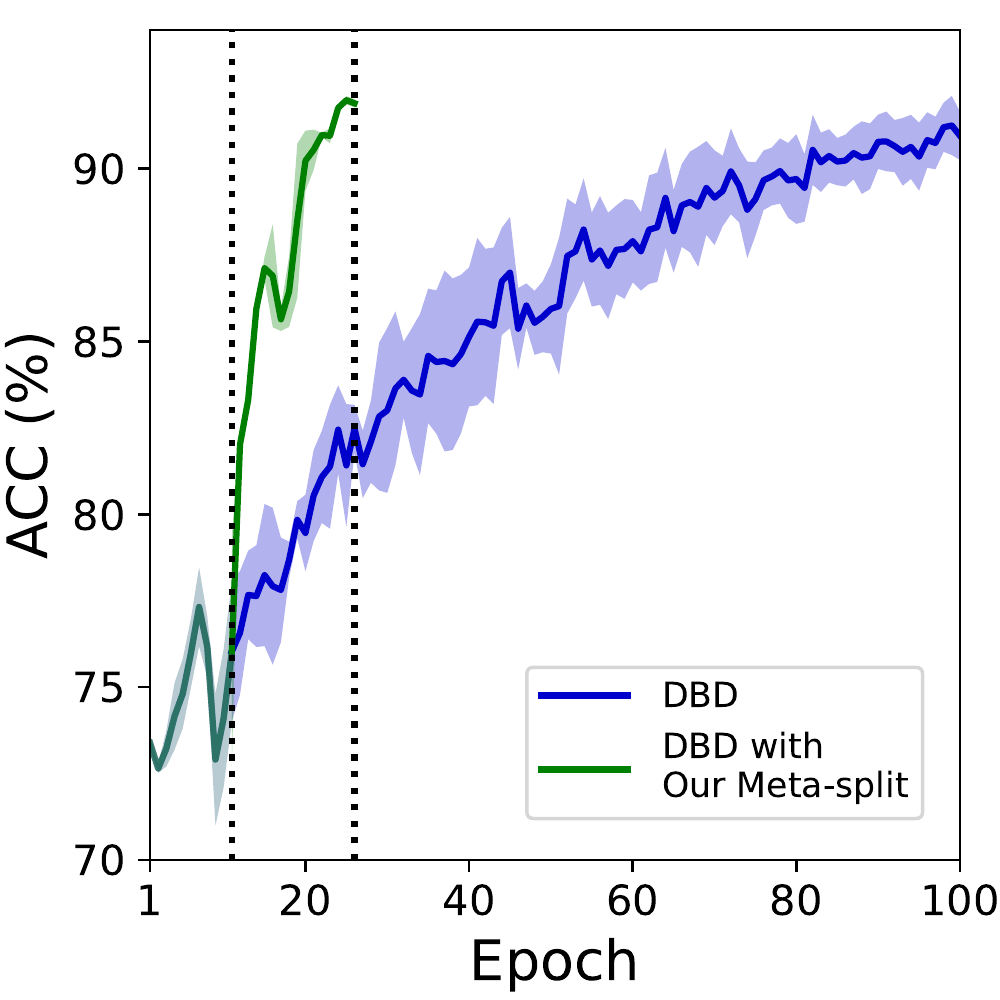}
        \caption{Blend}
        \label{dbdtune-Blend}
    \end{subfigure}
    \begin{subfigure}[]{0.24\linewidth}
        \includegraphics[width=\linewidth]{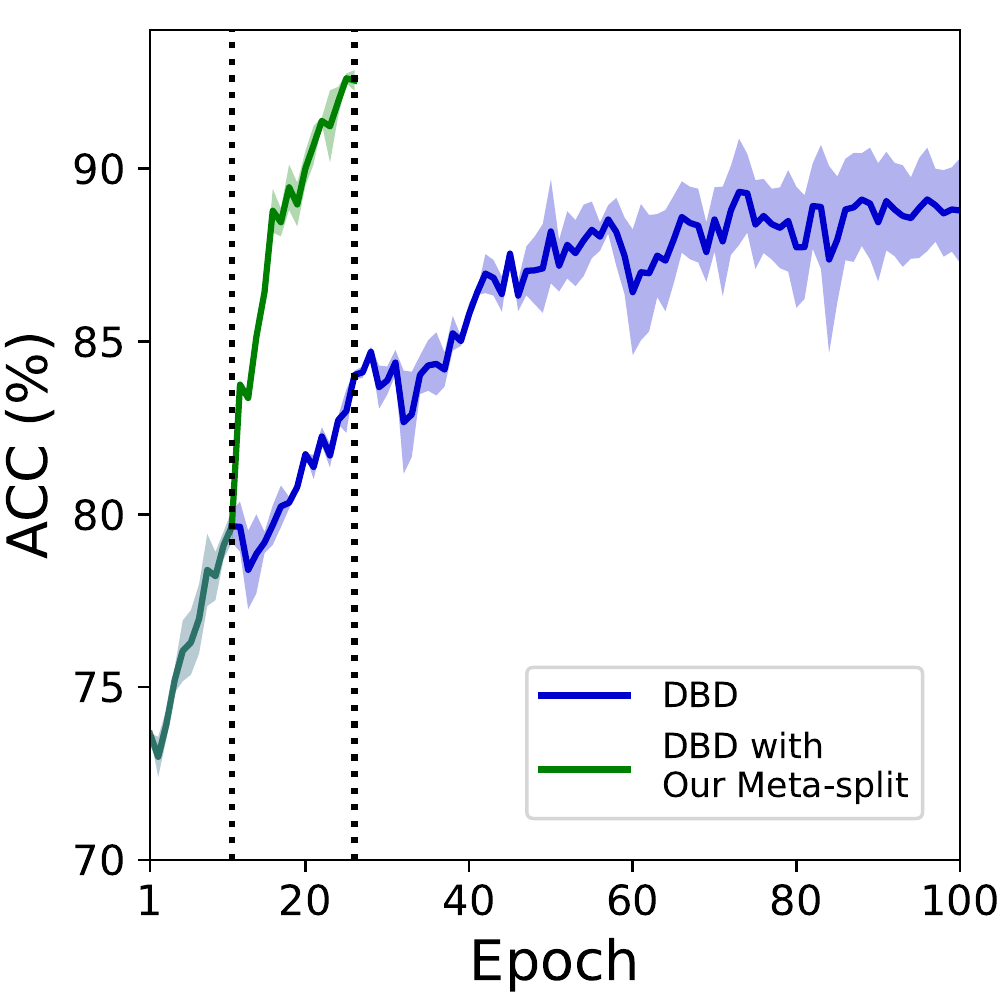}
        \caption{WaNet}
        \label{dbdtune-wanet}
    \end{subfigure}
    \begin{subfigure}[]{0.24\linewidth}
        \includegraphics[width=\linewidth]{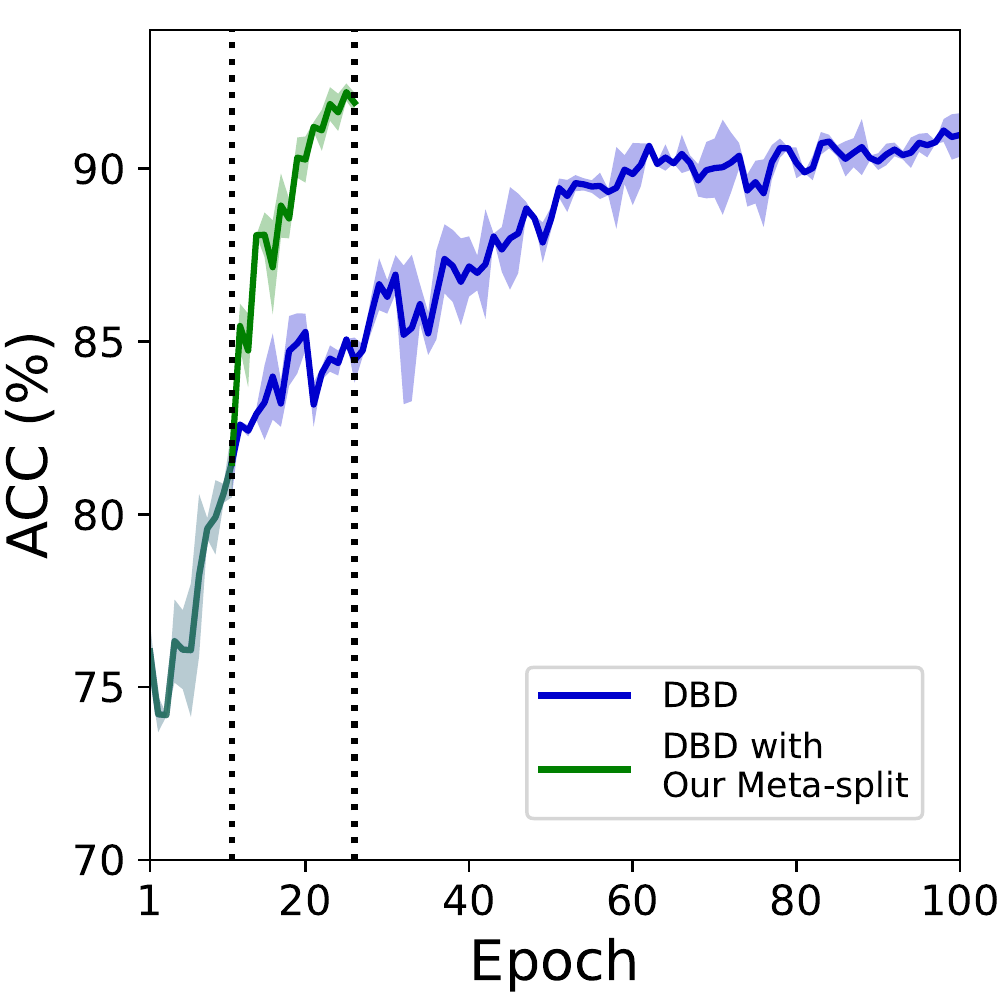}
        \caption{CLB}
        \label{dbdtune-clb}
    \end{subfigure}
	\centering
	\caption{Apply our meta-split to DBD on CIFAR-10 for four backdoor attacks, \textit{i.e.}, BadNets, Blend, WaNet, and CLB. Our proposed meta-split can help accelerate DBD.}
    \label{Apply our meta-split to DBD}
\end{figure*}

We show more results about combining our meta-split with DBD in 
Fig. \ref{Apply our meta-split to DBD}. From the overall results, we can observe that DBD can achieve 91$+\%$ ACC and 4$-\%$ ASR and its training time will reduce a lot with our meta-split.


\section{Comparison between DBD and our ASD}
\begin{figure*}[t]
	\centering
    \begin{subfigure}[]{0.24\linewidth}
        \includegraphics[width=\linewidth]{PDFs/hardsample/badnets.pdf}
        \caption{BadNets}
        \label{hardsample-BadNets}
    \end{subfigure}
    \begin{subfigure}[]{0.24\linewidth}
        \includegraphics[width=\linewidth]{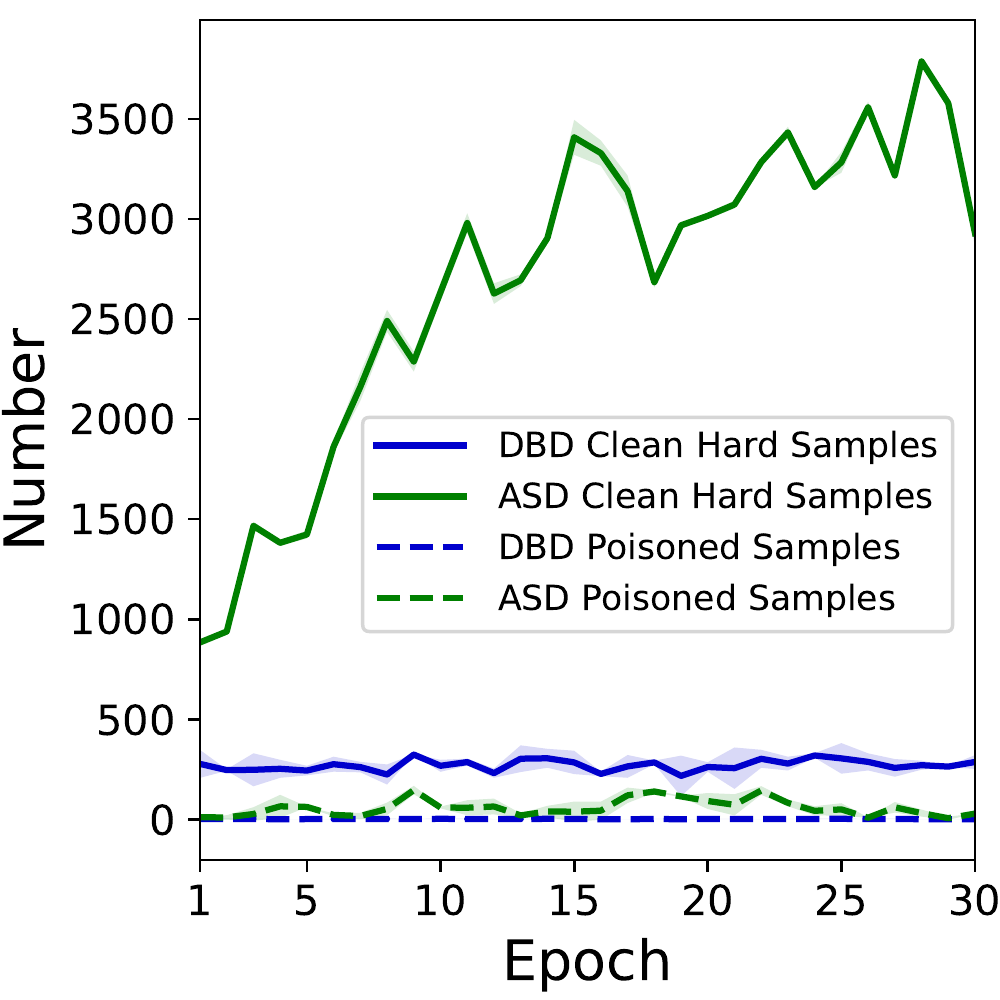}
        \caption{Blend}
        \label{hardsample-Blend}
    \end{subfigure}
    \begin{subfigure}[]{0.24\linewidth}
        \includegraphics[width=\linewidth]{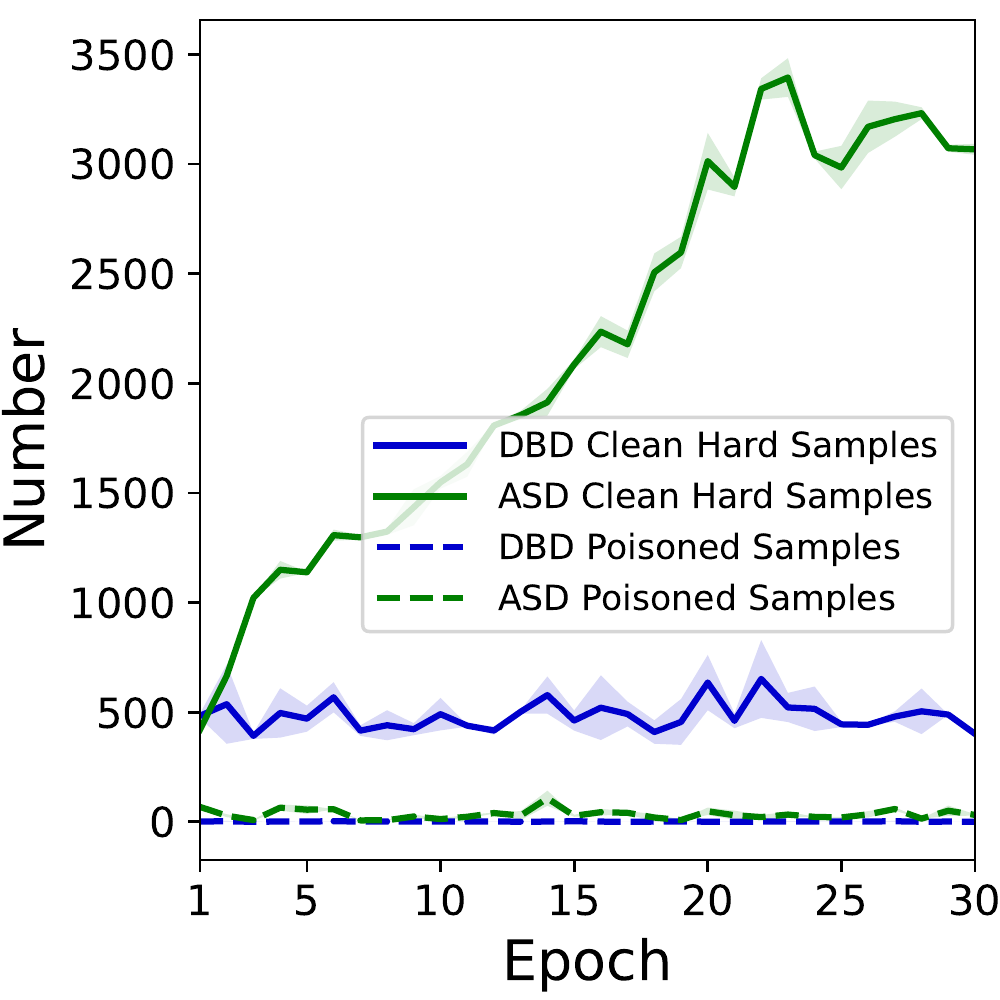}
        \caption{WaNet}
        \label{hardsample-wanet}
    \end{subfigure}
    \begin{subfigure}[]{0.24\linewidth}
        \includegraphics[width=\linewidth]{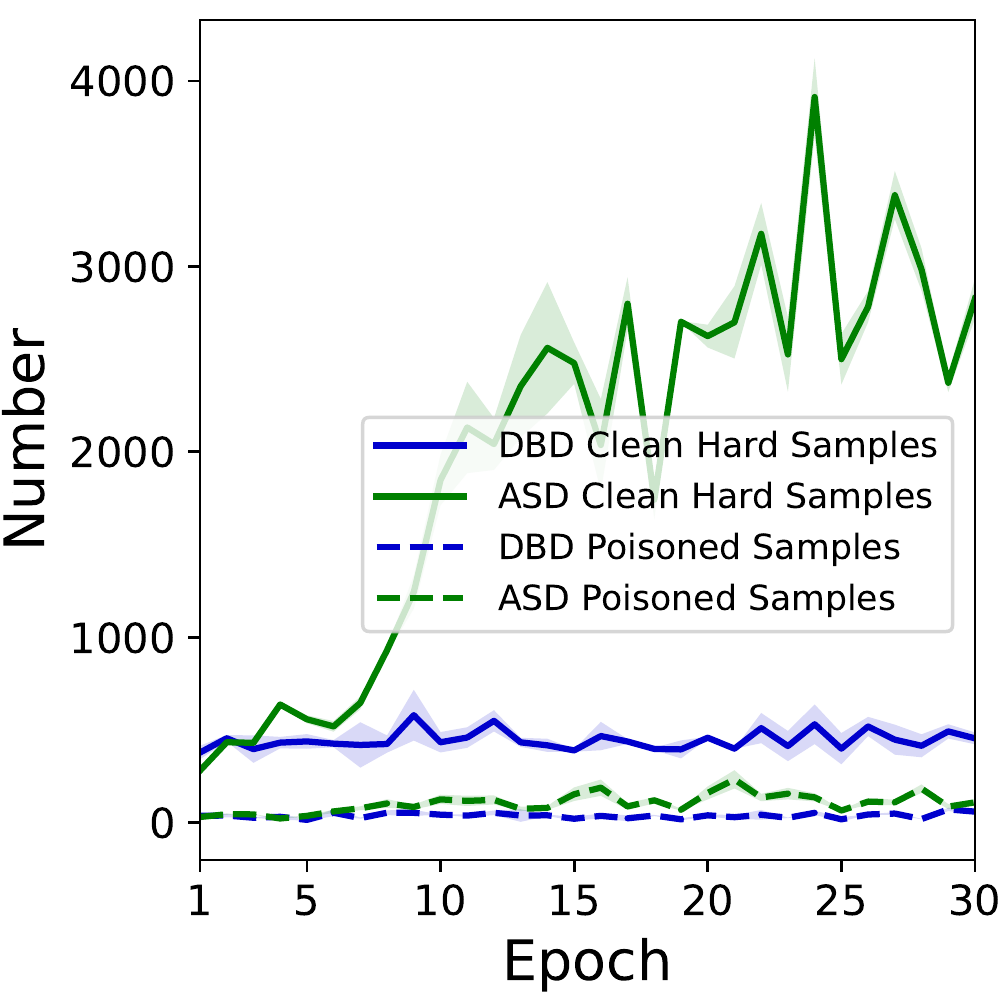}
        \caption{CLB}
        \label{hardsample-clb}
    \end{subfigure}
	\centering
	\caption{The number of clean hard samples and poisoned samples to be split in $\mathcal{D}_{C}$ for DBD and our ASD on CIFAR-10 for four backdoor attacks, \textit{i.e.}, BadNets, Blend, WaNet, and CLB. Our ASD can select clean hard samples.}
    \label{more results of hardsample}
\end{figure*}

We choose 5,000 samples with the largest $\mathcal{L}_1(\cdot)$ losses chosen by the model as clean hard samples. We show more results about the number of clean hard samples and poisoned samples to be split in $\mathcal{D}_{C}$ for DBD and our ASD in Fig. \ref{more results of hardsample}.

Furthermore, we list the split rate (\%) of poisoned samples in $\mathcal{D}_C$ along with the ACC (\%) / ASR (\%). Note that we report the split rate in the maximum value during the whole defense process. Here, we compare our ASD with ABL for WaNet in Table \ref{split rate ACC ASR ABL and ASD} and DBD for IAB in Table \ref{split rate ACC ASR DBD and ASD}. Our ASD can achieve the lower split rates, higher ACCs and lower ASRs. Specially, the split rate of poisoned samples in $\mathcal{D}_C$ is less than 1.2\% during the whole ASD training. 
%

\begin{table}[t]
\caption{The clean accuracy (ACC \%), the attack success rate (ASR \%) and the corresponding split rate of poisoned samples on CIFAR-10 for WaNet. Our ASD can achieve better results and a lower split rate of poisoned samples.}
\label{split rate ACC ASR ABL and ASD}
\footnotesize
\setlength{\tabcolsep}{3mm}{
\begin{tabular}{c|c|cc}
\toprule[0.68pt]
\addlinespace[0pt]
 & Split rate of poisoned samples & ACC & ASR  \\ \hline
 ABL  &  33.2   &   84.1   &  2.2  \\  
 ASD (Ours)  &  1.2  &  93.1  &  1.7  \\ 

\addlinespace[-0.22em]
\bottomrule[0.68pt]
\end{tabular}}
\end{table}

\begin{table}[t]
\caption{The clean accuracy (ACC \%), the attack success rate (ASR \%) and the corresponding split rate of poisoned samples on CIFAR-10 for IAB. Our ASD can achieve better results and a lower split rate of poisoned samples.}
\label{split rate ACC ASR DBD and ASD}
\footnotesize
\setlength{\tabcolsep}{3mm}{
\begin{tabular}{c|c|cc}
\toprule[0.68pt]
\addlinespace[0pt]
 & Split rate of poisoned samples & ACC & ASR  \\ \hline
 DBD  &  9.2   &   91.6   &  100  \\  
 ASD (Ours)  &  1.1  &  93.2  &  1.3  \\ 

\addlinespace[-0.22em]
\bottomrule[0.68pt]
\end{tabular}}
\end{table}


\begin{table}[t]
\caption{The clean accuracy (ACC \%) and the attack success rate (ASR \%) on CIFAR-10 for different target labels. Our ASD can work well under different target labels.}
\label{target label}
\footnotesize
\setlength{\tabcolsep}{1.9mm}{
\begin{tabular}{cllllll}
\toprule[0.68pt]
\addlinespace[0pt]
\multicolumn{2}{c}{Target Label} & $y_t=0$ & $y_t=1$ & $y_t=2$ & $y_t=3$ & $y_t=4$ \\ \hline
\multirow{2}{*}{BadNets}   & ACC  &  92.8   &  93.1   &   92.7   &   93.4   &   93.2   \\ \cline{2-7} 
                           & ASR  &   1.1  &   0.4  &   1.5   &   1.2   &   0.7   \\ \hline
\multirow{2}{*}{Blend}     & ACC  &  93.5 &   93.5  &   93.4  &   93.7   &   93.7        \\ \cline{2-7} 
                           & ASR  &  0.9 &   0.3  &   1.1  &   1.6   &   1.2        \\ \hline
\multirow{2}{*}{WaNet}     & ACC  &  93.8 &  92.6  &   92.4  &   93.1   &   93.5         \\ \cline{2-7} 
                           & ASR  &  1.1 &   0.6  &   1.4  &   1.7   &    2.1        \\ \hline
\multirow{2}{*}{CLB}     & ACC  &  93.4  &  93.7  &  92.8 &  93.1  &  93.2          \\ \cline{2-7} 
                           & ASR  &  0.7  &  1.4  &  1.1  &   0.9  &  0.4          \\ 
\addlinespace[-0.22em]
\bottomrule[0.68pt]
\end{tabular}}
\end{table}

\begin{table}[t]
\caption{The clean accuracy (ACC \%) and the attack success rate (ASR \%) on CIFAR-10 for different poisoned rates. Our ASD can work well under different poisoned rates.}
\label{poisoned rate}
\footnotesize
\setlength{\tabcolsep}{3.0mm}{
\begin{tabular}{cllllll}
\toprule[0.68pt]
\addlinespace[0pt]
\multicolumn{2}{c}{Poisoned rate} & 1\% & 5\% & 10\% & 15\% & 20\% \\ \hline
\multirow{2}{*}{BadNets}   & ACC  &   93.8  &  93.4   &   94.2   &   92.3   &  93.6    \\ \cline{2-7} 
                           & ASR  &   1.8  &  1.2   &   1.8   &   1.1   &   0.9   \\ \hline
\multirow{2}{*}{Blend}     & ACC  &   93.8  &   93.7  &   92.6   &   93.5   &  93.8    \\ \cline{2-7} 
                           & ASR  &   3.9  &  1.6   &   1.2   &   1.8   &   1.7   \\ \hline
\multirow{2}{*}{WaNet}     & ACC  &   93.8  &  93.1   &  93.5    &   93.6   &  93.2    \\ \cline{2-7} 
                           & ASR  &   3.6  &   1.7  &   0.9   &   1.6   &   1.9   \\ 
\addlinespace[-0.22em]
\bottomrule[0.68pt]
\end{tabular}}
\end{table}

\section{Ablation study on attack settings}
\label{sec:ablation study on attack settings}
\noindent \textbf{Different target labels.} We evaluate our ASD using different target labels $y_t \in \{0,1,2,3,4\}$. The results are shown in Table \ref{target label}, which verifies the effectiveness of the proposed ASD. \par

\begin{table}[t]
\caption{The clean accuracy (ACC \%) and the attack success rate (ASR \%) on CIFAR-10 for different poisoned rates for CLB. Our ASD can work well under different poisoned rates for CLB.}
\label{poisoned rate for clb}
\footnotesize
\setlength{\tabcolsep}{4.5mm}{
\begin{tabular}{cllll}
\toprule[0.68pt]
\addlinespace[0pt]
\multicolumn{2}{c}{Poisoned rate} & 0.6\% & 2.5\% & 5\%  \\ \hline
\multirow{2}{*}{CLB ($\epsilon=16$)}   & ACC  &   93.4  &  93.1   &   93.1    \\ \cline{2-5} 
                           & ASR  &   3.1  &  0.9   &   0     \\ \hline
\multirow{2}{*}{CLB ($\epsilon=32$)}     & ACC  &   94.1  &   93.4  &   93.3     \\ \cline{2-5} 
                           & ASR  &   2.0  &  1.5   &   1.3   \\ 
\addlinespace[-0.22em]
\bottomrule[0.68pt]
\end{tabular}}
\end{table}

\begin{table}[t]
\caption{The clean accuracy (ACC \%) and the attack success rate (ASR \%) on CIFAR-10 for different trigger locations of BadNets. Our ASD can work well under different trigger locations of BadNets.}
\label{trigger location of BadNets}
\footnotesize
\setlength{\tabcolsep}{2.3mm}{
\begin{tabular}{cllllll}
\toprule[0.68pt]
\addlinespace[0pt]
\multicolumn{2}{c}{\multirow{2}{*}{Location}} & Upper & Upper & Lower & Lower  & \multicolumn{1}{c}{\multirow{2}{*}{Center}}  \\
 & & left & right & left & right &   \\\hline
No    & ACC  &  94.9  &  93.7  &  94.5 & 94.1 & 93.8     \\ \cline{2-7} 
                     Defense      & ASR  &  100  & 99.7  & 99.8 & 100 & 100       \\ \hline
ASD    & ACC  &   93.4  &   93.1  &  93.7 & 92.8 & 93.6      \\ \cline{2-7} 
                        (Ours)   & ASR  &   1.2  &  1.1   &  0.9 & 0.8 & 1.7    \\ 
                        \addlinespace[-0.22em]
\bottomrule[0.68pt]
\end{tabular}}
\end{table}

\begin{table}[t]
\caption{The clean accuracy (ACC \%) and the attack success rate (ASR \%) on CIFAR-10 for different trigger sizes of BadNets. Our ASD can work well under different trigger sizes of BadNets.}
\label{trigger size of BadNets}
\footnotesize
\setlength{\tabcolsep}{3.5mm}{
\begin{tabular}{clllll}
\toprule[0.68pt]
\addlinespace[0pt]
\multicolumn{2}{c}{\multirow{1}{*}{Trigger size}} & $1\times 1$ & $2\times 2$ & $3\times 3$ & $4\times 4$   \\ \hline
No    & ACC  &  94.7  &  94.9  &  94.2  &  93.7     \\ \cline{2-6} 
                     Defense      & ASR  &  93.4  &  100  &   99.7  &  99.8       \\ \hline
ASD    & ACC  &  93.2  &   93.4  &  92.7  &  93.1   \\ \cline{2-6} 
                        (Ours)   & ASR  &   0.8  &   1.2  &  2.2  &  1.4   \\ 
\addlinespace[-0.22em]
\bottomrule[0.68pt]
\end{tabular}}
\end{table}

\noindent \textbf{Different poisoned rates.} We test our ASD under different poisoned rates $\in \{1\%, 5\%, 10\%, 15\%, 20\%\}$. We demonstrate the results in Table \ref{poisoned rate}, which verifies the superiority of the proposed ASD. 
Besides, as suggested in \cite{turner2018clean}, we also perform the CLB attack under the poisoned rate $\in \{0.6\%, 2.5\%, 5\%\}$ and the maximum perturbation magnitude $\epsilon \in \{16, 32\}$ to evaluate ASD. The results are shown in Table \ref{poisoned rate for clb}, which proves that our ASD can also defend the CLB attack under different attack settings.\par

\noindent \textbf{Different trigger patterns.} For simplicity, we adopt the BadNets on CIFAR-10 as an example to test the performance of our ASD under different trigger patterns. On the one hand, we set the trigger at different locations, as shown in Table \ref{trigger location of BadNets}. On the other hand, we also adjust the trigger size to evaluate our defense, as shown in Table \ref{trigger size of BadNets}. ASD can achieve $92+$\% ACC and $2-$\% ASR in both two cases.\par

\begin{figure}[t]
	\centering
    \begin{subfigure}[]{0.48\linewidth}
        \includegraphics[width=\linewidth]{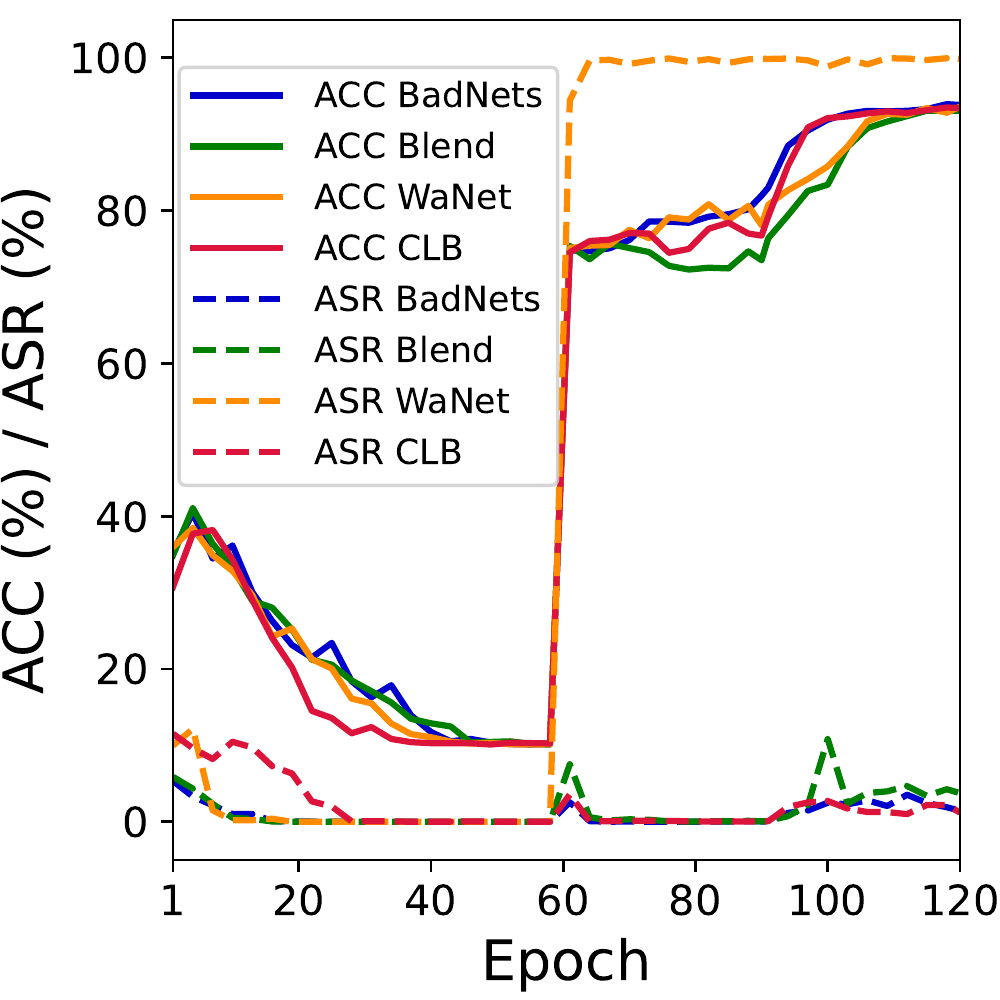}
        \caption{}
        \label{stage1withentireaware}
    \end{subfigure}
    \begin{subfigure}[]{0.48\linewidth}
        \includegraphics[width=\linewidth]{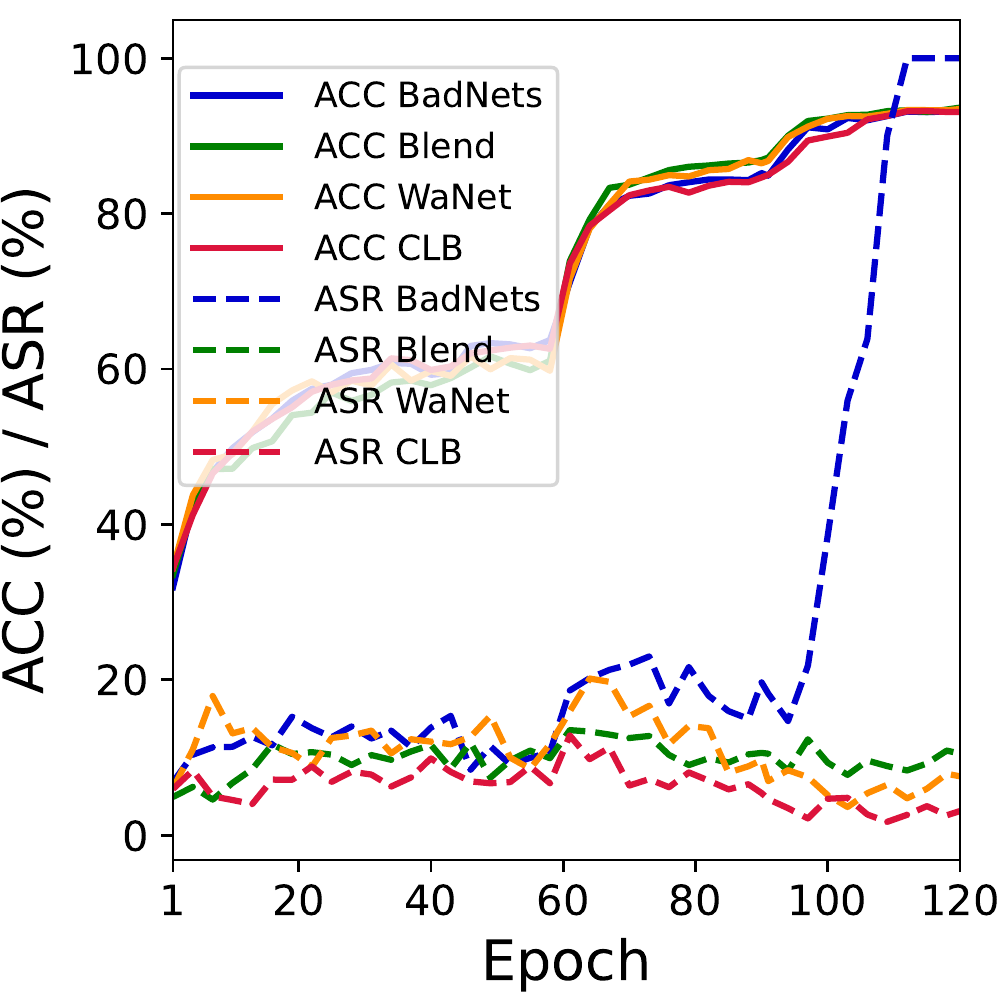}
        \caption{}
        \label{stage2withclassaware}
    \end{subfigure}
	\centering
	\caption{Ablation study for the loss-guided split of our ASD on CIFAR-10 for four backdoor attacks, \textit{i.e.}, BadNets, Blend, WaNet, and CLB. (a) Set class-agnostic loss-guided split in stage 1 instead of class-aware one. (b) Set class-aware loss-guided split in stage 2 instead of class-agnostic one. The results show the necessity of using class-aware one in stage 1 and using class-agnostic one in stage 2.}
\end{figure}

\section{Ablation study on defense settings}
\noindent \textbf{Different modes in loss-guided split.} 
In summary, loss-guided split contains two modes during the previous two stages: (1) Class-aware data split. (2) Class-agnostic data split. In particular, we will discuss the necessity of the corresponding mode in different stages by conducting the followed two experiments. Note that we only change the mode of loss-guided data split and will not change the number in $\mathcal{D}_{C}$ of data split during every epoch. \par
Set class-agnostic loss-guided split in stage 1 instead of class-aware one.  As shown in Fig. \ref{stage1withentireaware}, if we use samples with the lowest losses of the entire set to increase the seed sample, the ACC will crash in stage 1 due to class imbalance.  \par
Set class-aware loss-guided split in stage 2 instead of class-agnostic one. More poisoned samples in the target class will be filled in $\mathcal{D}_{C}$ with class-aware split in stage 2 and ASR can not be suppressed. Fig. \ref{stage2withclassaware} demonstrates that BadNets will break the defense in this setting at $20\%$ poisoned rate. In contrast, our defaulted ASD can still work at 20\% poisoned rate, as shown in Table \ref{poisoned rate}. 

\noindent \textbf{Different hyper-parameters in meta-split.} The epoch, learning rate and updated layer number are three key hyper-parameters in meta-split.\par
\begin{figure*}[t]
    \begin{minipage}[b]{0.7\textwidth}
        \begin{subfigure}[]{0.32\textwidth}
        \includegraphics[width=\textwidth]{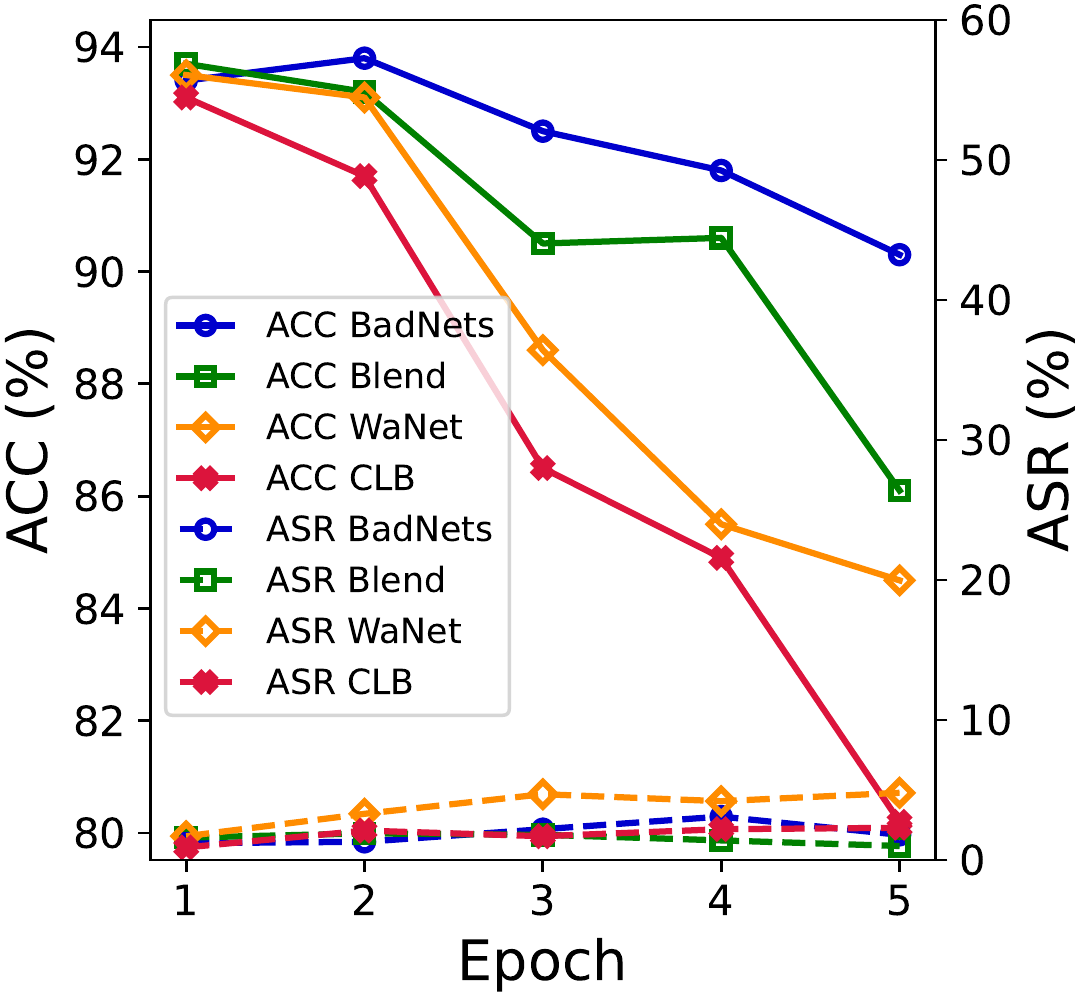}
        \caption{}
        \label{ablEpoch}
    \end{subfigure}
    \begin{subfigure}[]{0.32\textwidth}
        \includegraphics[width=\textwidth]{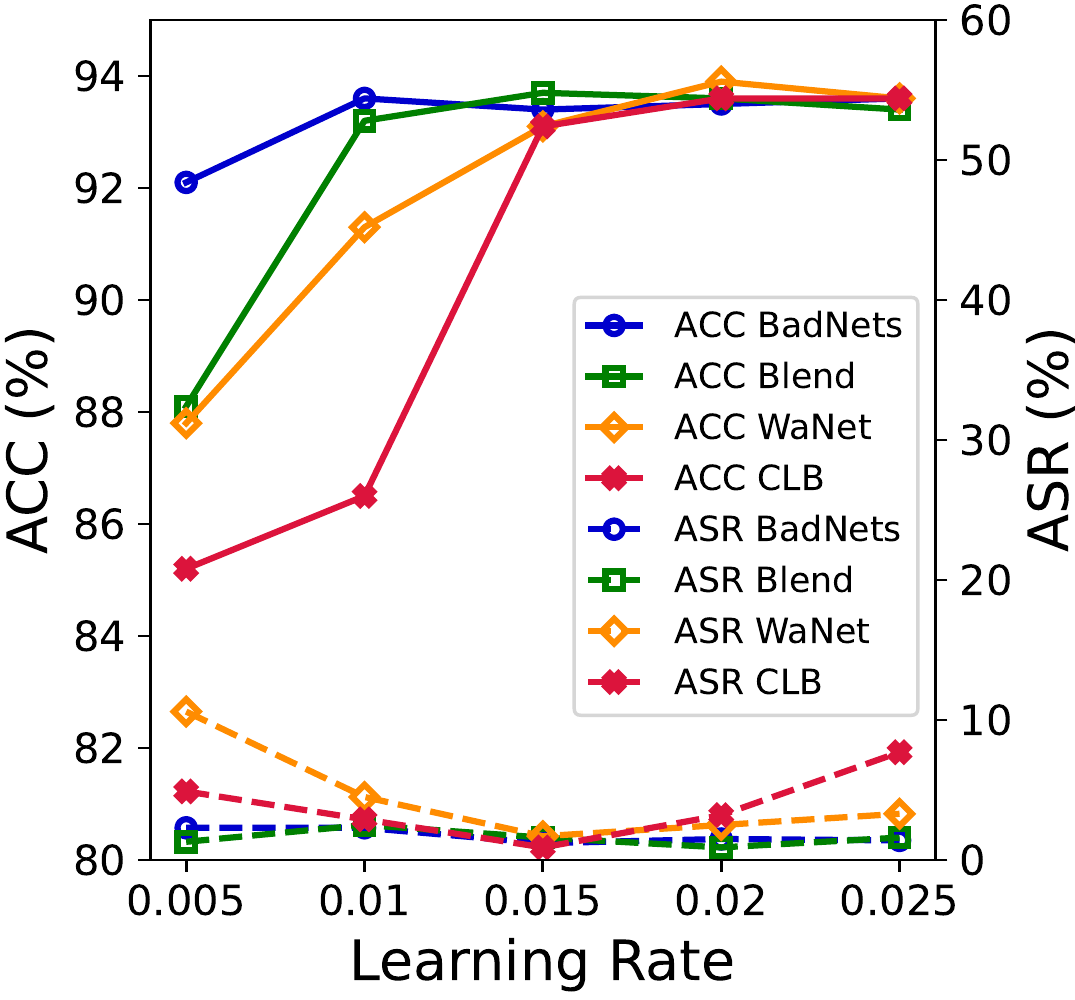}
        \caption{}
        \label{ablLR}
    \end{subfigure}
    \begin{subfigure}[]{0.32\textwidth}
        \includegraphics[width=\textwidth]{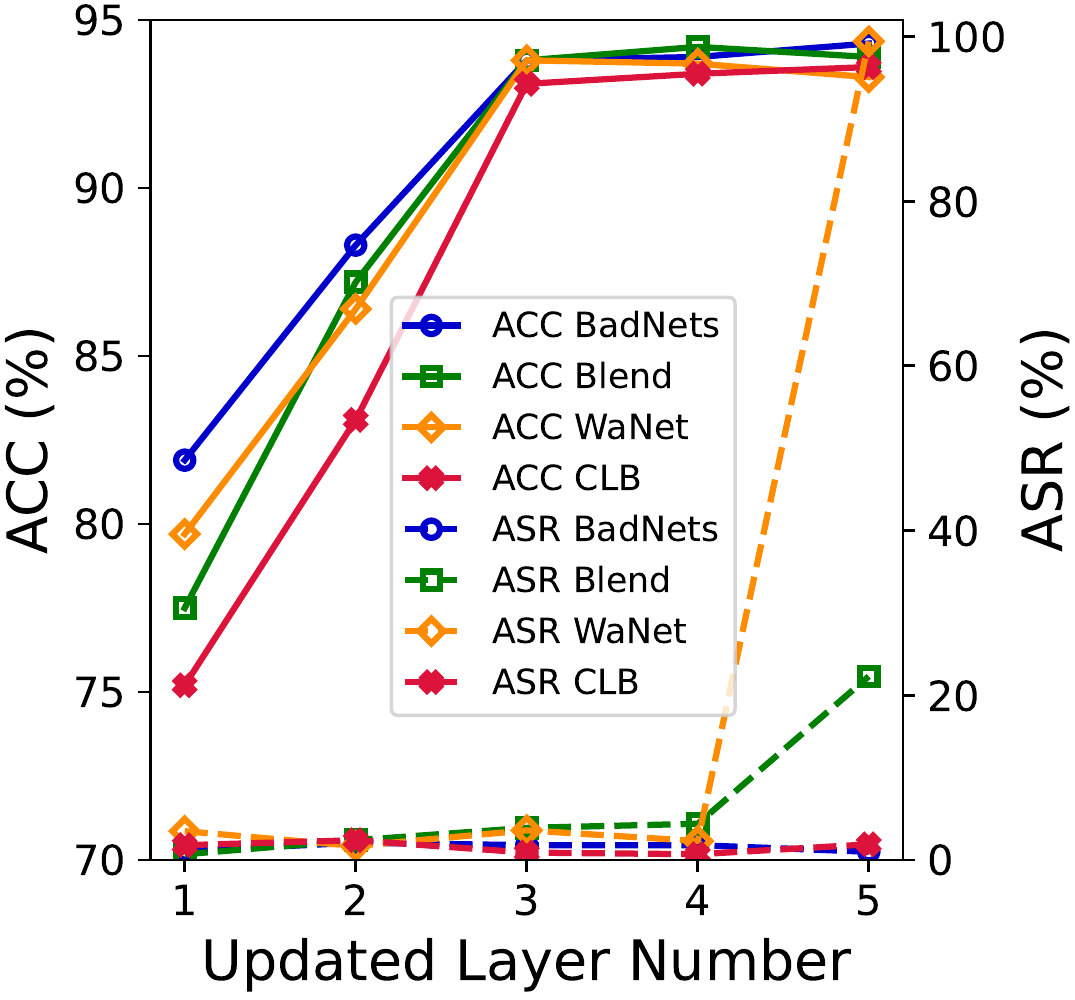}
        \caption{}
        \label{ablLayer}
    \end{subfigure}
	\centering
	\caption{Ablation study for the meta-split on CIFAR-10 for four backdoor attacks, \textit{i.e.}, BadNets, Blend, WaNet, and CLB. (a) The epoch of supervised learning. (b) Learning rate. (c) Updated layer number. }
    \end{minipage}
    \hspace{1em}
	\centering
        \begin{minipage}[b]{0.22\textwidth}
        \includegraphics[width=\textwidth]{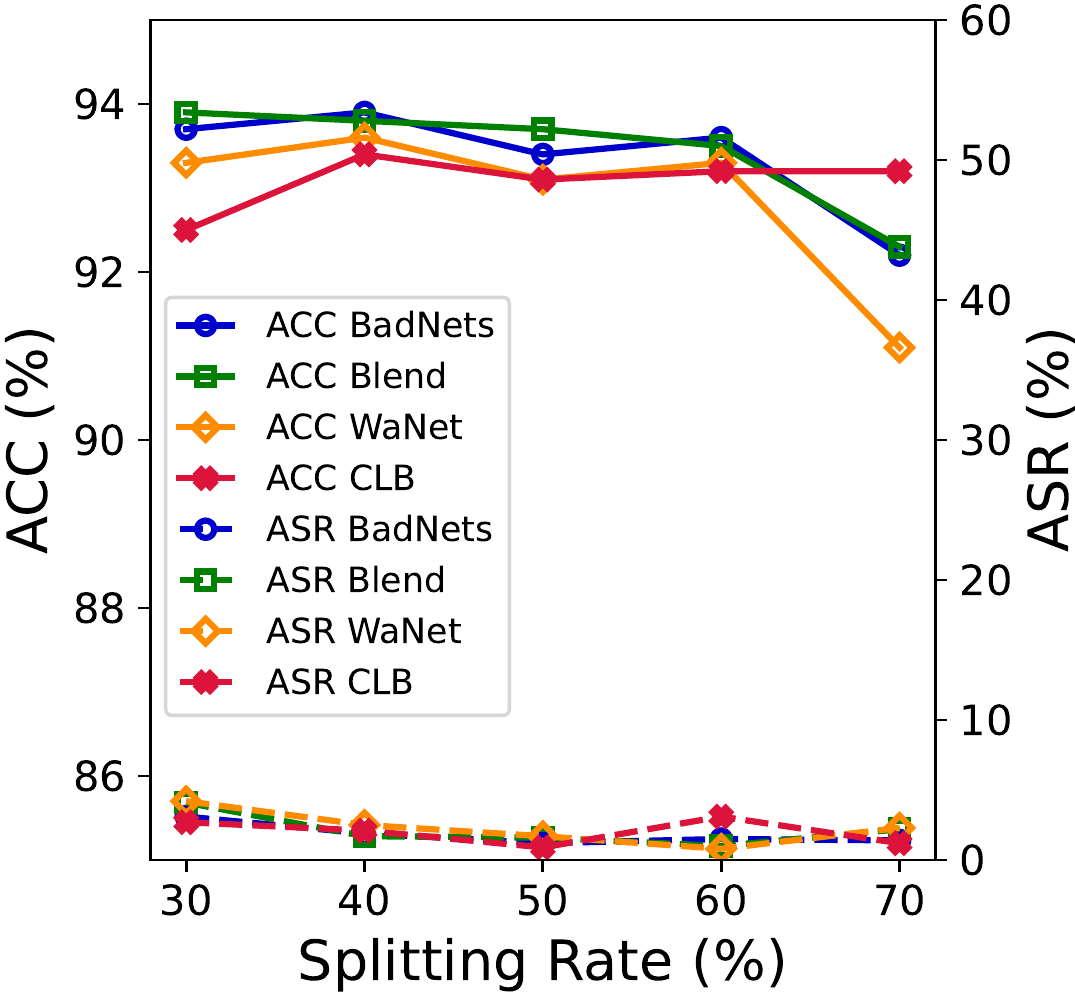}
        \caption{Ablation study for the splitting rate in Stage 2 and Stage 3 on CIFAR-10 for four backdoor attacks.}
        \label{splitting rate}
        \end{minipage}
\end{figure*}

 
From Fig. \ref{ablEpoch}, ACC will drop with the increase of the epoch because the multi-epoch update can enable the model to learn not only the poisoned samples but also the clean hard samples well. As such, the large loss reduction makes the clean hard samples hard to appear in $\mathcal{D}_{C}$  based on the proposed meta-split.\par 
As for the learning rate in Fig. \ref{ablLR}, we can observe that ACC will decrease as the learning rate is smaller. We suspect that the reason may be that the low learning rate can also prevent clean easy samples to be learned by the model. Hence, the small loss reduction makes $\mathcal{D}_C$ contain more clean easy samples. \par
Fig. \ref{ablLayer} indicates that the number of layers to be updated in meta-split can also have an effect on the final performance of the proposed ASD. We show the results of our ASD at poisoned rate $1\%$ in Fig. \ref{ablLayer}. If we update fewer layers in the meta-split, the clean easy samples can not be learned by the model either, which introduces more clean easy samples to 
$\mathcal{D}_C$ and thus leads to a lower ACC. Once all the layers are updated and the poisoned rate is low, the difference in the loss reduction will decrease between the poisoned samples and clean samples. Hence, $\mathcal{D}_{C}$ can contain more poisoned samples and induce the model to create the backdoor mapping.

\noindent \textbf{Different splitting rates 
$\alpha\%$ in stage 2 and $\gamma\%$ stage 3.} We keep the same splitting rate in stage 2 and stage 3 to conduct the experiment. Fig. \ref{splitting rate} shows that our ASD can achieve 90$+\%$ ACC and 5$-\%$ ASR in different splitting rates during stage 2 and stage 3. In other words, ASD is not sensitive to the hyper-parameter splitting rate.

\noindent \textbf{Different model architectures under our ASD.} We test our ASD under ResNet-18 \cite{he2016deep}, VGG-11 \cite{simonyan2014very}, MobileNet \cite{howard2017mobilenets} and DenseNet-121 \cite{huang2017densely}. As suggested in \cite{wu2022backdoorbench}, we conduct the backdoor attacks without backdoor defenses. Besides, we set the learning rate as 0.01 in meta-split for our ASD when using MobileNet. Unless otherwise specified, other settings remain unchanged. As shown in Table \ref{different architecture}, our ASD can defend against backdoor attacks under different model architectures.


\noindent \textbf{Ablation study about seed samples.} Seed samples in each dataset are randomly sampled and then fixed during ASD. Here, we report the results (mean$\pm$std) of 5 runs in Table \ref{seed samples random different}. The results demonstrate the stability of our ASD under different sampled seed samples. Besides, we also conduct the experiments under different numbers of seed samples. As shown in Table \ref{number of clean seed samples}, it might result in the failure of ASD when the number of seed samples is less than 100. Note that 100 is much smaller than that (10,000) required in previous defenses \cite{liu2018fine,li2021neural,wu2021adversarial,wang2022trap}. Besides, we also show seed samples can be taken from a different available dataset in Sec. 5.4, which indicates the flexibility of our seed sample selection.

\noindent \textbf{Performance on clean dataset.} Our ASD can achieve 93.8\% ACC on clean CIFAR10, preserving the clean ACC well.

\begin{table}[t]
\caption{The clean accuracy (ACC \%) and the attack success rate (ASR \%) on CIFAR-10 for different model architectures. Our ASD can work well under different model architectures.}
\label{different architecture}
\footnotesize
\setlength{\tabcolsep}{0.67mm}{

\begin{tabular}{llllllllll}
\toprule[0.68pt]
\addlinespace[0pt]
\multicolumn{1}{l}{\multirow{2}{*}{Attack}} & \multicolumn{1}{l}{\multirow{2}{*}{Method}} & \multicolumn{2}{c}{ResNet18} & \multicolumn{2}{c}{VGG11} & \multicolumn{2}{c}{MobileNet} & \multicolumn{2}{c}{DenseNet121} \\ \cline{3-10} 
\multicolumn{1}{c}{}                        & \multicolumn{1}{c}{}                        & ACC           & ASR          & ACC         & ASR         & ACC            & ASR            & ACC           & ASR           \\ \hline
\multirow{2}{*}{BadNets}                    & No Defense                                  &  94.9          &    100            &        91.0     &   99.9      &    90.1          &   100    &          94.4      &     100                   \\
    & ASD(Ours)                                   & 93.4              &      1.2        &  90.4         &    3.7      &         89.4     &   4.6       &     93.1          &       2.2                     \\ \hline
\multirow{2}{*}{Blend}                      & No Defense                                  &     94.1         &       98.3        &   90.6         &    98.4        &    87.7          &     99.7      &      94.1         &      98.2                  \\
    & ASD(Ours)                                   &      93.7        &   1.6            &      87.5       &     2.4     &      89.8        &     0.7    &      92.4         &        3.1                   \\ \hline
\multirow{2}{*}{WaNet}                    & No Defense                                  &      93.6        &     99.9        &     89.7     &   99.4         &      86.9        &     99.8            &       93.9      &   100            \\
     & ASD(Ours)                                   &        93.1       &       1.7       &       90.3      &     0.9        &         86.8      &    3.4             &          93.2    &    3.2            \\ \hline
\multirow{2}{*}{CLB}                    & No Defense                                  &    94.4         &    99.9          &       91.0   &     99.9         &        88.9     &        7.2        &        94.2      &     2.3         \\
         & ASD(Ours)                                   &       93.1      &   0.9             &     89.2       &    3.1          &     88.1         &   2.4               &       93.1        &   1.4            \\ 
\addlinespace[-0.22em]
\bottomrule[0.68pt]
\end{tabular}}
\end{table}

\begin{table}[t]
\caption{The clean accuracy (ACC \%) and the attack success rate (ASR \%) on CIFAR-10 for different randomly sampled seed samples. The experiments ($\pm$std over 5 random runs) are conducted on CIFAR-10. Our ASD can achieve the stable performance when the seed samples are differently sampled.}
\label{seed samples random different}
\footnotesize
\setlength{\tabcolsep}{2.4mm}{
\begin{tabular}{cllll}
\toprule[0.68pt]
\addlinespace[0pt]
 & BadNets & Blend & WaNet & CLB  \\ \hline
 ACC  &   92.5 ($\pm$0.7)  &   92.9 ($\pm$0.6)    &  93.2 ($\pm$0.6)  &  93.0 ($\pm$0.5) \\  
 ASR  &   1.9 ($\pm$0.6)   &  1.5 ($\pm$0.5)   &   2.0 ($\pm$0.7)  & 2.2 ($\pm$0.8) \\ 

\addlinespace[-0.22em]
\bottomrule[0.68pt]
\end{tabular}}
\end{table}

\begin{table}[t]
\caption{The clean accuracy (ACC \%) and the attack success rate (ASR \%) on CIFAR-10 for different numbers of seed samples.  The default value
(i.e., 100) used in our ASD is feasible on CIFAR-10.}
\label{number of clean seed samples}
\footnotesize
\setlength{\tabcolsep}{5.1mm}{
\begin{tabular}{cllll}
\toprule[0.68pt]
\addlinespace[0pt]
\multicolumn{2}{c}{Number of seed samples} & 10 & 50 & 100  \\ \hline
\multirow{2}{*}{BadNets}   & ACC  &  80.6  &  91.4  & 93.4      \\ \cline{2-5} 
                           & ASR  &  0  &  4.8  &  1.2    \\ \hline
\multirow{2}{*}{Blend}   & ACC  &  92.5  &  86.8  &  93.7   \\ \cline{2-5} 
                           & ASR &  99.1  &  10.4  &  1.6    \\ \hline
\multirow{2}{*}{WaNet}   & ACC &  85.9  &  92.7  &  93.1    \\ \cline{2-5} 
                           & ASR  &  99.7  &  6.2  &  1.7    \\ \hline
\multirow{2}{*}{CLB}     & ACC  &  93.0  &  93.0  & 93.1    \\ \cline{2-5} 
                           & ASR &  3.6  &  2.4  &  0.9   \\ 
\addlinespace[-0.22em]
\bottomrule[0.68pt]
\end{tabular}}
\end{table}

\section{Details about different semi-supervised methods}
Semi-supervised learning \cite{sohn2020fixmatch,berthelot2019mixmatch,xie2020unsupervised,berthelot2019remixmatch,zhu2009introduction} studies how to leverage a training dataset with both labeled data and unlabeled data to obtain a model with high accuracy. 
In addition to its application in normal training, semi-supervised learning also serves as a powerful means for the security of DNNs \cite{alayrac2019labels,carmon2019unlabeled,huang2022backdoor}. 
\par
\noindent \textbf{MixMatch Loss} \cite{berthelot2019mixmatch}. Given a batch $\mathcal{X} \subset \mathcal{D}_{C}$ of labeled samples, and a batch $\mathcal{U} \subset \mathcal{D}_{P}$ of unlabeled samples, MixMatch generates a guessed label distribution $\bar{q}$ for each unlabeled sample $u \in \mathcal{U}$ and adopts MixUp to augment $\mathcal{X}$ and $\mathcal{U}$ to $\mathcal{X}'$ and $\mathcal{U}'$. The supervised loss $\mathcal{L}_s$ is defined as:
\begin{equation}
\mathcal{L}_{s}= \sum_{(x, q) \in \mathcal{X}^{\prime}} \mathrm{H}\left(p_x, q\right),
\end{equation}
where $p_x$ is the prediction of $x$, $q$ is the one-hot label and $\mathrm{H}(\cdot, \cdot)$ is the cross-entropy loss. The unsupervised loss $\mathcal{L}_u$ is defined as:
\begin{equation}
\mathcal{L}_u=\sum_{(u, \bar{q}) \in \mathcal{U}^{\prime}}\left\|p_u-\bar{q}\right\|_2^2,
\end{equation}
where $p_u$ is the prediction of $u$.\par
Finally, the MixMatch loss can be defined as:
\begin{equation}
\mathcal{L}=\mathcal{L}_s+\lambda \cdot \mathcal{L}_u,
\end{equation}
where $\lambda$ is a hyper-parameter for trade-off.
\par

\noindent \textbf{UDA} \cite{xie2020unsupervised}.
Given a batch $\mathcal{X} \subset \mathcal{D}_{C}$ of labeled samples, and a batch $\mathcal{U} \subset \mathcal{D}_{P}$ of unlabeled samples, UDA constructed a guessed label distribution $\bar{q}$ for each unlabeled sample $u \in \mathcal{U}$ after the weak augmentation. Moreover, it adopts the strong augmentation (RandAugment) to augment $\mathcal{U}$ to $\mathcal{U}'$ and generates a guessed label distribution $\bar{q}'$. The supervised loss $\mathcal{L}_s$ is defined as:
\begin{equation}
\mathcal{L}_{s}= \sum_{(x, q) \in \mathcal{X}} \mathrm{H}\left(p_x, q\right),
\end{equation}
where $\mathrm{H}(\cdot, \cdot)$ is the cross-entropy loss. The unsupervised loss $\mathcal{L}_u$ is defined as:
\begin{equation}
\mathcal{L}_u=\sum_{(u, \bar{q}) \in \mathcal{U},(u, \bar{q}') \in \mathcal{U}^{\prime}}\mathrm{H}\left(\bar{q}\ ||\ \bar{q}'\right),
\end{equation}
where $p_u$ is the prediction of $u$.\par
Finally, the UDA loss can be defined as:
\begin{equation}
\mathcal{L}=\mathcal{L}_s+\lambda \cdot \mathcal{L}_u,
\end{equation}
where $\lambda$ is a hyper-parameter for trade-off.

\par
\noindent \textbf{ReMixMatch Loss} \cite{berthelot2019remixmatch}. Given a batch $\mathcal{X} \subset \mathcal{D}_{C}$ of labeled samples, and a batch $\mathcal{U} \subset \mathcal{D}_{P}$ of unlabeled samples, ReMixMatch produces a guessed label distribution $\bar{q}$ for each unlabeled sample $u \in \mathcal{U}$ after the weak augmentation. Besides, it adopts MixUp, the strong augmentation (CTAugment) and the weak augmentation to augment $\mathcal{X}$, $\mathcal{U}$, $\mathcal{U}$ to $\mathcal{X}'$, $\mathcal{U}'$, $\hat{\mathcal{U}}_1$. In total, the ReMixMatch loss can be defined as:
\begin{equation}
\begin{aligned}
\mathcal{L}=
& \sum_{(x, q) \in \mathcal{X}^{\prime}} \mathrm{H}\left(p_x,q\right)
+\lambda_{\mathcal{U}} \sum_{(u, \bar{q}) \in \mathcal{U}^{\prime}} \mathrm{H}\left(p_u,\bar{q}\right)\\
+ & \lambda_{\hat{\mathcal{U}}_1}  \sum_{(u_1, \bar{q}) \in \hat{\mathcal{U}}_1} \mathrm{H}\left(p_{u_1}, \bar{q}\right)\\
+\lambda_r & \sum_{u_1 \in \hat{\mathcal{U}}_1} \mathrm{H}\left(p_{\bm{\theta}}\ (r \mid \operatorname{Rotate}(u_1, r)),r\right),
\end{aligned}
\end{equation}
where $\operatorname{Rotate}(u_1, r)$ denotes that rotate an image $u_1 \in \hat{\mathcal{U}}_1$ the rotation angle $r$ uniformly from $r \sim\{0,90,180,270\}$ and $\mathrm{H}(\cdot, \cdot)$ is the cross-entropy loss.
\par



\section{Details of the adaptive attack}
We state the details of the adaptive attack in the main paper.\par
\noindent \textbf{Problem formulation.} Suppose that the attackers choose a number of samples to be poisoned $\mathcal{D}_p=\{(\bm{x}_i,y_i)\}^N_{i=1}$ and $\mathcal{D}_c=\{(\bm{x}_i,y_i)\}^M_{i=1}$ denotes the remain clean samples, $f_{\bm{\theta}}$ denotes a trained model. 
The objective function for the trigger pattern $\bm{p}$ in the adaptive attacks can be formulated as (\ref{eq:adaptive attack}), \textit{i.e.},  minimizing the gradient for the poisoned samples \textit{w.r.t} the trained model $f_{\bm{\theta}}$ and maximizing that for the clean samples.
\begin{equation}
\begin{aligned}
\min_{\bm{p}} &\frac{1}{N} \sum_{(\bm{x}, y) \in \mathcal{D}_{p}} \frac{\mathrm{d} \mathcal{L}\left(f_{\bm{\theta}}(\bm{x}+\bm{p}), y\right)} {\mathrm{d} \bm{\theta}}\\-&\frac{1}{M} \sum_{(\bm{x}, y) \in \mathcal{D}_{c}} \frac{\mathrm{d} \mathcal{L}\left(f_{\bm{\theta}}(\bm{x}), y\right)} {\mathrm{d} \bm{\theta}}, \textit{s.t.}, \parallel \bm{p} \parallel_{\infty} \leq \epsilon,
\end{aligned}
\label{eq:adaptive attack}
\end{equation}
where $\epsilon$ is the magnitude of the trigger pattern.

\noindent  \textbf{Settings and more results.} We adjust the perturbation magnitudes $\epsilon$ of the trigger pattern for ABL, DBD and our ASD. As shown in Table \ref{adaptive attack perturbation}, our ASD can achieve the best average results among three backdoor defenses. Besides, we also study the effect of the loss objectives to train the surrogate model on the results. Specially, our ASD can obtain $91+$\% ACC and $5-$\% ASR by using either the supervised loss or the semi-supervised loss to train the surrogate model in the adaptive attack.

\begin{table}[t]

\centering
\scriptsize
\caption{The results of ABL, DBD and our ASD under the adaptive attack in different perturbation magnitudes $\epsilon$ of the trigger.}
\label{adaptive attack perturbation}
\setlength{\tabcolsep}{1.65mm}{
\begin{tabular}{l|cccc|c}
	\hline
	$\epsilon$ & 4 & 8 & 16 & 32 & Average \\  \hline 
    ABL & 86.1 / 0.8 & 71.6 / 99.7 & 75.1 / 99.8 & 86.7 / 99.4 & 79.9 / 74.9 \\ 
    DBD & 90.4 / 0.2 & 91.2 / 0.7 & 90.4 / 0.8 & 91.7 / 99.9 & 90.9 / 25.4\\ 
    ASD (Ours) & 93.2 / 1.1 & 93.5 / 1.3 & 92.8 / 0.9 & 93.3 / 1.2 & \textbf{93.2} / \textbf{1.1} \\ \hline
\end{tabular}}
\end{table}

\noindent  \textbf{Reasons for our successful defense against the adaptive attack.} 
The superiority of ASD in adaptive attack benefits a lot from the \textbf{\textit{semi-supervised loss objective}} and \textbf{\textit{two dynamic data pools}}. Adaptive attacks aim at optimizing triggers to minimize the gaps between clean and poisoned samples on surrogate models, which makes poisoned samples difficult to defend. However, such reduced gaps are highly dependent on model checkpoint, which means the gaps might be large on some other checkpoints, especially during ASD training with semi-supervised loss on two dynamic data pools ($D_C$, $D_P$), which can greatly increase the diversity of optimized checkpoints. Besides, as shown in manuscript, ASD is good at separating model checkpoint-dependent clean hard examples from poisoned ones with meta-split. Moreover, the strong data augmentation and pseudo-labeling of MixMatch used in ASD also help destroy the trigger pattern.

\section{Resistance to another adaptive attack}
In this section, we propose another adaptive attack for our proposed ASD. We adopt the same poisoning-based threat model \cite{gu2017badnets,turner2018clean,chen2017targeted} as that in the main paper. \par

\noindent \textbf{Problem formulation.} Suppose that the attackers choose a number of samples to be poisoned $\mathcal{D}_p=\{(\bm{x}_i,y_i)\}^N_{i=1}$ and $\mathcal{D}_t=\{(\bm{x}_i,y_t)\}^M_{i=1}$ denotes the remaining clean samples with the attacker-specified target label $y_t$, $g$ denotes a trained model. Since we adopt semi-supervised learning to purify the polluted pool and this adaptive attack aims to destruct the purification process, the trigger pattern $\bm{p}$ can be optimized by minimizing the distance between poisoned samples and the target class in the feature space as: 
\begin{equation}
\begin{aligned}
\min_{\bm{p}} \Bigg\| \frac{1}{N} \sum_{(\bm{x}, y) \in \mathcal{D}_{p}} &g\left(\bm{x}+\bm{p} \right)- \frac{1}{M} \sum_{(\bm{x}, y) \in \mathcal{D}_{t}} g\left(\bm{x}\right) \Bigg\|_2,\\ \textit{s.t.},&
\parallel \bm{p} \parallel_{\infty} \leq \epsilon,
\end{aligned}
\label{eq:another adaptive attack}
\end{equation}
where $\epsilon$ is the magnitude of the trigger pattern.


\noindent \textbf{Settings and results.} We adopt the same settings as that in our main paper. The adaptive attack can achieve 94.9\% ACC and 99.9\% ASR without any defense. This attack can obtain 93.7\% ACC and only 1.5\% ASR under our ASD. Hence, our ASD can still work well under this adaptive attack due to the low transferability of the trigger pattern.\par


\section{Details about the loss distribution during meta-split}
\begin{figure*}[t]
	\centering
    \begin{subfigure}[]{0.24\linewidth}
        \includegraphics[width=\linewidth]{PDFs/distribution/badnets_91.pdf}
        \caption{BadNets}
    \end{subfigure}
    \begin{subfigure}[]{0.24\linewidth}
        \includegraphics[width=\linewidth]{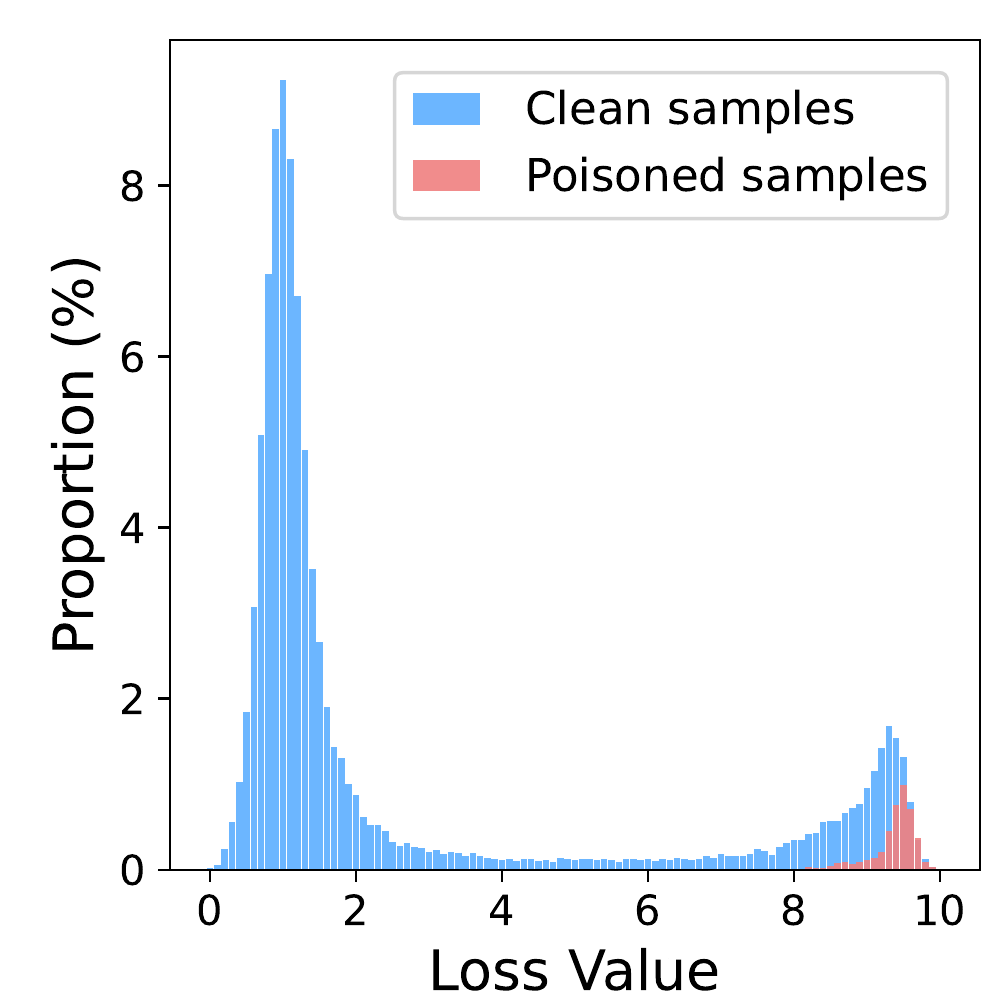}
        \caption{Blend}
    \end{subfigure}
    \begin{subfigure}[]{0.24\linewidth}
        \includegraphics[width=\linewidth]{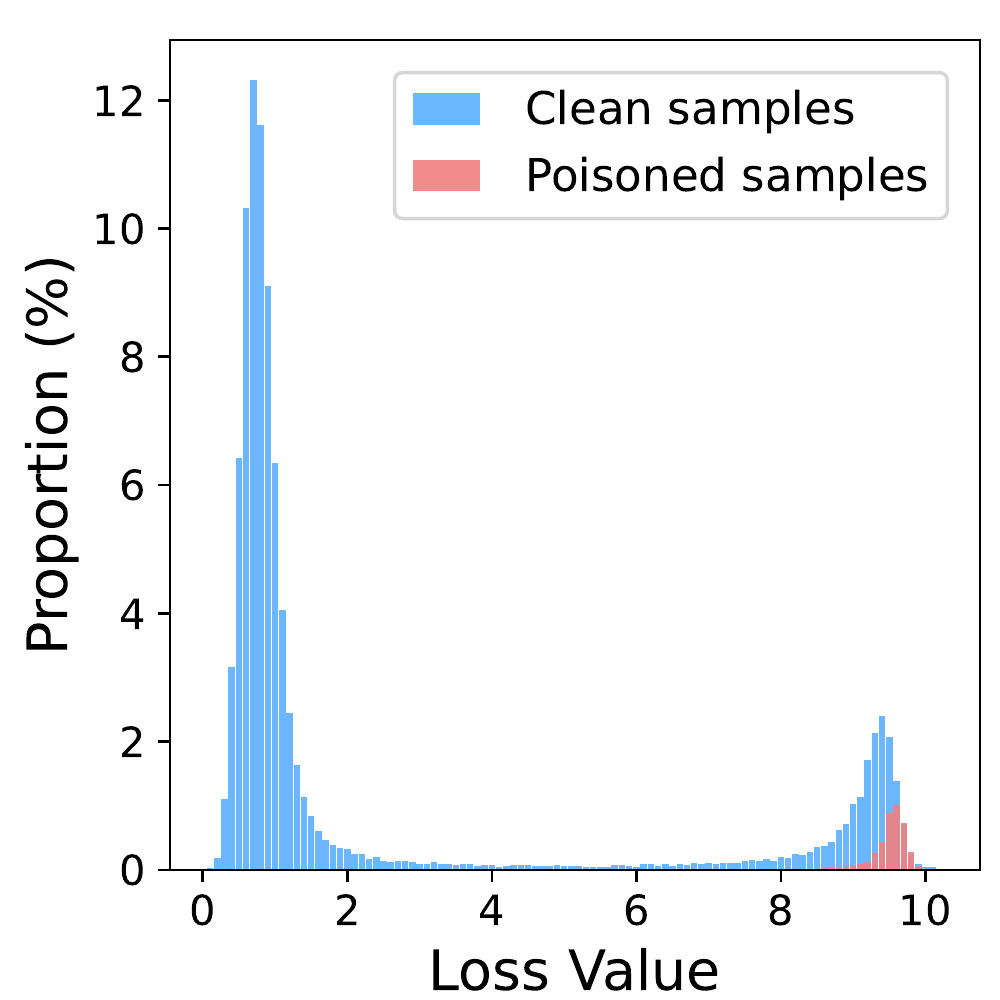}
        \caption{WaNet}
    \end{subfigure}
    \begin{subfigure}[]{0.24\linewidth}
        \includegraphics[width=\linewidth]{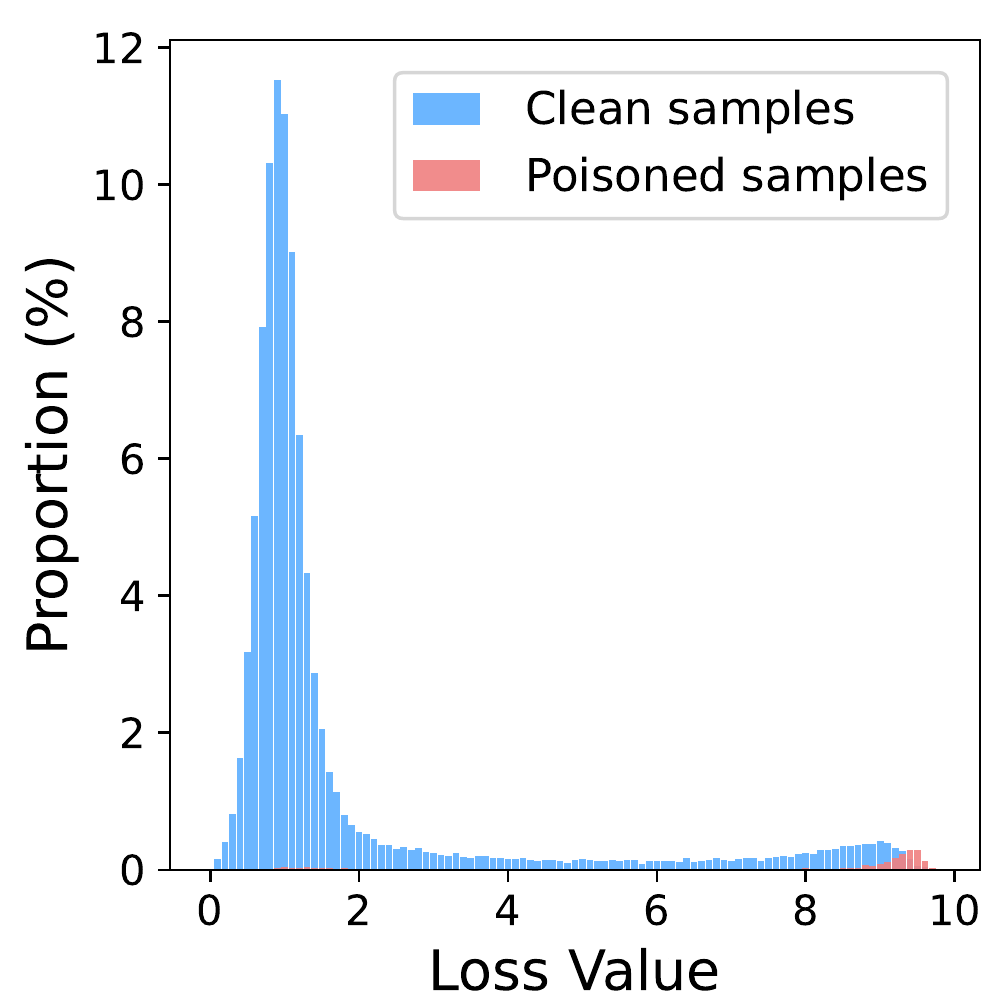}
        \caption{CLB}
    \end{subfigure}
	\centering
	\caption{The loss distribution of samples on the model $f_{\bm{\theta}}$ after the first two stages on CIFAR-10 for four backdoor attacks.}
    \label{distribution at epoch t2}
\end{figure*}

\begin{figure*}[t]
    \vspace{-0.5em}
	\centering
    \begin{subfigure}[]{0.24\linewidth}
        \includegraphics[width=\linewidth]{PDFs/distribution_copy/badnets_91.pdf}
        \caption{BadNets}
    \end{subfigure}
    \begin{subfigure}[]{0.24\linewidth}
        \includegraphics[width=\linewidth]{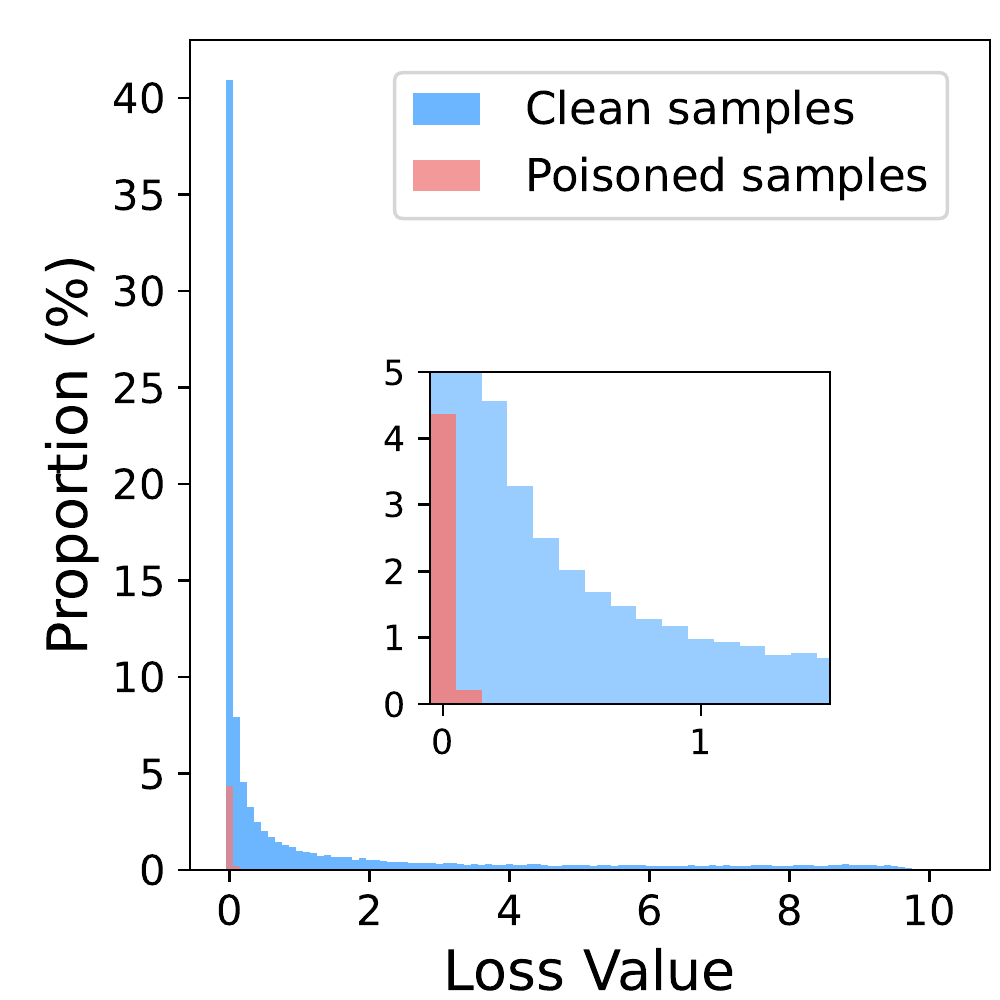}
        \caption{Blend}
    \end{subfigure}
    \begin{subfigure}[]{0.24\linewidth}
        \includegraphics[width=\linewidth]{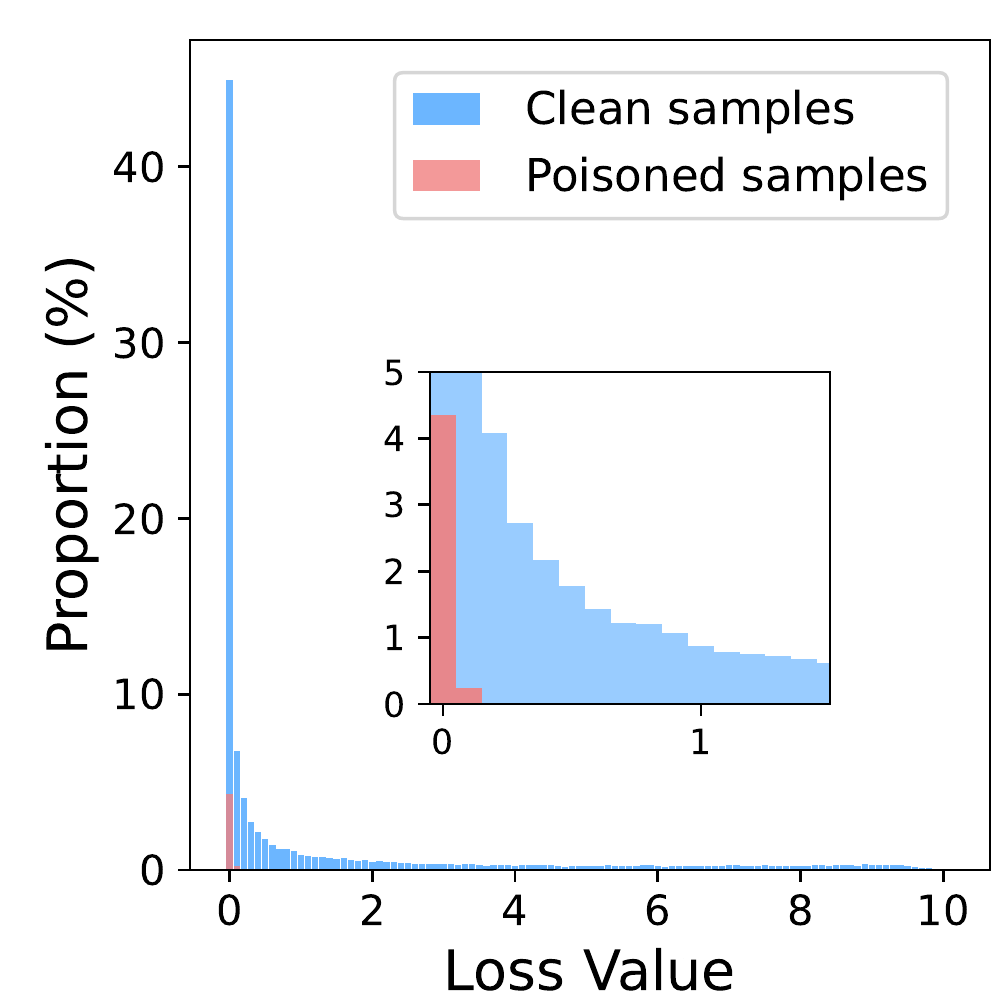}
        \caption{WaNet}
    \end{subfigure}
    \begin{subfigure}[]{0.24\linewidth}
    \includegraphics[width=\linewidth]{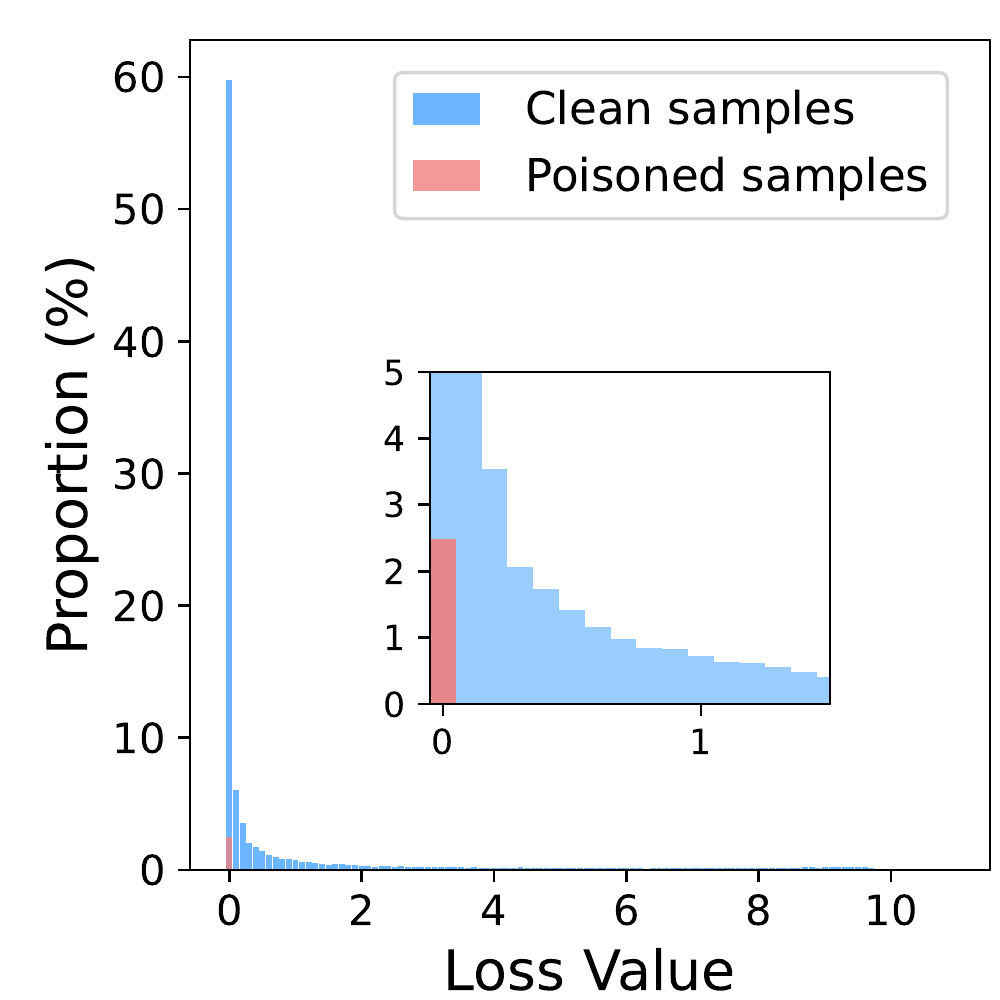}
        \caption{CLB}
    \end{subfigure}
	\centering
	\caption{The loss distribution of samples on the `virtual model' $f_{\bm{\theta}'}$ in Fig. \ref{distribution at epoch t2} after one-epoch supervised learning on CIFAR-10 for four backdoor attacks.}
    \label{loss distribution at epoch t2 after one supervised training}
\end{figure*}

\begin{figure*}[t]
    \vspace{-0.5em}
	\centering
    \begin{subfigure}[]{0.24\linewidth}
        \includegraphics[width=\linewidth]{PDFs/distribution_minus/badnets_91.pdf}
        \caption{BadNets}
    \end{subfigure}
    \begin{subfigure}[]{0.24\linewidth}
        \includegraphics[width=\linewidth]{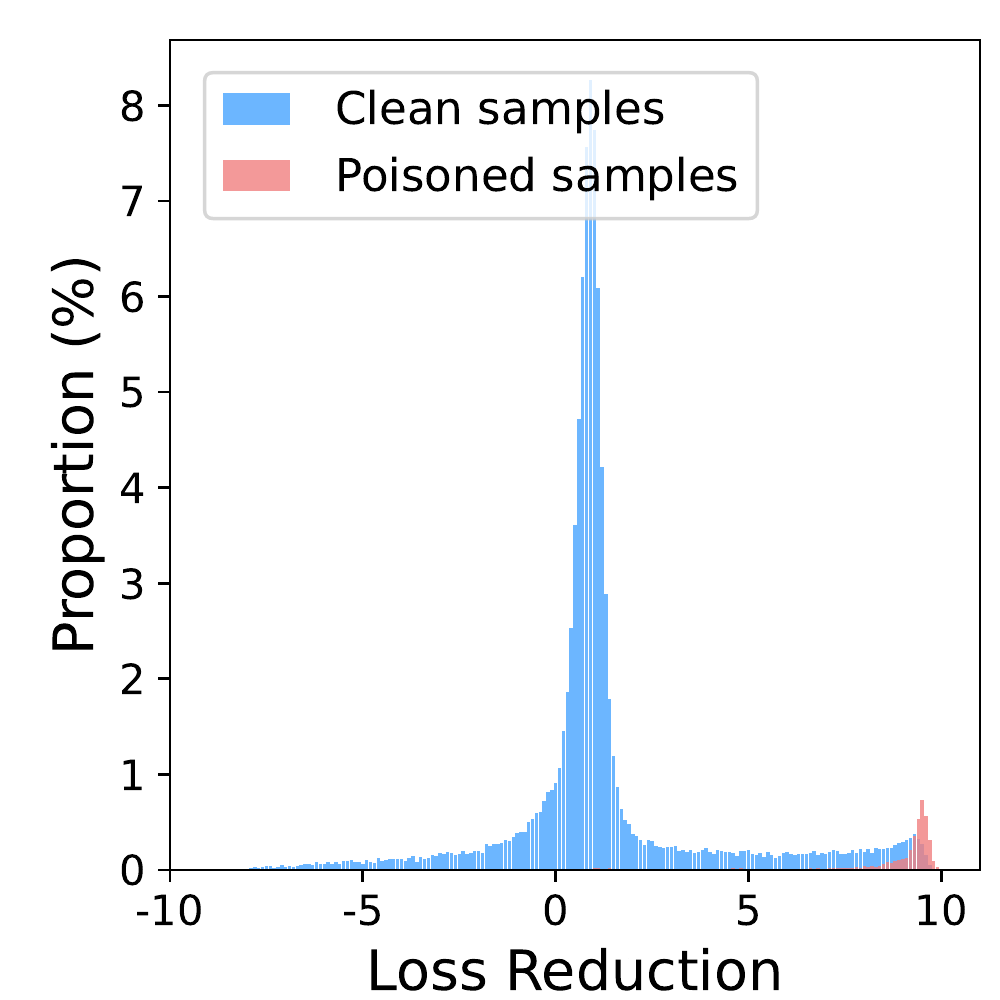}
        \caption{Blend}
    \end{subfigure}
    \begin{subfigure}[]{0.24\linewidth}
        \includegraphics[width=\linewidth]{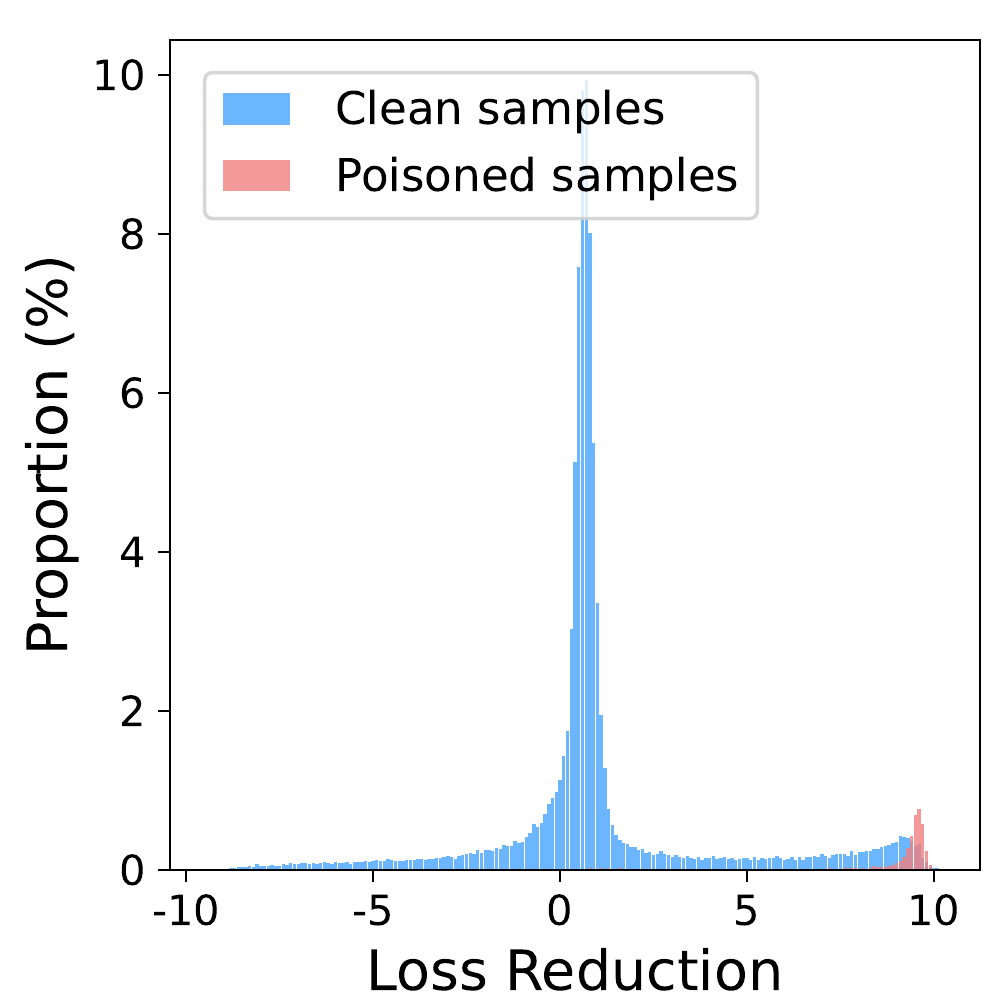}
        \caption{WaNet}
    \end{subfigure}
    \begin{subfigure}[]{0.24\linewidth}
        \includegraphics[width=\linewidth]{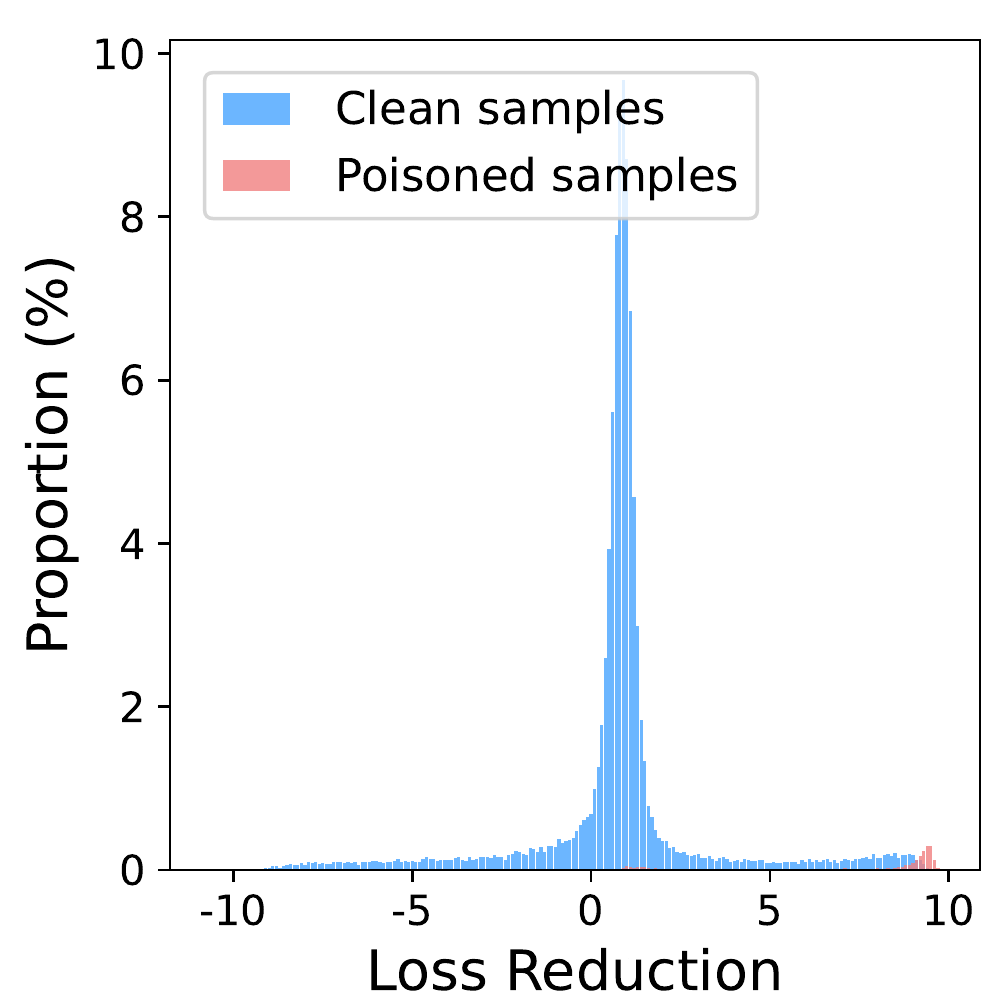}
        \caption{CLB}
    \end{subfigure}
	\centering
	\caption{The loss reduction between $f_{\bm{\theta}}$ in Fig. \ref{distribution at epoch t2} and $f_{\bm{\theta}'}$ in Fig. \ref{loss distribution at epoch t2 after one supervised training} on CIFAR-10 for four backdoor attacks.}
    \label{loss reduction}
\end{figure*}

\begin{figure*}[t]
    \vspace{-0.5em}
	\centering
    \begin{subfigure}[]{0.24\linewidth}
        \includegraphics[width=\linewidth]{PDFs/distribution/badnets_119.pdf}
        \caption{BadNets}
    \end{subfigure}
    \begin{subfigure}[]{0.24\linewidth}
        \includegraphics[width=\linewidth]{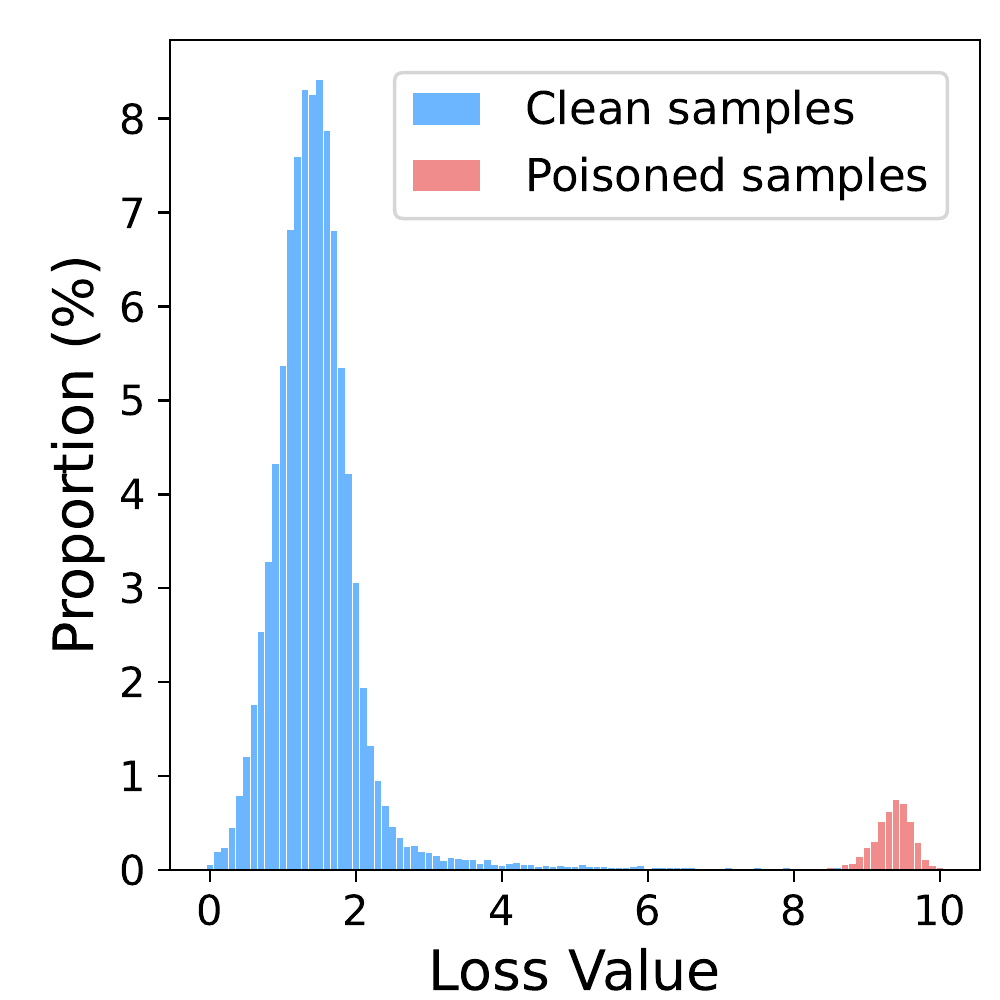}
        \caption{Blend}
    \end{subfigure}
    \begin{subfigure}[]{0.24\linewidth}
        \includegraphics[width=\linewidth]{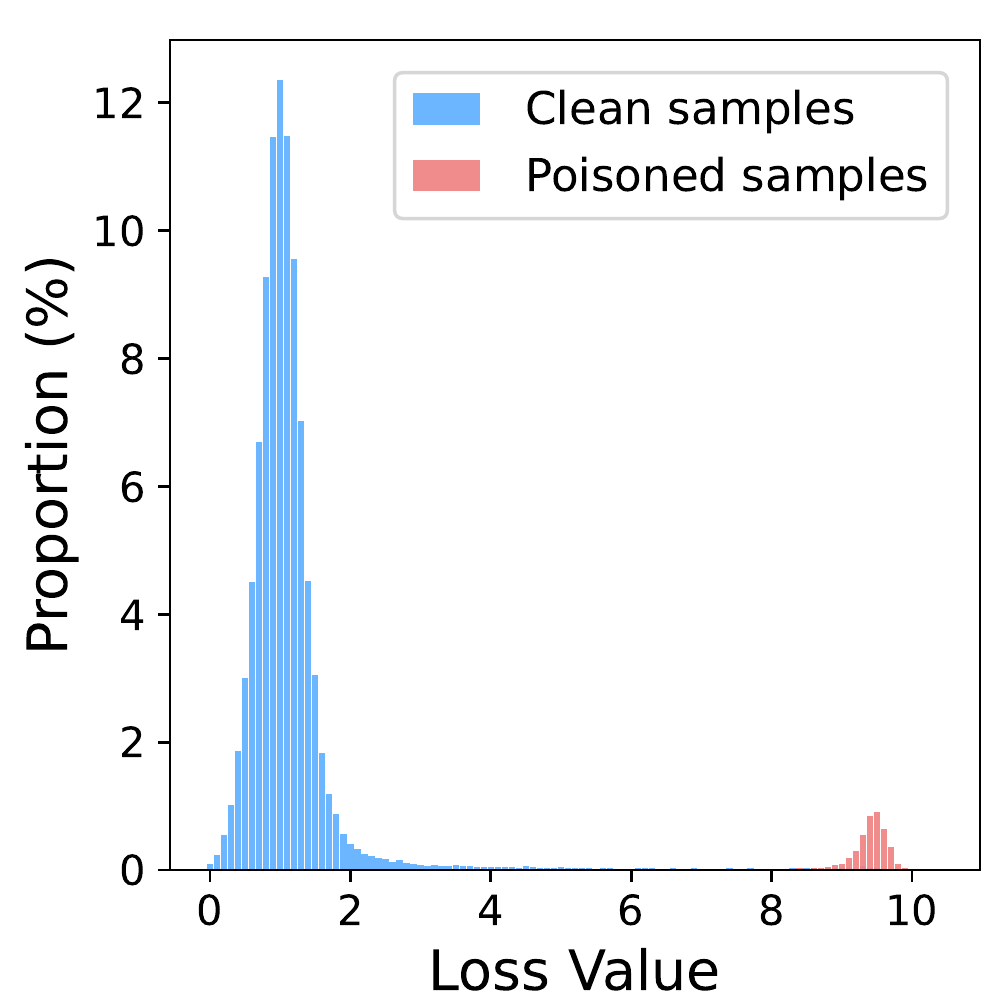}
        \caption{WaNet}
    \end{subfigure}
    \begin{subfigure}[]{0.24\linewidth}
        \includegraphics[width=\linewidth]{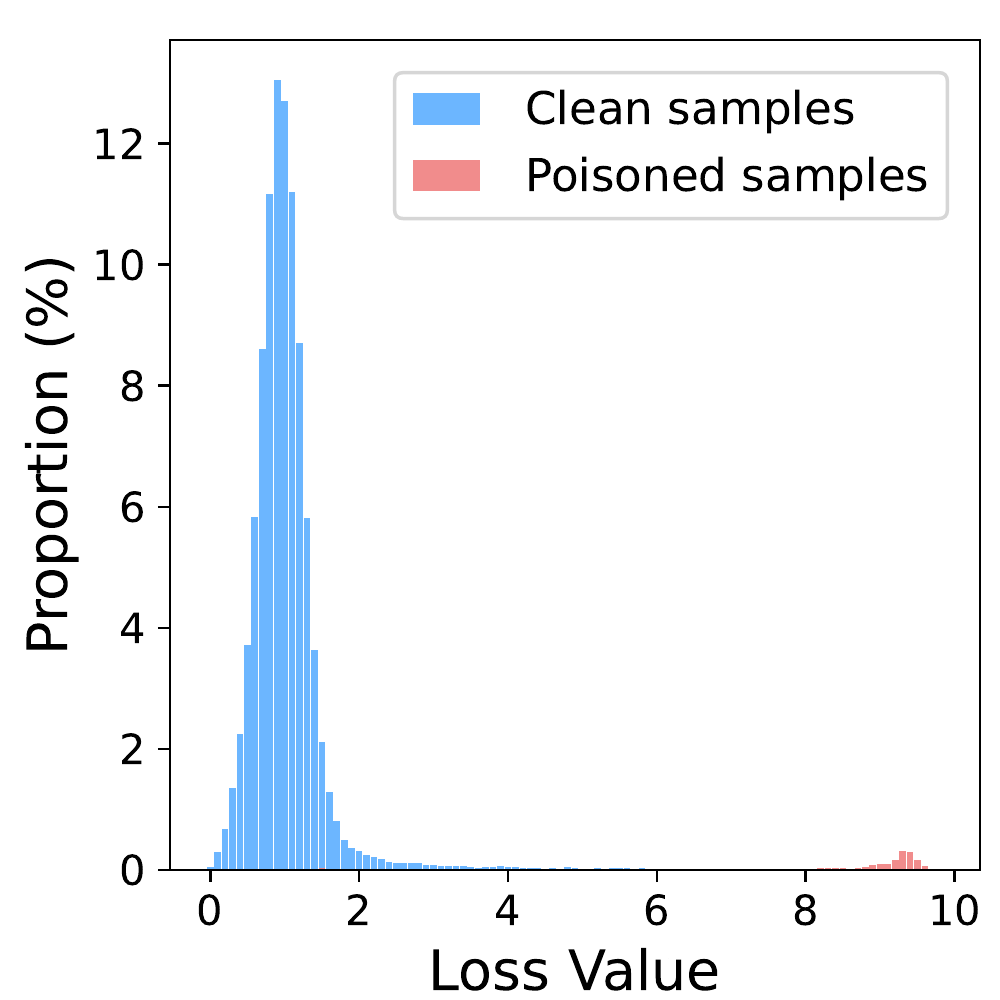}
        \caption{CLB}
    \end{subfigure}
	\centering
	\caption{The loss distribution of samples on the model $f_{\bm{\theta}}$ after all three stages on CIFAR-10 for four backdoor attacks.}
    \label{final distribution}
\end{figure*}

We show more results of the loss distribution during the meta-split of our proposed ASD in Fig. \ref{distribution at epoch t2}, Fig. \ref{loss distribution at epoch t2 after one supervised training}, Fig. \ref{loss reduction} and Fig. \ref{final distribution}. \par

\section{Details about the grid-search for FP, NAD, ABL, and DPSGD}
We search for the best results by grid search for FP, NAD, ABL and DPSGD and show the results in Table \ref{grid search fp cifar}, Table \ref{grid search fp gtsrb}, Table \ref{grid search fp imagenet}, Table \ref{grid search fp vggface2}, Table \ref{grid search nad cifar}, Table \ref{grid search nad gtsrb}, Table \ref{grid search nad imagenet}, Table \ref{grid search nad vggface2}, Table \ref{grid search abl cifar}, Table \ref{grid search abl gtsrb}, Table \ref{grid search abl imagenet}, Table \ref{grid search abl vggface2}, Table \ref{grid search dpsgd cifar}. The details of the grid search have been stated in Appendix \ref{sec:Implementation details}.

\clearpage

\begin{table*}[t]
\caption{Search for the best results by the grid-search for FP on CIFAR-10.}
\label{grid search fp cifar}
\footnotesize
\setlength{\tabcolsep}{3.8mm}{
\begin{tabular}{l|ll|ll|ll|ll|ll|ll}
\toprule[0.68pt]
\addlinespace[0pt]
\multicolumn{1}{c|}{\multirow{2}{*}{Ratio}} & \multicolumn{2}{c|}{BadNets}                    & \multicolumn{2}{c|}{Blend}                            & \multicolumn{2}{c|}{WaNet}                           & \multicolumn{2}{c|}{IAB}                         & \multicolumn{2}{c|}{Refool}                           & \multicolumn{2}{c}{CLB}                                 \\ \cline{2-13} 
\multicolumn{1}{c|}{}                        & \multicolumn{1}{c}{ACC} & \multicolumn{1}{c|}{ASR} & \multicolumn{1}{c}{ACC} & \multicolumn{1}{c|}{ASR} & \multicolumn{1}{c}{ACC} & \multicolumn{1}{c|}{ASR} & \multicolumn{1}{c}{ACC} & \multicolumn{1}{c|}{ASR} & \multicolumn{1}{c}{ACC} & \multicolumn{1}{c|}{ASR} & \multicolumn{1}{c}{ACC} & \multicolumn{1}{c}{ASR} \\ \hline
\multicolumn{1}{c|}{0.2}                 &  94.5 & 100	& 93.9 & 97.9	& 93.4 & 100	& 94.1 & 99.9	& 93.9 & 91.8 & 	\textbf{90.2} & \textbf{92.8}                          \\
\multicolumn{1}{c|}{0.3}                 &  94.4 & 100	& 93.8 & 97.2	& 93.9 & 99.9	& 93.9 & 99.9	& 93.9 & 91.9 &	90.4 & 94.4                                \\
\multicolumn{1}{c|}{0.4}                 &  94.5 & 100 &	93.8 & 96.1	& 93.6 & 99.8	& 93.9 & 99.6	& 93.6 & 92.3 & 	90.3 & 96.9                                \\
\multicolumn{1}{c|}{0.5}                 &  94.6 & 100 & 	93.7 & 96.2	& 93.7 & 99.7	& 93.9 & 100	& 93.2 & 92.5 &	90.4 & 98.8                               \\
\multicolumn{1}{c|}{0.6}                 &  94.4 & 100	& 93.6 & 96.9	& 93.7 & 99.9	& 93.3 & 99.9	& 92.3 & 92.1 &	94.9 & 99.9                              \\
\multicolumn{1}{c|}{0.7}                 &  94.2 & 100 & 	93.4 & 93.2	& 93.3 & 99.8	& 92.9 & 99.8	& 91.2 & 91.8 &	94.4 & 99.8                             \\
\multicolumn{1}{c|}{0.8}                 &  \textbf{93.9} & \textbf{1.8}	 & 93.8 & 89.5& 	93.1 & 99.8	& 92.3 & 99.7	& 92.7 & 87.9 & 	93.3 & 100                              \\
\multicolumn{1}{c|}{0.9}                 &  94.2 & 100	 & \textbf{92.9} & \textbf{77.1}	& \textbf{90.4} & \textbf{98.6}	& \textbf{89.3} & \textbf{98.1}	& \textbf{92.1} & \textbf{86.1}	& 91.17 & 99.9                              \\
\addlinespace[-0.22em]
\bottomrule[0.68pt]
\end{tabular}}
\end{table*}

\begin{table*}[t]
\caption{Search for the best results by the grid-search for FP on GTSRB.}
\label{grid search fp gtsrb}
\footnotesize
\setlength{\tabcolsep}{3.8mm}{
\begin{tabular}{l|ll|ll|ll|ll|ll|ll}
\toprule[0.68pt]
\addlinespace[0pt]
\multicolumn{1}{c|}{\multirow{2}{*}{Ratio}} & \multicolumn{2}{c|}{BadNets}                    & \multicolumn{2}{c|}{Blend}                            & \multicolumn{2}{c|}{WaNet}                           & \multicolumn{2}{c|}{IAB}                         & \multicolumn{2}{c|}{Refool}                           & \multicolumn{2}{c}{CLB}                                 \\ \cline{2-13} 
\multicolumn{1}{c|}{}                        & \multicolumn{1}{c}{ACC} & \multicolumn{1}{c|}{ASR} & \multicolumn{1}{c}{ACC} & \multicolumn{1}{c|}{ASR} & \multicolumn{1}{c}{ACC} & \multicolumn{1}{c|}{ASR} & \multicolumn{1}{c}{ACC} & \multicolumn{1}{c|}{ASR} & \multicolumn{1}{c}{ACC} & \multicolumn{1}{c|}{ASR} & \multicolumn{1}{c}{ACC} & \multicolumn{1}{c}{ASR} \\ \hline
\multicolumn{1}{c|}{0.2}                 &  97.4 & 100 & 	96.8 & 98.7 & 	97.6 & 100	& 97.6 & 100	& 97.9 & 98.2 & 	94.5 & 99.7                               \\
\multicolumn{1}{c|}{0.3}                 &  97.2 & 100	& 96.7 & 98.6	& 97.2 & 100	& 97.3 & 100	& 98.1 & 96.4 &	\textbf{93.6} & \textbf{99.3}                                \\
\multicolumn{1}{c|}{0.4}                 &  97.3 & 100 &	96.7 & 98.5	& 97.2 & 100	& 97.7 & 100	& 97.9 & 94.2 & 	89.3 & 99.4                                \\
\multicolumn{1}{c|}{0.5}                 &  97.1 & 100	& 96.5 & 98.2	& 97.4 & 100	& 97.5 & 100	& 97.3 & 60.9 &	83.2 & 99.5                               \\
\multicolumn{1}{c|}{0.6}                 &  97.1 & 100	 & 96.5 & 98.1	& 97.2 & 99.9	& 97.5 & 100	& 96.6 & 49.1 &	67.6 & 99.7                                \\
\multicolumn{1}{c|}{0.7}                 &  96.5 & 99.9 & 	96.1 & 85.2	& 96.3 & 99.9	& 97.4 & 99.5	& 95.7 & 47.3 & 	52.6 & 99.2                              \\
\multicolumn{1}{c|}{0.8}                 &  93.4 & 73.7 & 	\textbf{91.4} & \textbf{68.1}	& \textbf{92.5} & \textbf{21.4}	& 96.4 & 99.2	& \textbf{91.5} & \textbf{0.2}	 & 39.4 & 99.6                             \\
\multicolumn{1}{c|}{0.9}                 &  \textbf{84.2} & 
\textbf{0}	& 78.8 & 85.7	& 85.3 & 23.6	 & \textbf{86.9} & \textbf{0}	& 87.3 & 0.4	 & 23.6 & 99.4                              \\
\addlinespace[-0.22em]
\bottomrule[0.68pt]
\end{tabular}}
\end{table*}

\begin{table*}[t]
\caption{Search for the best results by the grid-search for FP on ImageNet.}
\label{grid search fp imagenet}
\footnotesize
\setlength{\tabcolsep}{3.8mm}{
\begin{tabular}{l|ll|ll|ll|ll|ll|ll}
\toprule[0.68pt]
\addlinespace[0pt]
\multicolumn{1}{c|}{\multirow{2}{*}{Ratio}} & \multicolumn{2}{c|}{BadNets}                    & \multicolumn{2}{c|}{Blend}                            & \multicolumn{2}{c|}{WaNet}                           & \multicolumn{2}{c|}{IAB}                         & \multicolumn{2}{c|}{Refool}                           & \multicolumn{2}{c}{CLB}                                 \\ \cline{2-13} 
\multicolumn{1}{c|}{}                        & \multicolumn{1}{c}{ACC} & \multicolumn{1}{c|}{ASR} & \multicolumn{1}{c}{ACC} & \multicolumn{1}{c|}{ASR} & \multicolumn{1}{c}{ACC} & \multicolumn{1}{c|}{ASR} & \multicolumn{1}{c}{ACC} & \multicolumn{1}{c|}{ASR} & \multicolumn{1}{c}{ACC} & \multicolumn{1}{c|}{ASR} & \multicolumn{1}{c}{ACC} & \multicolumn{1}{c}{ASR} \\ \hline
\multicolumn{1}{c|}{0.2}                 &  76.7 & 51.1 &	78.1 & 93.9	& 79.1 & 95.9	& 78.9 & 99.4	& 77.7 & 84.4 &	79.9 & 88.3                                \\
\multicolumn{1}{c|}{0.3}                 &  75.9 & 46.1 & 	77.6 & 93.4	& 78.9 & 96.3	& 78.3 & 99.9	& 76.7 & 74.1 & 	79.1 & 69.2                                \\
\multicolumn{1}{c|}{0.4}                 &  73.9 & 20.7 & 	76.1 & 90.9	& 77.5 & 95.5	& 76.9 & 98.9	 & 75.9 & 75.5 & 	77.6 & 60.1                                \\
\multicolumn{1}{c|}{0.5}                 &  \textbf{70.3} & \textbf{1.6} & 	74.4 & 86.2	& 74.2 & 94.4	& 76.3 & 99.9	& 73.8 & 62.7 & 75.7 & 67.2                              \\
\multicolumn{1}{c|}{0.6}                 &  73.1 & 14.9 & 	72.7 & 85.9	& 72.5 & 95.2	& 72.8 & 98.2	& 70.5 & 48.3 & 	\textbf{73.2} & \textbf{38.3}                                \\
\multicolumn{1}{c|}{0.7}                 &  70.1 & 0.3	& 71.1 & 87.8	& 75.1 & 95.1	& 73.9 & 99.9	& 73.6 & 63.8	& 69.8 & 49.5                              \\
\multicolumn{1}{c|}{0.8}                 &  66.4 & 0	& 72.9 & 77.9	& 71.4 & 92.2	& 70.9 & 99.4	& 71.2 & 52.5 &	70.9 & 53.7                              \\
\multicolumn{1}{c|}{0.9}                 &  84.3 & 0	& \textbf{63.4} & \textbf{9.5}	& \textbf{58.2} & \textbf{84.4}	& \textbf{58.7} & \textbf{84.2}	& \textbf{61.4} & \textbf{10.3} & 	54.2 & 0                             \\
\addlinespace[-0.22em]
\bottomrule[0.68pt]
\end{tabular}}
\end{table*}

\begin{table*}[t]

\caption{Search for the best results by the grid-search for FP on VGGFace2.}
\label{grid search fp vggface2}
\footnotesize
\setlength{\tabcolsep}{3.8mm}{
\begin{tabular}{l|ll|ll|ll|ll|ll|ll}
\toprule[0.68pt]
\addlinespace[0pt]
\multicolumn{1}{c|}{\multirow{2}{*}{Ratio}} & \multicolumn{2}{c|}{BadNets}                    & \multicolumn{2}{c|}{Blend}                            & \multicolumn{2}{c|}{WaNet}                           & \multicolumn{2}{c|}{IAB}                         & \multicolumn{2}{c|}{Refool}                           & \multicolumn{2}{c}{CLB}                                 \\ \cline{2-13} 
\multicolumn{1}{c|}{}                        & \multicolumn{1}{c}{ACC} & \multicolumn{1}{c|}{ASR} & \multicolumn{1}{c}{ACC} & \multicolumn{1}{c|}{ASR} & \multicolumn{1}{c}{ACC} & \multicolumn{1}{c|}{ASR} & \multicolumn{1}{c}{ACC} & \multicolumn{1}{c|}{ASR} & \multicolumn{1}{c}{ACC} & \multicolumn{1}{c|}{ASR} & \multicolumn{1}{c}{ACC} & \multicolumn{1}{c}{ASR} \\ \hline
\multicolumn{1}{c|}{0.2}                 &  91.2 & 100 & 90.2 & 99.9 & 91.8 & 75.3 & 90.7 & 97.5 &	90.7 & 98.6 & 90.8 & 99.8                               \\
\multicolumn{1}{c|}{0.3}                 &  91.1 & 100 & 90.1 & 99.9 & 91.8 & 75.7 & 90.8 & 97.4 &	90.8 & 98.7 & \textbf{90.9} & \textbf{99.9}                              \\
\multicolumn{1}{c|}{0.4}                 &  91.1 & 100 & 90.2 & 99.9 & 91.7 & 75.8 & 90.7 & 97.3 &	90.7 & 98.6 & 90.8 & 99.9                             \\
\multicolumn{1}{c|}{0.5}                 &  \textbf{91.5} & \textbf{100} & 90.2 & 99.9 & 91.6 & 76.7 & 90.6 & 97.3 & 90.9 & 98.7 & 90.8 & 99.8                             \\
\multicolumn{1}{c|}{0.6}                 &  91.0 & 100 & 91.1 & 99.9 & 91.6 & 78.5 & \textbf{90.6} & \textbf{97.2} &	\textbf{90.4} & \textbf{98.4}	& 90.6 & 99.9                               \\
\multicolumn{1}{c|}{0.7}                 &  91.1 & 100 & 90.3 & 100 & 91.4 & 81.1 & 90.5 & 97.6 &	90.7 & 98.9 & 90.7 & 99.9                             \\
\multicolumn{1}{c|}{0.8}                 &  91.5 & 100 & 89.9 & 100 & 90.8 & 79.6 & 90.6 & 97.4 &	90.2 & 97.9 & 90.2 & 100                         \\
\multicolumn{1}{c|}{0.9}                 &  89.3 & 100 & \textbf{87.1} & \textbf{96.0} & \textbf{89.2} & \textbf{33.4} & 89.3 & 98.4	& 86.8 & 98.5 & 88.8 & 99.9                           \\
\addlinespace[-0.22em]
\bottomrule[0.68pt]
\end{tabular}}
\end{table*}

\begin{table*}[t]

\caption{Search for the best results by the grid-search for NAD on CIFAR-10.}
\label{grid search nad cifar}
\footnotesize
\setlength{\tabcolsep}{3.8mm}{
\begin{tabular}{l|ll|ll|ll|ll|ll|ll}
\toprule[0.68pt]
\addlinespace[0pt]
\multicolumn{1}{c|}{\multirow{2}{*}{$\beta$}} & \multicolumn{2}{c|}{BadNets}                    & \multicolumn{2}{c|}{Blend}                            & \multicolumn{2}{c|}{WaNet}                           & \multicolumn{2}{c|}{IAB}                         & \multicolumn{2}{c|}{Refool}                           & \multicolumn{2}{c}{CLB}                                 \\ \cline{2-13} 
\multicolumn{1}{c|}{}                        & \multicolumn{1}{c}{ACC} & \multicolumn{1}{c|}{ASR} & \multicolumn{1}{c}{ACC} & \multicolumn{1}{c|}{ASR} & \multicolumn{1}{c}{ACC} & \multicolumn{1}{c|}{ASR} & \multicolumn{1}{c}{ACC} & \multicolumn{1}{c|}{ASR} & \multicolumn{1}{c}{ACC} & \multicolumn{1}{c|}{ASR} & \multicolumn{1}{c}{ACC} & \multicolumn{1}{c}{ASR} \\ \hline
\multicolumn{1}{c|}{500}                 &  90.6 & 12.9                           &  89.8 & 17.2                           &  89.2 & 21.8                           &  88.5 & 29.7                             &  89.9 & 9.7                           &  \textbf{86.4} & \textbf{9.5}                                \\
\multicolumn{1}{c|}{1000}                 &  89.7 & 10.0                           &  87.4 & 4.3                           &  87.5 & 11.8                            &  85.8 & 8.3                             &  89.7 & 10.5                           &  81.8 & 8.6                                \\
\multicolumn{1}{c|}{1500}                 &  \textbf{88.2} & \textbf{4.6}                            &  \textbf{85.8} & \textbf{3.4}                           & 83.1 & 13.1                            &  \textbf{82.8} & \textbf{4.2}                             &  87.7 & 5.4                           &  72.5 & 6.2                                \\
\multicolumn{1}{c|}{2000}                 &  84.7 & 6.9                            &  80.2 & 5.9                           &  71.3 & 6.7                            &  75.5 & 2.5                             &  \textbf{86.2} & \textbf{3.6}                           &  65.3 & 6.3                               \\
\multicolumn{1}{c|}{2500}                 &  83.1 & 4.5                            &  75.8 & 4.5                           &  64.3 & 8.1                            &  67.9 & 1.1                             &  81.1 & 3.1                           &  43.9 & 10.9                                \\
\multicolumn{1}{c|}{5000}                 &  32.2 & 2.7	& 32.1 & 3.7	& 40.2 & 6.3	& 39.4 & 7.2 &	45.5 & 2.8	& 32.3 & 5.1                              \\
\multicolumn{1}{c|}{7500}                 &  18.2 & 5.1	& 29.8 & 3.1	& 25.7 & 4.1	& 24.5 & 1.9	& 30.3 & 8.1 & 18.4 & 11.1                              \\
\multicolumn{1}{c|}{10000}                 &  20.8 & 1.1	& 20.2 & 7.4	 & 23.9 & 10.4	& 20.1 & 6.1 & 	24.1 & 6.2	& 21.4 & 14.4                              \\
\addlinespace[-0.22em]
\bottomrule[0.68pt]
\end{tabular}}
\end{table*}

\begin{table*}[t]
\caption{Search for the best results by the grid-search for NAD on GTSRB.}
\label{grid search nad gtsrb}
\footnotesize
\setlength{\tabcolsep}{3.8mm}{
\begin{tabular}{l|ll|ll|ll|ll|ll|ll}
\toprule[0.68pt]
\addlinespace[0pt]
\multicolumn{1}{c|}{\multirow{2}{*}{$\beta$}} & \multicolumn{2}{c|}{BadNets}                    & \multicolumn{2}{c|}{Blend}                            & \multicolumn{2}{c|}{WaNet}                           & \multicolumn{2}{c|}{IAB}                         & \multicolumn{2}{c|}{Refool}                           & \multicolumn{2}{c}{CLB}                                 \\ \cline{2-13} 
\multicolumn{1}{c|}{}                        & \multicolumn{1}{c}{ACC} & \multicolumn{1}{c|}{ASR} & \multicolumn{1}{c}{ACC} & \multicolumn{1}{c|}{ASR} & \multicolumn{1}{c}{ACC} & \multicolumn{1}{c|}{ASR} & \multicolumn{1}{c}{ACC} & \multicolumn{1}{c|}{ASR} & \multicolumn{1}{c}{ACC} & \multicolumn{1}{c|}{ASR} & \multicolumn{1}{c}{ACC} & \multicolumn{1}{c}{ASR} \\ \hline
\multicolumn{1}{c|}{500}                 &  \textbf{97.1}  & \textbf{0.2} &	96.9 & 99.9 &	97.2 & 67.8	& 96.9 & 0.1	& 97.3 & 93.6	& 5.7 & 40.1                               \\
\multicolumn{1}{c|}{1000}                 &  96.8 & 0 & 	96.8 & 99.5 & 	97.1 & 69.1	 & \textbf{97.1} & \textbf{0.1}	& 97.3 & 72.4	& 4.1 & 41.7                            \\
\multicolumn{1}{c|}{1500}                 &  96.5 & 0	& 96.3 & 99.9	& 96.9 & 76.1	& 95.9 & 0.7	& 97.1 & 47.5 & 4.7 & 44.6                              \\
\multicolumn{1}{c|}{2000}                 &  93.5 & 0	& 96.2 & 99.3	& 96.7 & 70.9	& 94.5 & 0.5 &	\textbf{95.5} & \textbf{1.4}	& 4.8 & 40.1                             \\
\multicolumn{1}{c|}{2500}                 &  19.7 & 0	& 96.2 & 99.1	& \textbf{96.5} & \textbf{47.1}	& 20.8 & 0	& 93.6 & 3.8	& 5.5 & 36.3                            \\
\multicolumn{1}{c|}{5000}                 &  6.9 & 0 & 	\textbf{93.3} & \textbf{62.4}	& 78.1 & 2.4	& 5.9 & 0	& 7.1 & 0	& \textbf{3.3} & \textbf{21.1}                              \\
\multicolumn{1}{c|}{7500}                 &  5.7 & 1.2 & 	55.7 & 1.2 & 4.3 & 0	& 8.4 & 0.5	& 4.3 & 0	& 4.6 & 34.3                              \\
\multicolumn{1}{c|}{10000}                 &  5.9 & 0.7	& 10.3 & 0	& 5.8 & 31.7	& 7.2 & 1.4	& 6.6 & 0	& 2.9 & 29.4                              \\
\addlinespace[-0.22em]
\bottomrule[0.68pt]
\end{tabular}}
\end{table*}

\begin{table*}[t]
\caption{Search for the best results by the grid-search for NAD on ImageNet.}
\label{grid search nad imagenet}
\footnotesize
\setlength{\tabcolsep}{3.8mm}{
\begin{tabular}{l|ll|ll|ll|ll|ll|ll}
\toprule[0.68pt]
\addlinespace[0pt]
\multicolumn{1}{c|}{\multirow{2}{*}{$\beta$}} & \multicolumn{2}{c|}{BadNets}                    & \multicolumn{2}{c|}{Blend}                            & \multicolumn{2}{c|}{WaNet}                           & \multicolumn{2}{c|}{IAB}                         & \multicolumn{2}{c|}{Refool}                           & \multicolumn{2}{c}{CLB}                                 \\ \cline{2-13} 
\multicolumn{1}{c|}{}                        & \multicolumn{1}{c}{ACC} & \multicolumn{1}{c|}{ASR} & \multicolumn{1}{c}{ACC} & \multicolumn{1}{c|}{ASR} & \multicolumn{1}{c}{ACC} & \multicolumn{1}{c|}{ASR} & \multicolumn{1}{c}{ACC} & \multicolumn{1}{c|}{ASR} & \multicolumn{1}{c}{ACC} & \multicolumn{1}{c|}{ASR} & \multicolumn{1}{c}{ACC} & \multicolumn{1}{c}{ASR} \\ \hline
\multicolumn{1}{c|}{500}                 &  64.1 & 6.22 & 	\textbf{64.8} & \textbf{0.3}	& \textbf{63.8} & \textbf{1.3}	& 63.1 & 4.8 &	\textbf{63.7} & \textbf{0.3}	& 63.4 & 3.3                             \\
\multicolumn{1}{c|}{1000}                 &  \textbf{65.1} & \textbf{5.1} & 	63.6 & 0.6	& 62.8 & 0.7	& 63.4 & 1.1	 & 62.5 & 0	& 62.2 & 1.9                              \\
\multicolumn{1}{c|}{1500}                 &  61.6 & 4.2	& 62.27 & 0.5	& 62.2 & 0.8	& \textbf{63.8} & \textbf{0.6} & 	60.8 & 0	& 61.8 & 4.6                             \\
\multicolumn{1}{c|}{2000}                 &  60.1 & 2.1 & 	59.6 & 0.4	& 59.7 & 1.2	& 60.6 & 0.3 &	59.5 & 0.3	& \textbf{62.7} & \textbf{1.7}                             \\
\multicolumn{1}{c|}{2500}                 &  54.5 & 1.5& 	57.5 & 0	& 56.4 & 0.5 & 57.9 & 0.2 &	58.5 & 0.1	& 53.2 & 1.3                              \\
\multicolumn{1}{c|}{5000}                 &  51.7 & 3.2 &	51.5 & 0.4	& 50.2 & 0.5	& 48.5 & 0.6 &	51.5 & 0	& 47.1 & 0.6                             \\
\multicolumn{1}{c|}{7500}                 &  43.8 & 1.8	& 44.7 & 0	& 38.2 & 0.6	& 43.4 & 0.4	& 41.4 & 0	& 40.9 & 0.2                              \\
\multicolumn{1}{c|}{10000}                 &  33.8 & 1.4	& 33.9 & 0.6	& 41.1 & 1.2	& 35.4 & 0.6 & 	35.2 & 0.1 &	37.1 & 0                              \\
\addlinespace[-0.22em]
\bottomrule[0.68pt]
\end{tabular}}
\end{table*}

\begin{table*}[t]
\caption{Search for the best results by the grid-search for NAD on VGGFace2.}
\label{grid search nad vggface2}
\footnotesize
\setlength{\tabcolsep}{3.8mm}{
\begin{tabular}{l|ll|ll|ll|ll|ll|ll}
\toprule[0.68pt]
\addlinespace[0pt]
\multicolumn{1}{c|}{\multirow{2}{*}{$\beta$}} & \multicolumn{2}{c|}{BadNets}                    & \multicolumn{2}{c|}{Blend}                            & \multicolumn{2}{c|}{WaNet}                           & \multicolumn{2}{c|}{IAB}                         & \multicolumn{2}{c|}{Refool}                           & \multicolumn{2}{c}{CLB}                                 \\ \cline{2-13} 
\multicolumn{1}{c|}{}                        & \multicolumn{1}{c}{ACC} & \multicolumn{1}{c|}{ASR} & \multicolumn{1}{c}{ACC} & \multicolumn{1}{c|}{ASR} & \multicolumn{1}{c}{ACC} & \multicolumn{1}{c|}{ASR} & \multicolumn{1}{c}{ACC} & \multicolumn{1}{c|}{ASR} & \multicolumn{1}{c}{ACC} & \multicolumn{1}{c|}{ASR} & \multicolumn{1}{c}{ACC} & \multicolumn{1}{c}{ASR} \\ \hline
\multicolumn{1}{c|}{500}                 &  42.6 & 5.5 & 49.1 & 10.9 & 48.7 & 3.7 & \textbf{43.1} & \textbf{5.5} &	50.8 & 4.3 & 42.9 & 15.1                            \\
\multicolumn{1}{c|}{1000}                 &  53.4 & 10.6 & 46.2 & 8.1 & 43.9 & 12.2 & 25.7 & 5.1 &	52.9 & 2.1 & 46.3 & 18.4                              \\
\multicolumn{1}{c|}{1500}                 &  48.5 & 5.8 & 43.7 & 5.6 & \textbf{50.4} & \textbf{4.2} & 37.9 & 2.3 &	\textbf{53.0} & \textbf{3.1} & 48.7 & 15.6                            \\
\multicolumn{1}{c|}{2000}                 &  \textbf{56.1} & \textbf{6.5} & 47.3 & 4.1 & 43.7 & 4.1 & 44.7 & 8.8 &	52.8 & 5.6 & 34.6 & 3.0                            \\
\multicolumn{1}{c|}{2500}                 & 41.8 & 1.4 & \textbf{50.8} & \textbf{7.3} & 43.9 & 3.7 & 42.7 & 8.9 &	53.3 & 7.1 & \textbf{40.0} & \textbf{3.3}                     \\
\multicolumn{1}{c|}{5000}                 & 53.8 & 11.4 & 28.9 & 2.6 & 41.1 & 2.2 & 31.9 & 5.3 &	52.5 & 4.3 & 40.2 & 11.2                      \\
\multicolumn{1}{c|}{7500}                 & 50.7 & 2.7	& 47.9 & 2.5 & 49.5 & 2.8 & 32.6 & 40.9 &	53.2 & 5.3 & 38.7 & 2.2                       \\
\multicolumn{1}{c|}{10000}                 & 52.7 & 8.5 & 45.5 & 6.4 & 40.7 & 5.5	& 30.6 & 12.3 &	50.5 & 5.1	& 27.2 & 13.9                              \\
\addlinespace[-0.22em]
\bottomrule[0.68pt]
\end{tabular}}
\end{table*}

\begin{table*}[t]
\caption{Search for the best results by the grid-search for ABL on CIFAR-10.}
\label{grid search abl cifar}
\footnotesize
\setlength{\tabcolsep}{3.8mm}{
\begin{tabular}{l|ll|ll|ll|ll|ll|ll}
\toprule[0.68pt]
\addlinespace[0pt]
\multicolumn{1}{c|}{\multirow{2}{*}{$\gamma$}} & \multicolumn{2}{c|}{BadNets}                    & \multicolumn{2}{c|}{Blend}                            & \multicolumn{2}{c|}{WaNet}                           & \multicolumn{2}{c|}{IAB}                         & \multicolumn{2}{c|}{Refool}                           & \multicolumn{2}{c}{CLB}                                 \\ \cline{2-13} 
\multicolumn{1}{c|}{}                        & \multicolumn{1}{c}{ACC} & \multicolumn{1}{c|}{ASR} & \multicolumn{1}{c}{ACC} & \multicolumn{1}{c|}{ASR} & \multicolumn{1}{c}{ACC} & \multicolumn{1}{c|}{ASR} & \multicolumn{1}{c}{ACC} & \multicolumn{1}{c|}{ASR} & \multicolumn{1}{c}{ACC} & \multicolumn{1}{c|}{ASR} & \multicolumn{1}{c}{ACC} & \multicolumn{1}{c}{ASR} \\ \hline
\multicolumn{1}{c|}{0}                 &  \textbf{93.8} & \textbf{1.1}	& 90.9 & 2.1 & \textbf{84.1} & \textbf{2.2} & \textbf{93.4} & \textbf{5.1} &	79.9 & 99.7	& \textbf{86.6} & \textbf{1.3}                               \\
0.1                                        &  66.2 & 100	& \textbf{91.9} & \textbf{1.6} & 75.7 & 100 & 88.2 & 100 &	\textbf{82.7} & \textbf{1.3} & 79.9 & 14.4                                  \\ 
0.2                                          & 70.8 & 100 & 81.3 & 99.3 & 80.1 & 100 & 	85.7 & 100 & 80.3 & 99.1 & 83.8 & 7.67                                \\ 
0.3                                        &  72.8 & 100 & 80.1 & 99.3 & 77.7 & 100 & 80.6 & 100	& 69.7 & 99.9 & 83.8 & 25.6                                  \\ 
0.4                                          &  64.9 & 100 & 86.8 & 99.2 & 78.6 & 100 &	73.8 & 100 & 79.5 & 99.9 & 78.3 & 22.6                               \\ 
0.5                                        &  71.9 & 100 & 74.5 & 99.9 & 77.5 & 100 & 76.9 & 100 &	71.9 & 99.9 & 77.9 & 12.5                                  \\ 
\addlinespace[-0.22em]
\bottomrule[0.68pt]
\end{tabular}}
\end{table*}

\begin{table*}[t]
\caption{Search for the best results by the grid-search for ABL on GTSRB.}
\label{grid search abl gtsrb}
\footnotesize
\setlength{\tabcolsep}{3.8mm}{
\begin{tabular}{l|ll|ll|ll|ll|ll|ll}
\toprule[0.68pt]
\addlinespace[0pt]
\multicolumn{1}{c|}{\multirow{2}{*}{$\gamma$}} & \multicolumn{2}{c|}{BadNets}                    & \multicolumn{2}{c|}{Blend}                            & \multicolumn{2}{c|}{WaNet}                           & \multicolumn{2}{c|}{IAB}                         & \multicolumn{2}{c|}{Refool}                           & \multicolumn{2}{c}{CLB}                                 \\ \cline{2-13} 
\multicolumn{1}{c|}{}                        & \multicolumn{1}{c}{ACC} & \multicolumn{1}{c|}{ASR} & \multicolumn{1}{c}{ACC} & \multicolumn{1}{c|}{ASR} & \multicolumn{1}{c}{ACC} & \multicolumn{1}{c|}{ASR} & \multicolumn{1}{c}{ACC} & \multicolumn{1}{c|}{ASR} & \multicolumn{1}{c}{ACC} & \multicolumn{1}{c|}{ASR} & \multicolumn{1}{c}{ACC} & \multicolumn{1}{c}{ASR} \\ \hline
\multicolumn{1}{c|}{0}                 &  \textbf{97.1} & \textbf{0} & 95.6 & 12.5 & 94.2 & 10.9 & 93.3 & 100	 & 95.4 & 0 & \textbf{90.4} & \textbf{2.3}                           \\
0.1                                        &  80.9 & 100 & 94.2 & 99.9 & 91.8 & 100 & 91.4 & 100 &	96.2 & 0 & 86.7 & 100                                \\ 
0.2                                          &  85.5 & 100 & 94.4 & 99.9 & 91.9 & 100 &	90.9 & 100 & 95.7 & 0 & 80.1 & 100                             \\ 
0.3                                        &  97.1 & 0 & \textbf{97.1} & \textbf{0.5} & \textbf{97.0} & \textbf{0.4} & 97.1 & 0.8 &	\textbf{96.2} & \textbf{0} & 75.2 & 100                                \\ 
0.4                                          &  96.8 & 0 & 96.9 & 0.7 & 96.7 & 0 & \textbf{97.4} & \textbf{0.6} &	95.5 & 0 & 72.3 & 100                             \\ 
0.5                                        &  97.1 & 0 & 96.7 & 2.1 & 96.1 & 0.2 & 96.9 & 2.1 &	95.6 & 0 & 69.1 & 100                                \\ 
\addlinespace[-0.22em]
\bottomrule[0.68pt]
\end{tabular}}
\end{table*}

\begin{table*}[t]
\caption{Search for the best results by the grid-search for ABL on ImageNet.}
\label{grid search abl imagenet}
\footnotesize
\setlength{\tabcolsep}{3.8mm}{
\begin{tabular}{l|ll|ll|ll|ll|ll|ll}
\toprule[0.68pt]
\addlinespace[0pt]
\multicolumn{1}{c|}{\multirow{2}{*}{$\gamma$}} & \multicolumn{2}{c|}{BadNets}                    & \multicolumn{2}{c|}{Blend}                            & \multicolumn{2}{c|}{WaNet}                           & \multicolumn{2}{c|}{IAB}                         & \multicolumn{2}{c|}{Refool}                           & \multicolumn{2}{c}{CLB}                                 \\ \cline{2-13} 
\multicolumn{1}{c|}{}                        & \multicolumn{1}{c}{ACC} & \multicolumn{1}{c|}{ASR} & \multicolumn{1}{c}{ACC} & \multicolumn{1}{c|}{ASR} & \multicolumn{1}{c}{ACC} & \multicolumn{1}{c|}{ASR} & \multicolumn{1}{c}{ACC} & \multicolumn{1}{c|}{ASR} & \multicolumn{1}{c}{ACC} & \multicolumn{1}{c|}{ASR} & \multicolumn{1}{c}{ACC} & \multicolumn{1}{c}{ASR} \\ \hline
\multicolumn{1}{c|}{0}                 &  80.2 & 0.1 & 83.6 & 100 & 84.8 & 99.8 & 79.7 & 3.9 &	\textbf{76.2} & \textbf{0.2} & \textbf{82.8} & \textbf{0.8}                           \\
0.1                                        &  82.9 & 0 & 83.4 & 100 & 82.5 & 100 & 81.8 & 1.1	& 78.5 & 0.3 & 82.7 & 64.3                                \\ 
0.2                                          & 73.4 & 100 & 82.1 & 100 & 81.5 & 99.9 &	82.6 & 99.9 & 79.4 & 0.5 & 82.8 & 56.2                              \\ 
0.3                                        &  82.8 & 0 & 75.9 & 1.0 & 69.2 & 2.3 & 80.6 & 0	 & 80.1 & 1.6	& 80.6 & 60.1                                \\ 
0.4                                          &  \textbf{83.1} & \textbf{0} & 78.6 & 1.1 & \textbf{74.9} & \textbf{1.1} & 81.7 & 0.1 & 	80.1 & 2.6	& 82.3 & 52.5                             \\ 
0.5                                        &  83.1 & 0.1 & \textbf{82.6} & \textbf{0.7} & 71.4 & 2.5 & \textbf{81.7} & \textbf{0}	& 80.4 & 2.7 & 80.2 & 59.4                                \\ 
\addlinespace[-0.22em]
\bottomrule[0.68pt]
\end{tabular}}
\end{table*}

\begin{table*}[t]
\caption{Search for the best results by the grid-search for ABL on VGGFace2.}
\label{grid search abl vggface2}
\footnotesize
\setlength{\tabcolsep}{3.8mm}{
\begin{tabular}{l|ll|ll|ll|ll|ll|ll}
\toprule[0.68pt]
\addlinespace[0pt]
\multicolumn{1}{c|}{\multirow{2}{*}{$\gamma$}} & \multicolumn{2}{c|}{BadNets}                    & \multicolumn{2}{c|}{Blend}                            & \multicolumn{2}{c|}{WaNet}                           & \multicolumn{2}{c|}{IAB}                         & \multicolumn{2}{c|}{Refool}                           & \multicolumn{2}{c}{CLB}                                 \\ \cline{2-13} 
\multicolumn{1}{c|}{}                        & \multicolumn{1}{c}{ACC} & \multicolumn{1}{c|}{ASR} & \multicolumn{1}{c}{ACC} & \multicolumn{1}{c|}{ASR} & \multicolumn{1}{c}{ACC} & \multicolumn{1}{c|}{ASR} & \multicolumn{1}{c}{ACC} & \multicolumn{1}{c|}{ASR} & \multicolumn{1}{c}{ACC} & \multicolumn{1}{c|}{ASR} & \multicolumn{1}{c}{ACC} & \multicolumn{1}{c}{ASR} \\ \hline
\multicolumn{1}{c|}{0}                 & 90.9 & 65.2 & \textbf{90.1} & \textbf{96.7} & 90.3 & 89.7 & 91.2 & 99.3	 & 90.9 & 56.1 & 91.4 & 0.3                            \\
0.1                                        &  90.4 & 99.9 & 90.6 & 97.6 & 91.8 & 100 & 91.2 & 79.1 &	\textbf{91.1} & \textbf{51.1}	& 90.6 & 0.3                                \\ 
0.2                                          & \textbf{91.2} & \textbf{19.6} & 90.2 & 100 & \textbf{92.6} & \textbf{74.6} &	\textbf{91.3} & \textbf{59.7} & 91.3 & 51.7 & 90.9 & 0.1                              \\ 
0.3                                        &   90.8 & 85.4 & 91.1 & 99.9 & 91.9 & 99.8 &	92.0 & 80.1 & 91.6 & 62.8 & 91.2 & 0.1                               \\ 
0.4                                          &  90.5 & 100 & 90.3 & 99.9 & 91.5 & 81.4 &	92.8 & 100 & 90.0 & 58.8 & 90.8 & 0                             \\ 
0.5                                        &  90.2 & 100 & 90.8 & 100 & 91.3 & 81.1 & 91.2 & 100	 & 90.2 & 61.2 & \textbf{91.3} & \textbf{0}                                \\ 
\addlinespace[-0.22em]
\bottomrule[0.68pt]
\end{tabular}}
\end{table*}

\begin{table*}[t]
\caption{Search for the best results by the grid-search for DPSGD on CIFAR-10.}
\label{grid search dpsgd cifar}
\footnotesize
\setlength{\tabcolsep}{3.8mm}{
\begin{tabular}{l|ll|ll|ll|ll|ll|ll}
\toprule[0.68pt]
\addlinespace[0pt]
\multicolumn{1}{c|}{\multirow{2}{*}{$\sigma$}} & \multicolumn{2}{c|}{BadNets}                    & \multicolumn{2}{c|}{Blend}                            & \multicolumn{2}{c|}{WaNet}                           & \multicolumn{2}{c|}{IAB}                         & \multicolumn{2}{c|}{Refool}                           & \multicolumn{2}{c}{CLB}                                 \\ \cline{2-13} 
\multicolumn{1}{c|}{}                        & \multicolumn{1}{c}{ACC} & \multicolumn{1}{c|}{ASR} & \multicolumn{1}{c}{ACC} & \multicolumn{1}{c|}{ASR} & \multicolumn{1}{c}{ACC} & \multicolumn{1}{c|}{ASR} & \multicolumn{1}{c}{ACC} & \multicolumn{1}{c|}{ASR} & \multicolumn{1}{c}{ACC} & \multicolumn{1}{c|}{ASR} & \multicolumn{1}{c}{ACC} & \multicolumn{1}{c}{ASR} \\ \hline
\multicolumn{1}{c|}{0}                 &   10.1 
 & 100	& 10.9 & 100	& 10.2 & 100 &	10.1 & 100	& 10.9 & 100	& 10.1 & 0                            \\
0.01                                        &  85.5 & 100	& 84.4 & 87.2	& 84.3 & 99.9	&\textbf{85.9} & \textbf{99.7}& 	83.2 & 91.7& 	84.6 & 38.2                          \\ 
0.05                                          & 79.2 & 99.8	& 68.5 & 63.1	& 76.4 & 94.5 &	77.7 & 99.8	& 76.8 & 82.7& 	76.7 & 10.3                             \\ 
0.1                                        & 67.2 & 100	& 64.3 & 75.2	& 63.5 & 75.4	 & 65.3 & 99.8& 	65.4 & 70.2	& 65.1 & 8.2                                \\ 
0.2                                          &  \textbf{55.9} & \textbf{10.9}	& \textbf{56.7} & \textbf{37.0}	& \textbf{55.1} & \textbf{15.8} & 	54.4 & 99.8	& \textbf{55.4} &\textbf{59.2}	& \textbf{55.7} & \textbf{7.6}                           \\ 
\addlinespace[-0.22em]
\bottomrule[0.68pt]
\end{tabular}}
\end{table*}

\end{document}